\documentclass{article}

\usepackage{microtype}
\usepackage{graphicx}
\usepackage{booktabs} 
\usepackage{amsmath}
\usepackage{amsfonts}
\usepackage{amssymb}
\usepackage{amsthm}
\usepackage{mathrsfs}
\usepackage{color}
\usepackage{float} 
\usepackage{caption}
\usepackage{soul}
\usepackage[T1]{fontenc}
\usepackage[font=footnotesize]{subcaption}

\usepackage{hyperref}

\newtheorem{lemma}{Lemma}
\newtheorem{theorem}{Theorem}
\newtheorem{proposition}{Proposition}

\usepackage[accepted]{icml2018}

\icmltitlerunning{A Theoretical Explanation for Perplexing Behaviors of Backpropagation-based Visualizations}

\graphicspath{{./figures/}}

\begin{document}

\twocolumn[
\icmltitle{A Theoretical Explanation for Perplexing Behaviors of Backpropagation-based Visualizations}




\begin{icmlauthorlist}
    \icmlauthor{Weili Nie}{rece}
    \icmlauthor{Yang Zhang}{rcs}
    \icmlauthor{Ankit B. Patel}{rece,bcm}
\end{icmlauthorlist}

\icmlaffiliation{rece}{Department of Electrical and Computer Engineering, Rice University, Houston, USA.}
\icmlaffiliation{rcs}{Department of Computer Science, Rice University, Houston, USA.}
\icmlaffiliation{bcm}{Department of Neuroscience, Baylor College of Medicine, Houston, USA}
\icmlcorrespondingauthor{Weili Nie}{wn8@rice.edu}
\icmlcorrespondingauthor{Ankit B. Patel}{abp4@rice.edu}

\icmlkeywords{saliency map, guided backpropagation, deconvolutional network, explainable AI}

\vskip 0.3in
]



\printAffiliationsAndNotice{}  

\begin{abstract}
Backpropagation-based visualizations have been proposed to interpret convolutional neural networks (CNNs), however a theory is missing to justify their behaviors:
Guided backpropagation (GBP) and deconvolutional network (DeconvNet) generate more human-interpretable but less class-sensitive visualizations than saliency map. 
Motivated by this, we develop a theoretical explanation revealing that GBP and DeconvNet are essentially doing (partial) image recovery which is unrelated to the network decisions. 
Specifically, our analysis shows that the backward ReLU introduced by GBP and DeconvNet, and the local connections in CNNs are the two main causes of compelling visualizations.
Extensive experiments are provided that support the theoretical analysis.
\end{abstract}

\section{Introduction}
\label{submission}

Driven by massive data and computational resources, modern convolutional neural networks (CNNs) and other network architectures have achieved many outstanding results, such as image recognition  \cite{krizhevsky2012imagenet}, 
neural machine translation \cite{sutskever2014sequence}, and playing Go games \cite{silver2016mastering}, etc. Despite their extensive applications, these neural networks are always considered as black boxes. 
Interpretability used to be for its own sake; now, due to safety-critical applications such as self-driving cars and tumor diagnosis, it is no longer satisfying to have a black box that is unaccountable for its decisions.
The demand for explainable artificial intelligence (XAI) \cite{gunning2017explainable} -- human interpretable explanations of model decisions -- has driven the development of visualization techniques, 
including image synthesis via activation maximization \cite{simonyan2013deep,johnson2016perceptual, nguyen2016synthesizing} and backpropagation-based visualizations \cite{simonyan2013deep,zeiler2014visualizing, springenberg2014striving, shrikumar2017learning, kindermans2017patternnet}. 

The basic idea of backpropagation-based visualizations is to highlight class-relevant pixels by propagating the network output back to the input image space. The intensity changes of these pixels have the most significant impact on network decisions. Specifically,
\cite{simonyan2013deep} visualizes the spatial support of a given class in a given image, i.e. saliency map, by using the true gradient which masks out negative entries of bottom data via the forward ReLU. Despite its simplicity, the results of saliency map are normally very noisy which makes the interpretation difficult. \cite{zeiler2014visualizing} visualize the reverse mapping from feature activities back to the input pixel space with the deconvolutional network (DeconvNet) method. The basic idea of DeconvNet is to mask out negative entries of the top gradients by resorting to the backward ReLU. \cite{springenberg2014striving} proposed the Guided Backpropagation (GBP) method which combines the above two methods: 
by considering both the forward and backward ReLUs,
it masks out the values for which either top gradients or bottom data are negative and produces sharper visualizations. 
More recently, DeepLift \cite{shrikumar2017learning} and PatternNet \cite{kindermans2017patternnet} have been proposed to further improve the visual quality of backpropagation-based methods.

This class of backpropagation-based visualizations, in particular GBP and DeconvNet, has attracted a lot of attention in both the deep learning community and other fields \cite{szegedy2013intriguing, dosovitskiy2016inverting, selvaraju2016grad,fong2017interpretable,kraus2016classifying}. 
Despite their good visual quality, the question of how they are actually related to the decision-making has remained largely unexplored. Do the pretty visualizations actually tell us reliably about what the network is doing internally?
Our experiments have confirmed previous observations \cite{mahendran2016salient,selvaraju2016grad,samek2017evaluating} that saliency map is indeed very sensitive to the change of class labels, while GBP and DeconvNet, though their visualization results are much cleaner than saliency map, remain almost the same given different class labels. 
It seems that the visual quality improvement of backpropagation-based methods is sacrificing the ability of highlighting important pixels to a specific output class. 
In this sense, GBP and DeconvNet may be unreliable in interpreting how deep neural networks make classification decisions. 

The most commonly used explanation for these visualizations is to approximate the neural networks with a linear function \cite{simonyan2013deep,kindermans2017patternnet}, where the derivative of output with respect to input image is just the weight vector of the model. In such sense, the backpropagation-based methods can be regarded as visualizing the \textit{learned weights}.
But apparently the approximate linear model is too simplistic to reflect the highly nonlinear property of deep neural networks. For example, GBP and DeconvNet essentially apply the same algorithm as saliency map, but treat ReLU, the nonlinear activation, differently. The linear model explanation thus cannot answer questions regarding why GBP and DeconvNet outperform saliency map in terms of visual quality whereas they are less class-sensitive than saliency map, as both of them reduce to saliency map in a linear model.
Therefore, we need a more complex model, which should at least capture the impact of both forward ReLU and backward ReLU, to better understand what the main causes of their visually compelling results are and what information, if not the classification decisions, we can extract from these visualizations.

\textbf{Our contributions.} We provide a theoretical explanation for why GBP and DeconvNet generate more human-interpretable but less class-sensitive visualizations than saliency map. Specifically, our analysis reveals that GBP and DeconvNet are essentially doing \textit{(partial) image recovery} instead of highlighting class-relevant pixels or visualizing the learned weights, which means in principle they are unrelated to the decision-making of neural networks. We also find that it is the \textit{backward ReLU} introduced by either GBP or DeconvNet, together with the \textit{local connections} in CNNs that results in crisp visualizations.
In particular, we explain how DeconvNet also relies on the max-pooling to recover the input.  
Finally, we do extensive experiments to support our theory and further reveal more detailed properties of these backpropagation-based visualizations\footnote{Code is available at https://github.com/weilinie/BackpropVis}.


\section{Backpropagation-based Visualizations}\label{Sec_obs}

In this section, we first give formal definitions of backpropagation-based visualizations: saliency map, DeconvNet and GBP, and then compare their empirical behaviors.

\begin{figure}
    \centering
    \includegraphics[width=\linewidth]{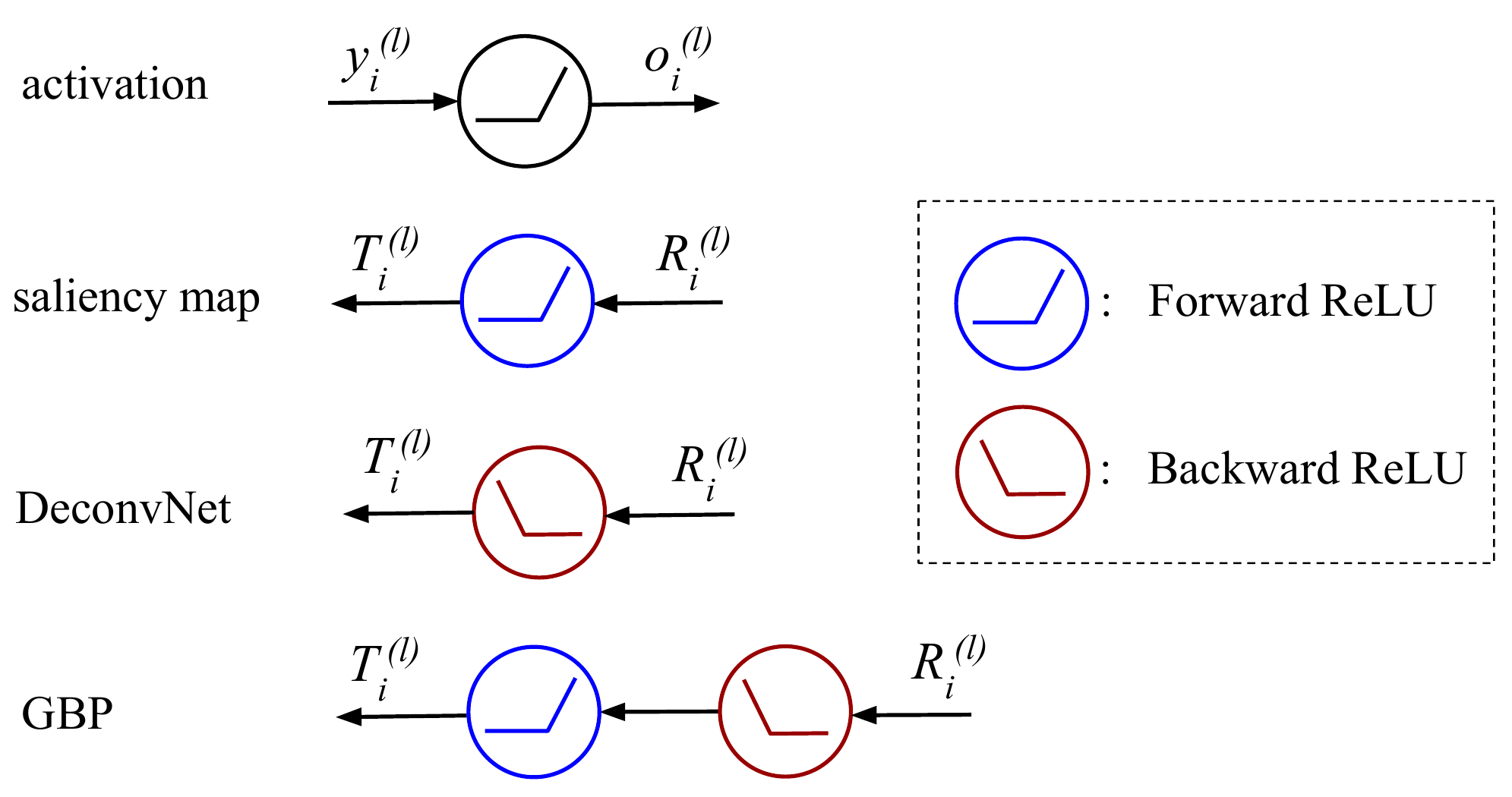}
    \caption{Illustration of how backpropagation-based methods propagate back through the $i$-th nonlinear activation in the $l$-th layer with input $y^{(l)}_i$ and output $o^{(l)}_i$, where $T^{(l)}_i$ denotes the (modified) gradient after passing through the activation and $R^{(l)}_i$ denotes the top gradient before the activation. }
    \label{model_vis}
\end{figure}

\subsection{Formal Definitions}

The key difference of backpropagation-based methods is the way they propagate the output score back through the ReLU activations. As illustrated by Figure \ref{model_vis}, we  consider the $i$-th ReLU activation in the $l$-th layer with its input $y^{(l)}_i$ and its output $o^{(l)}_i$ and denote by $\sigma(t) = \max(t, 0)$ the ReLU activation. 
Also, denote by $R^{(l)}_i$ the top gradient before activation, i.e., gradient of the output score with respect to $o^{(l)}_i$ and denote by $T^{(l)}_i$ the (modified) gradient after activation, i.e., gradient of the output score with respect to $y^{(l)}_i$. 
Then in the gradient calculations, the corresponding \textit{forward ReLU} could be formally defined as a function
$$\sigma^{(l)}_{f,i}(t) \triangleq \mathbb{I}\left({y^{(l)}_i}\right) t$$
where $\mathbb{I}(\cdot)$ is the indicator function and the corresponding \textit{backward ReLU} could be formally defined as a function
$$\sigma^{(l)}_{b,i}(t) \triangleq
\mathbb{I}\left({R^{(l)}_i}\right) t$$
Therefore, the formal definition of backpropagation-based methods for propagating the output score back through the $i$-th ReLU activation in the $l$-th layer is 
\begin{align*}
    \begin{split}
        T^{(l)}_i = \begin{cases}
            \sigma^{(l)}_{f,i} \left( R^{(l)}_i \right) & \text{for saliency map} \\
            \sigma^{(l)}_{b,i}\left( R^{(l)}_i \right)  & \text{for DeconvNet} \\
            \sigma^{(l)}_{f,i}\left( \sigma^{(l)}_{b,i} \left(R^{(l)}_i \right) \right) & \text{for GBP} \\
        \end{cases}
    \end{split}
\end{align*}
which can be further uniformly formulated as 
\begin{align} \label{formal_def}
    T^{(l)}_i = h\left(R^{(l)}_i\right) \frac{\partial g\left(y^{(l)}_i\right)}{\partial y^{(l)}_i}
\end{align}
where the two functions $h(\cdot)$ and $g(\cdot)$ are defined as
\begin{align} \label{two_funcs}
    \begin{split}
        h(t) &= \begin{cases}
t & \text{for saliency map}\\
\sigma(t) & \text{for DeconvNet and GBP}
\end{cases} \\
        g(t) &= \begin{cases}
t & \text{for DeconvNet}\\
\sigma(t) & \text{for saliency map and GBP}
\end{cases} \\
    \end{split}
\end{align}

\subsection{Empirical Observations}

\begin{figure}[!t]
	\begin{minipage}[c]{0.2\linewidth}
	    \centering
	    \begin{subfigure}[b]{\linewidth}
	    \centering
		\includegraphics[width=\linewidth]{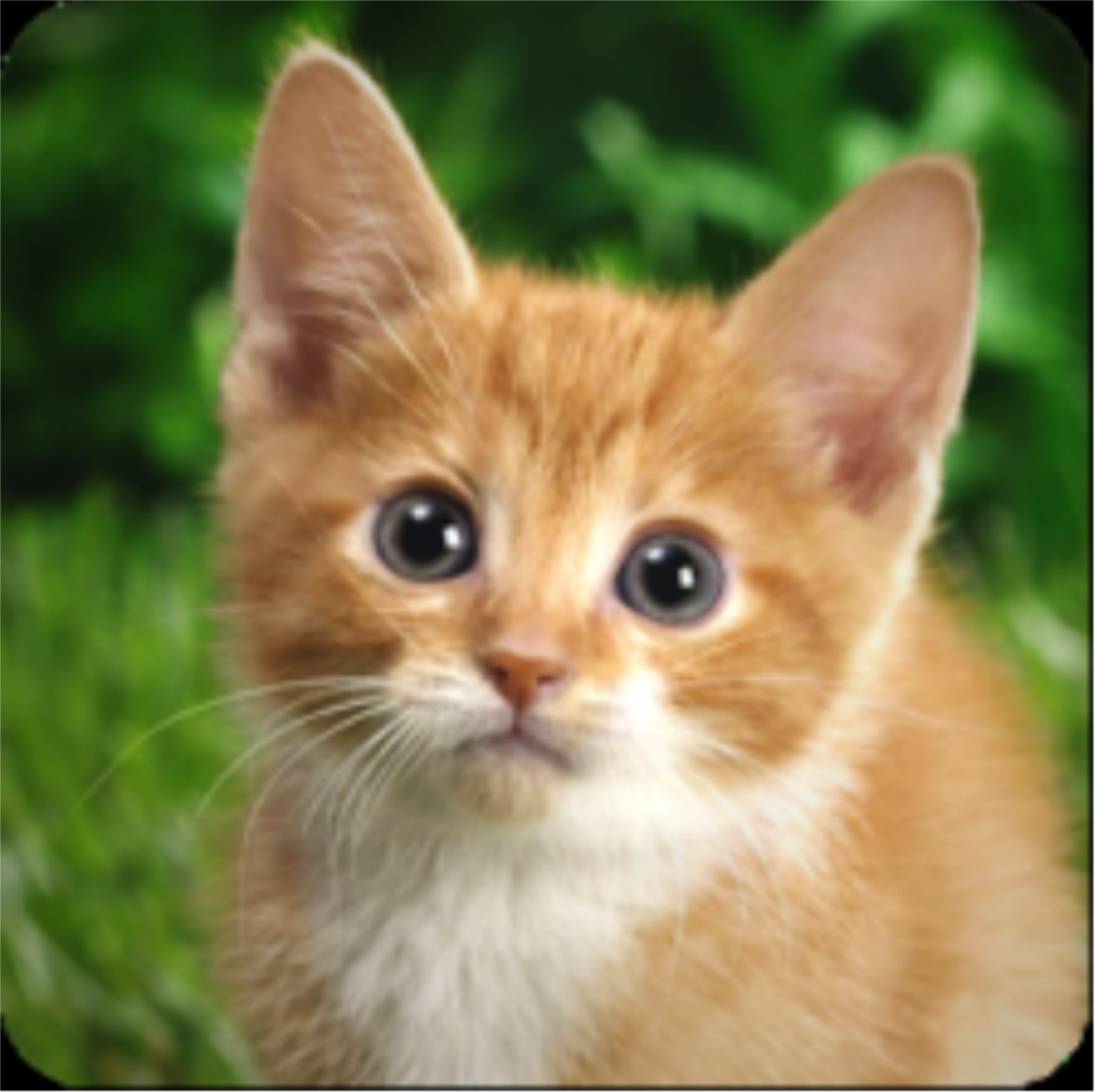}
		\caption*{tabby}
		\end{subfigure}
	\end{minipage}
	\hfill
	\begin{minipage}[c]{0.8\linewidth}
    \centering
	\begin{subfigure}[b]{0.27\linewidth}
		\centering
		\includegraphics[width=\linewidth]{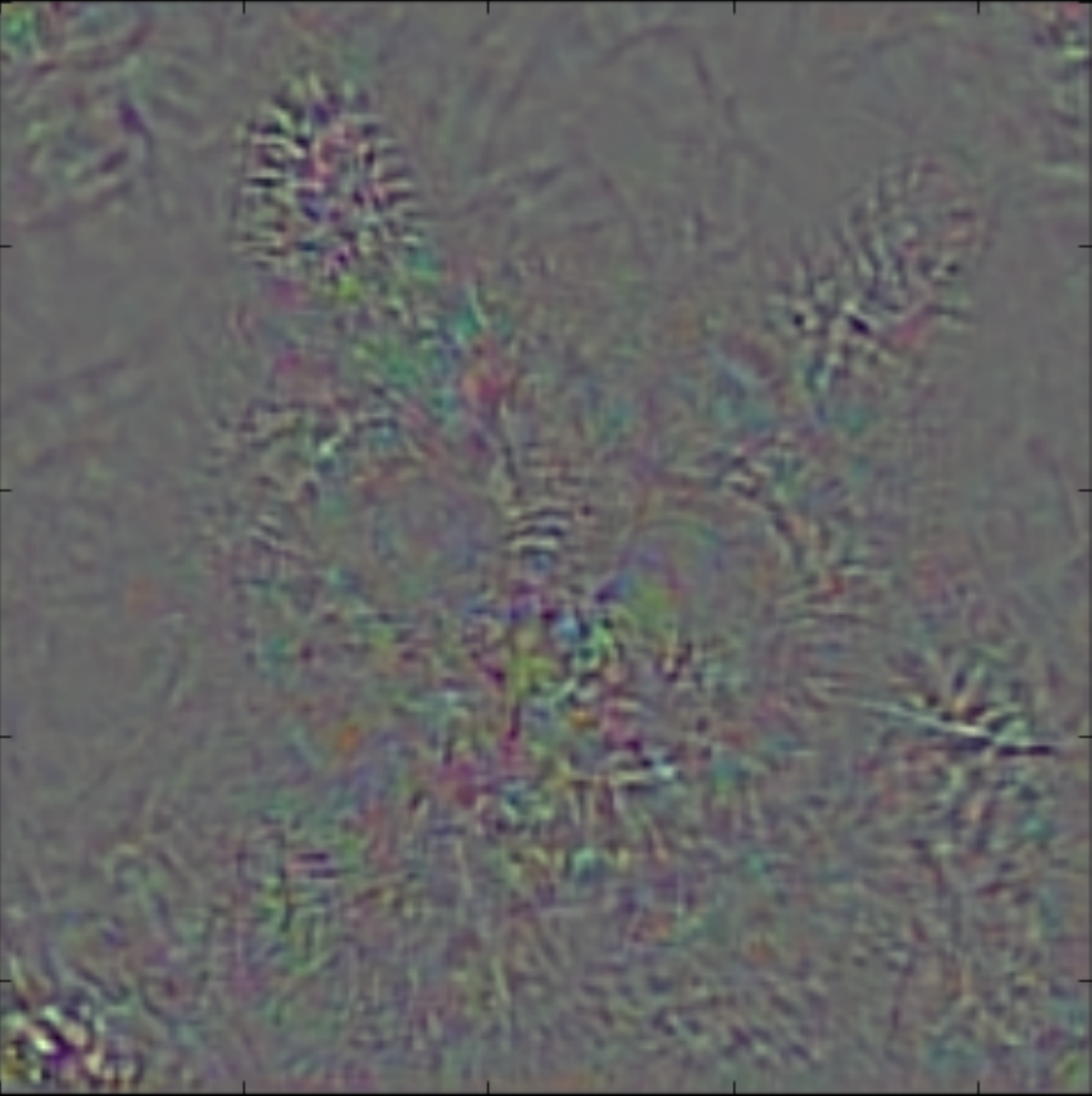}
		\caption*{\footnotesize Sal-max}
	\end{subfigure}
	\centering
	\begin{subfigure}[b]{0.27\linewidth}
		\centering
		\includegraphics[width=\linewidth]{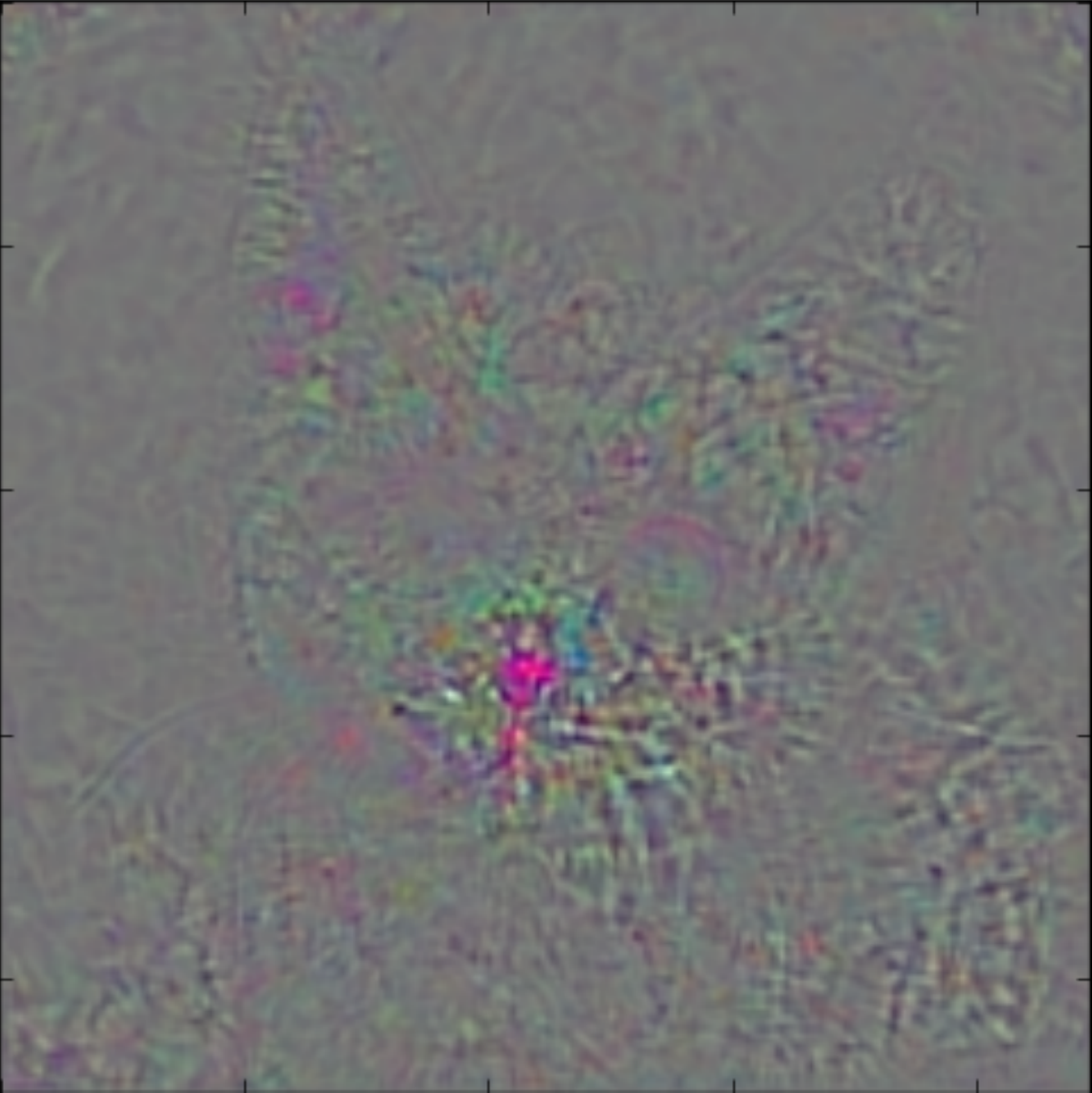}
		\caption*{\footnotesize Sal-482}
	\end{subfigure}
	\centering
	\begin{subfigure}[b]{0.27\linewidth}
		\centering
		\includegraphics[width=\linewidth]{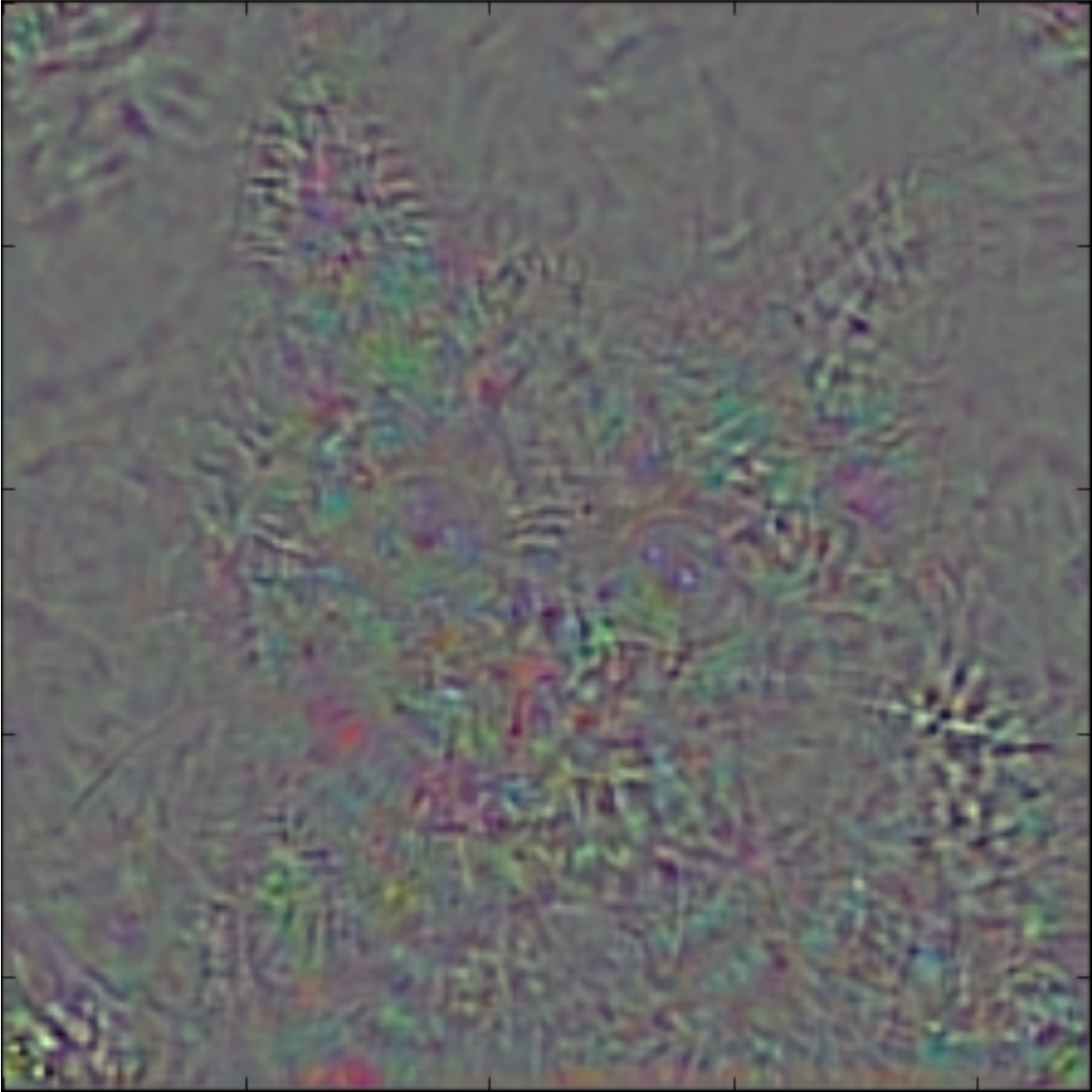}
		\caption*{\footnotesize{Sal-560}}
	\end{subfigure}
	
	\begin{subfigure}[b]{0.27\linewidth}
	    \centering
	    \includegraphics[width=\linewidth]{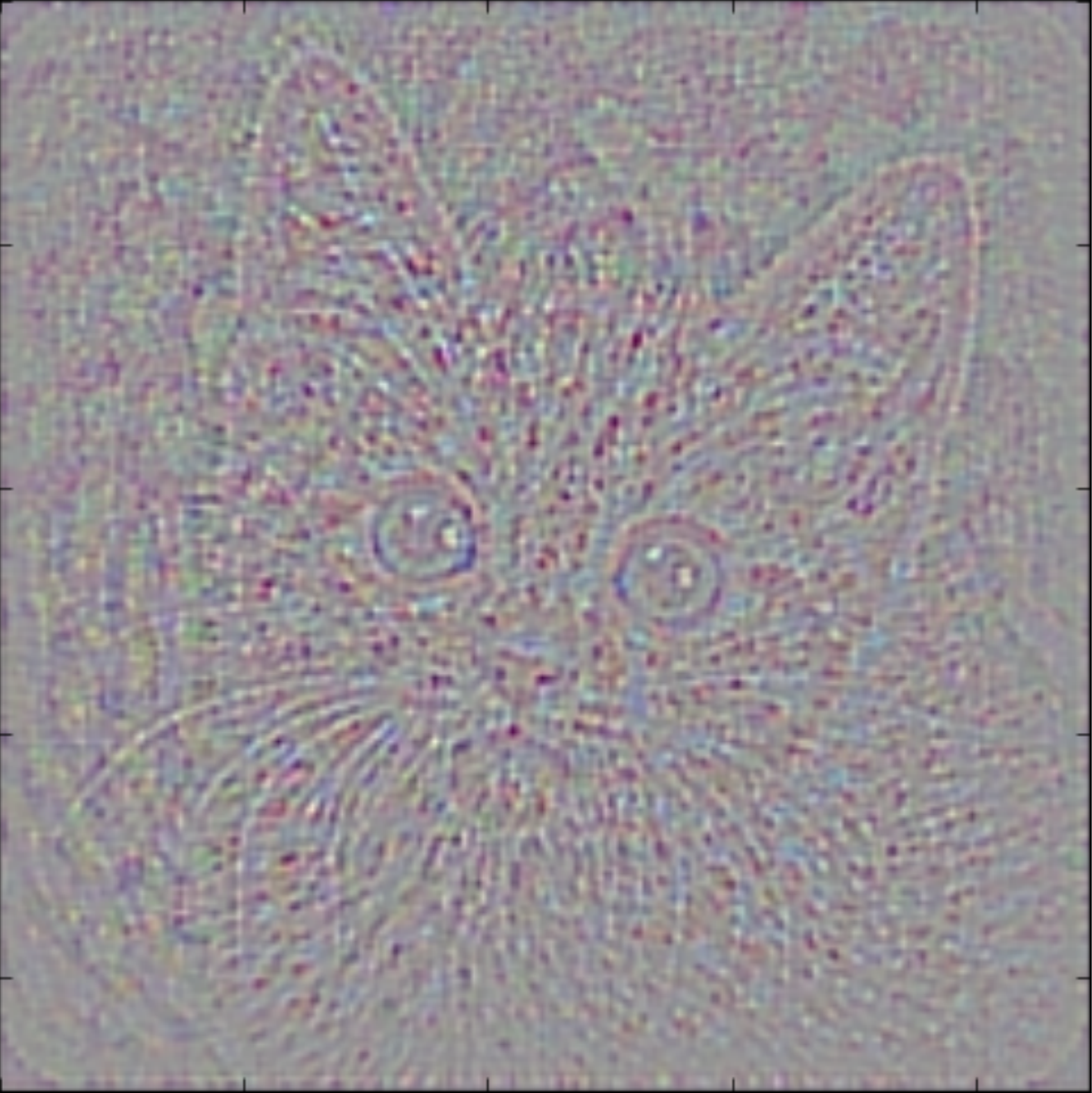}
	    \caption*{\footnotesize{Deconv-max}}
	\end{subfigure}
	\centering
	\begin{subfigure}[b]{0.27\linewidth}
		\centering
		\includegraphics[width=\linewidth]{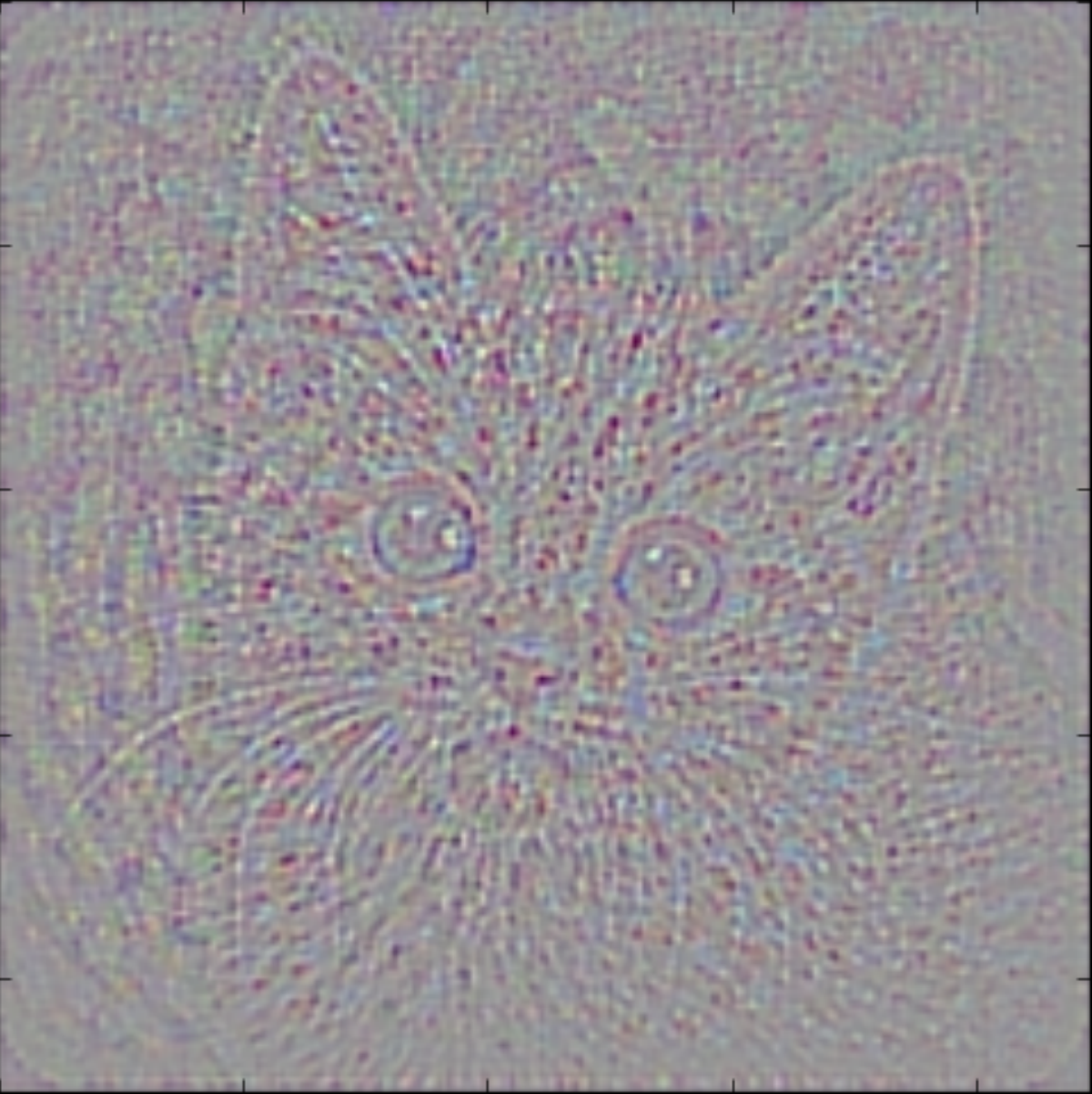}
		\caption*{\footnotesize Deconv-482}
	\end{subfigure}
	\centering
	\begin{subfigure}[b]{0.27\linewidth}
		\centering
		\includegraphics[width=\linewidth]{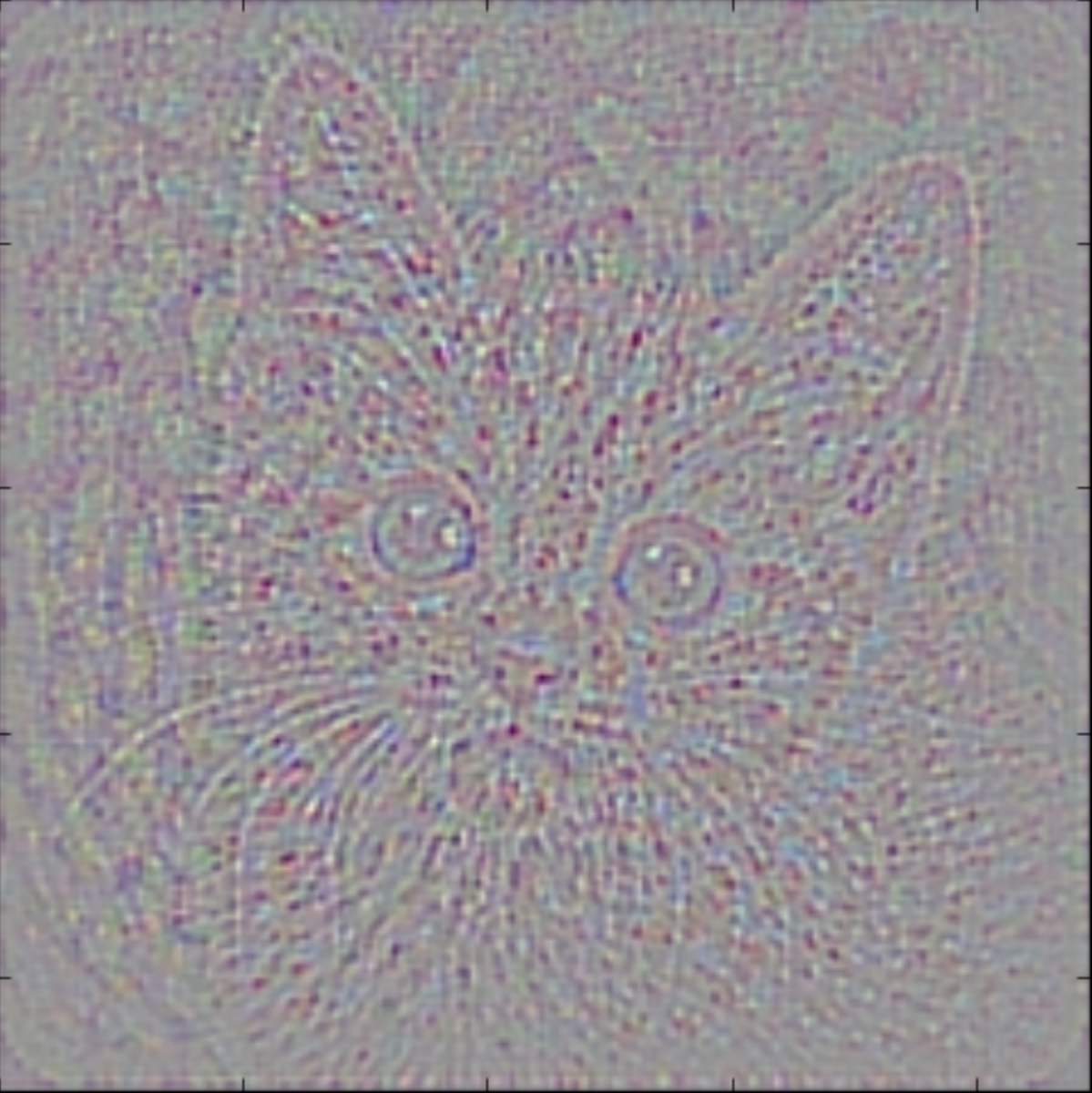}
		\caption*{\footnotesize Deconv-560}
	\end{subfigure}
	
	\begin{subfigure}[b]{0.27\linewidth}
		\centering
		\includegraphics[width=\linewidth]{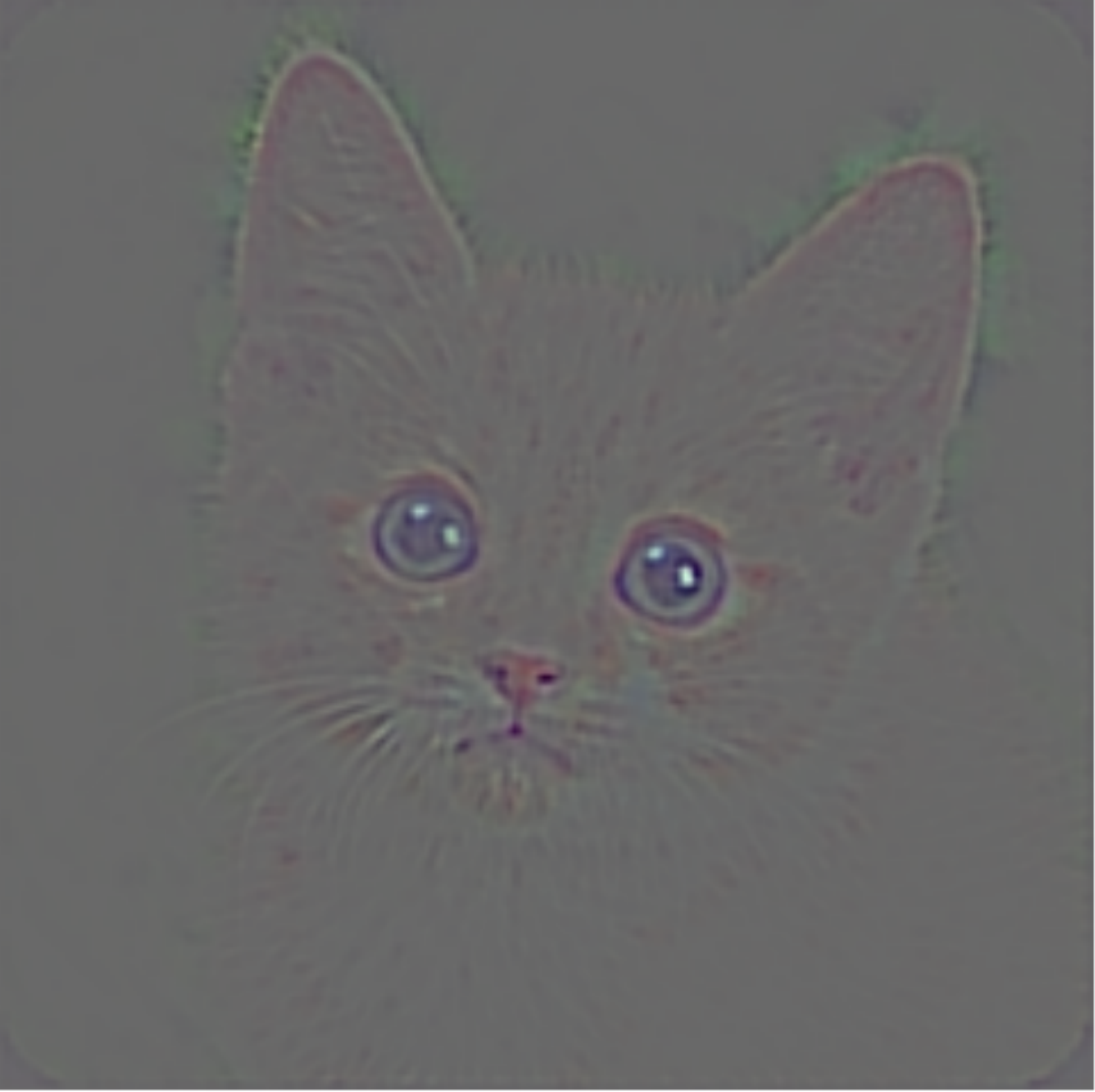}
		\caption*{\footnotesize GBP-max}
	\end{subfigure}
	\centering
	\begin{subfigure}[b]{0.27\linewidth}
		\centering
		\includegraphics[width=\linewidth]{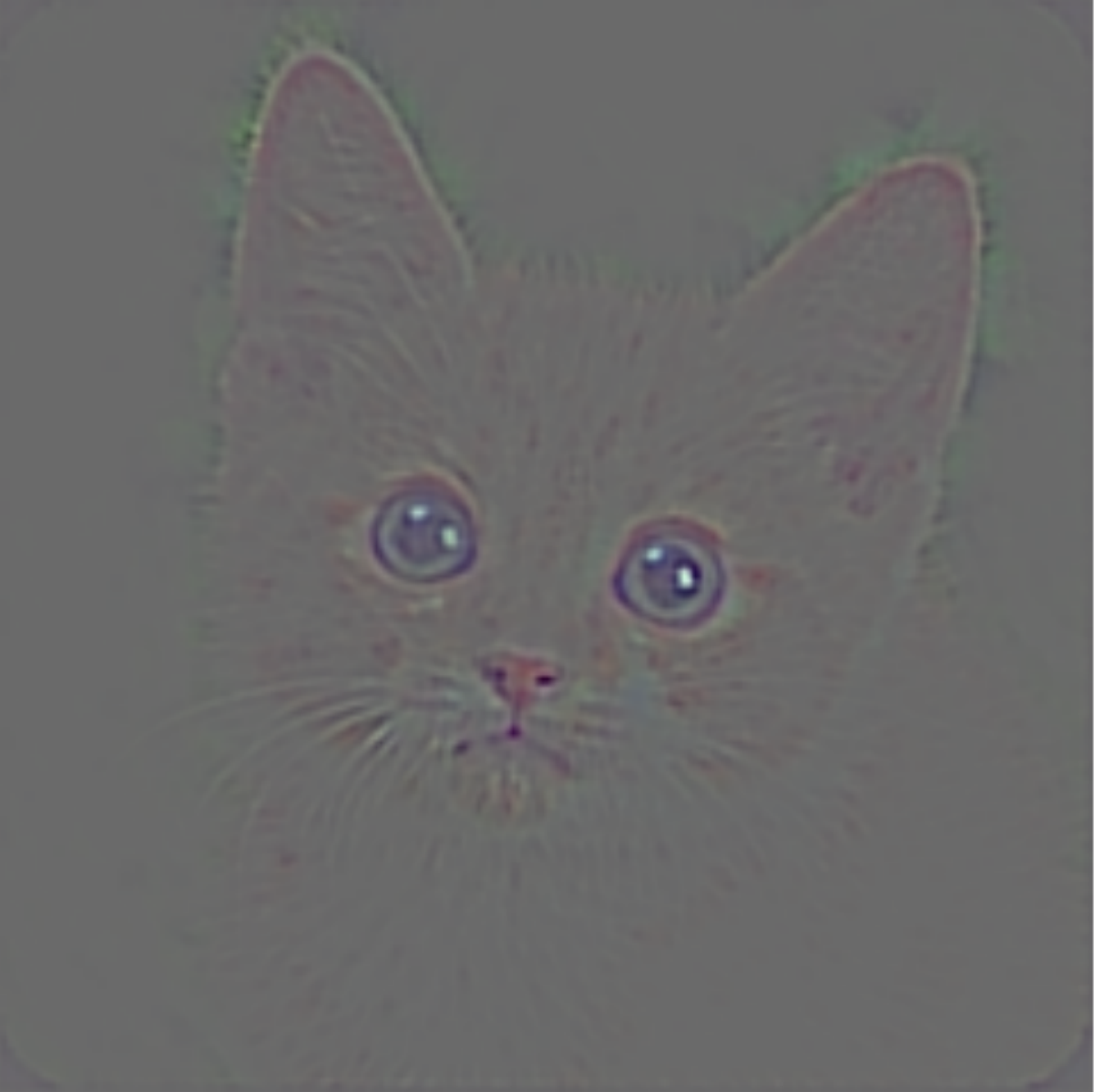}
		\caption*{\footnotesize GBP-482}
	\end{subfigure}
	\centering
	\begin{subfigure}[b]{0.27\linewidth}
		\centering
		\includegraphics[width=\linewidth]{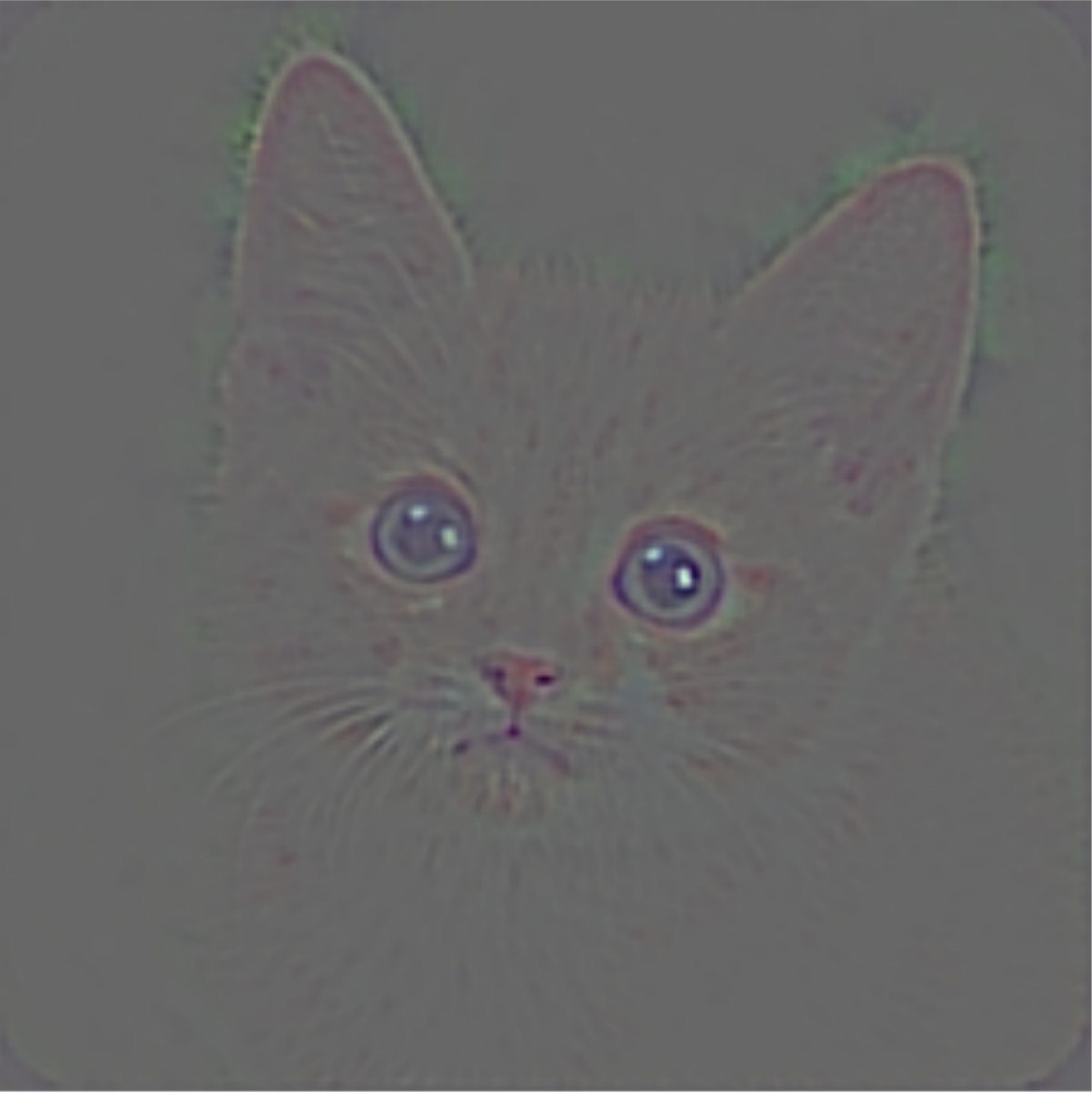}
		\caption*{\footnotesize GBP-560}
	\end{subfigure}

	\end{minipage}
	\caption{Backpropagation-based visualizations for the trained VGG-16 net given an input ``tabby''. From top row to the last row, it is saliency map, DeconvNet and GBP, where ``max'' refers to computing the (modified) gradient for the maximum class logit and the number, say ``482'', refers to computing the (modified) gradient for the $482$-th logit.  These numbers are randomly chosen for generality. Best viewed in the electronic version.}\label{pre_vgg_vis}
\end{figure}

To be a good visualization method, a clean and visually human-interpretable result is very desirable. More importantly, it should also reveal how the neural networks make decisions. Based on this, we provide the empirical behaviors of the backpropagation-based visualizations for a pre-trained VGG-16 net \cite{simonyan2014very} in Figure \ref{pre_vgg_vis}. Without loss of generality, the visualizations are obtained by choosing one of the class logits (i.e. the unnormalized class probability output right before the softmax function) as the output score to be taken derivative with respect to the input image.

For the visual quality, saliency map is very noisy while DeconvNet and GBP produce human-interpretable visualizations with a subtle difference: DeconvNet unexpectedly produces some kind of texture-like pattern, and GBP is cleaner with some background information filtered out. 
For the class-sensitivity,
saliency map changes greatly for different class logits while DeconvNet and GBP are almost invariant to which class logit we choose. 
This, together with more experiments, suggests that after introducing the backward ReLU, both DeconvNet and GBP modify the true gradient in a way that they create much cleaner results but their functionality as an indicator of important pixels to a specific class has disappeared. 
In the next section, we will explain these empirical behaviors and discuss the reason why GBP and DeconvNet differ greatly from saliency map.

\section{Theoretical Explanations} \label{sec: theory}
\label{explain}

We first analyze the backpropagation-based methods in a three-layer CNN with random Gaussian weights, which is then extended to more complicated models such as CNNs with max-pooling and deep CNNs. Besides, we also investigate their behaviors in well-trained CNNs. 

\subsection{A Random Three-Layer CNN}
\label{three_layer_cnn}

Consider a three-layer CNN, consisting of an input layer and a convolutional hidden layer, followed by a ReLU activation function and a fully connected layer of which its output is called class logits. Formally, let $x \in \mathbb{R}^d$ be a normalized input image with dimension $d$ and $\|x\|=1$, and let $W \in \mathbb{R}^{p \times N}$  be $N$ convolutional filters where each column $w^{(i)}$ denotes the $i$-th filter with size $p$. 
Note that here we use vectors to represent images and filters for simplicity, and the analysis also works for the more practical two-dimensional case. 
Then, we let $Y \in \mathbb{R}^{p \times J}$ be $J$ image patches extracted from $x$, and each column $y^{(j)}$ with size $p$ is generated by a linear function $y^{(j)} = {D_j} x$ where $D_j \triangleq \begin{bmatrix}
0_{p \times (j-1)b} & I_{p \times p} & 0_{p \times (d-(j-1)b-p)}
\end{bmatrix}$ with $b$ being the stride size\footnote{Here we assume a VALID padding method implicitly, and other padding methods do not impact our analysis.}. 
For example, given a filter with size $3$ and stride $1$, the resulting $j$-th patch $y^{(j)}$ is made of the $j$-th to $(j+2)$-th consecutive pixels. 
The weights in the fully-connected layer can be represented by $V \in \mathbb{R}^{NJ \times K}$ with $K$ being the number of output logits.
Therefore, the $k$-th logit is represented by
\begin{align} \label{conv_formula}
f_k(x) = \sum_{i=1}^{N}\sum_{j=1}^{J}{{V_{q_{ij},k}}\sigma({w^{(i)T}} y^{(j)})}
\end{align}
where the index $q_{ij}$ denotes the $((i-1)J+j)$-th entry in every column vector of weight matrix $V$. 

Assume every entry of $V$ and $W$ is sampled from an \textit{i.i.d.} Gaussian distribution $\mathcal{N}(0, c^2)$.
The following lemma provides the formula for backpropagation-based visualizations in a random three-layer CNN. 
Note that the norm of the final results will be in the range of $[0, 1]$ as we apply the normalization during visualizations.

\begin{lemma} \label{lem1}
    The backpropagation-based visualizations for the $k$-th logit in a random three-layer CNN is formalized as 
    \begin{align} \label{der_lem}
        s_k(x) = \frac{1}{Z_k} \sum_{j=1}^{J} {D_j}^T \sum_{i=1}^{N} {h({V_{q_{ij},k}}) \tilde{w}^{(i,j)}}
    \end{align}
    where $Z_k$ is the normalization coefficient
    to ensure $\|s_k(x)\| \in [0,1]$, $h(\cdot)$ is given by Eq. (\ref{two_funcs}) and 
    \begin{align*} 
        \tilde{w}^{(i,j)} = \begin{cases}
        w^{(i)} & \text{for DeconvNet}\\
        w^{(i)}\mathbb{I}\left({w^{(i)T}}y^{(j)}\right) & \text{for saliency map and GBP} 
        \end{cases}
    \end{align*}
\end{lemma}
\textit{Proof.} See Appendix A. \hfill $\square$

Next, we can analyze the different behaviors of these backpropagation-based methods case by case.

\subsubsection{Guided Backpropagation}

First, the behavior of GBP is given as follows.

\begin{theorem} \label{thm1}
In a random three-layer CNN, if the number of filters $N$ is sufficiently large, GBP at the $k$-th logit can be approximated as 
\begin{align} \label{GBP_final_thm}
	s^{\text{GBP}}_k(x) \approx  x
\end{align}
\end{theorem}


\textit{Proof.} See Appendix B. \hfill $\square$

The above theorem shows that after introducing the backward ReLU, the input image can be \textit{approximately recovered} by GBP in a random three-layer CNN, regardless of the class label.
However, according to the linear model explanation, backpropagation-based methods are visualizing learned weights, which should be random noise as they are all sampled from \textit{i.i.d} Gaussians. Obviously, it is inconsistent with the actual behavior of GBP.

As the approximation in Eq. (\ref{GBP_final_thm}) builds on an assumption that the number of filters $N$ is sufficiently large, a key question is: How many filters are needed to guarantee an accurate recovery? 
From \cite{lugosi2017sub}, we can set $N = \tilde{O} (\frac{p}{\epsilon^2})$ such that with high probability $\| \frac{1}{N} \sum_{i=1}^{N} { \tilde{w}^{(i,j)}} - \mathbb{E} [ { \tilde{w}^{(i,j)}} ] \| <\epsilon$, where $p$ denotes the filter size and $\tilde{O}(\cdot)$ hides some other factors. 
As an upper bound, it reveals that the number of convolutional filters needed heavily depends on the filter size $p$. As the filter size intrinsically determined by the local connections in CNNs is usually small, we could use a mild number of convolutional filters to recover the input image. For example, given a filter size $3 \times 3 \times 3$, we  need at most $O(10^3)$ filters to achieve an estimation error $\epsilon$ less than $0.1$. This strongly suggests that GBP visualizations are human-interpretable in most of the CNNs, and thus the \textit{local connections} property is another key factor underlying crisp visualizations.

\subsubsection{Saliency Map and DeconvNet}

Here we show the behaviors of saliency map and DeconvNet in a random three-layer CNN are largely different from GBP.

\begin{theorem}
    In a random three-layer CNN, if the number of filters $N$ is sufficiently large, saliency map and DeconvNet are approximated as Gaussian random variables satisfying 
    \begin{align*}
        s^{\text{Sal}}_k(x), s^{\text{Deconv}}_k(x) \sim \mathcal{N} (0, I)
    \end{align*}
\end{theorem}
\textit{Proof.} See Appendix C. \hfill $\square$

The above theorem shows that both saliency map and DeconvNet visualizations will yield \textit{random noise}, conveying little information about the input image and class logits. For saliency map, it is easily understood since saliency map represents the true gradient of the class logit, which heavily depends on the weights. For DeconvNet, although its behavior appears similar to saliency map in this simplistic scenario, we will show later on that it behaves more similarly to GBP, in particular with the existence of max-pooling.

\subsection{Extensions to More Realistic Models}

In this section, we extend our analysis of a simple random three-layer CNN to other more realistic cases, including the max-pooling, deeper nets and trained weights.

\subsubsection{CNNs with Max-Pooling} \label{theory_maxpool}

If we add a max-pooling layer between the ReLU and the fully-connected layer, the $k$-th logit becomes
\begin{align*} 
f_k(x) = \sum_{i=1}^{N}\sum_{j=1}^{J}{{V_{\tilde{q}_{ij},k}}\delta(\sigma({w^{(i)T}} y^{(j)}))}
\end{align*}
where $\delta(\cdot)$ denotes the max-pooling, which successively selects the maximum value in a fixed-size pooling window, and the new index $\tilde{q}_{ij}$ is the down-sampled version of $q_{ij}$.
Then the backpropagation-based visualizations for the $k$-th logit can be formulated as 
\begin{align} \label{maxpool_der}
s_k(x) = \frac{1}{Z_k} \sum_{j=1}^{J} {D_j}^T \sum_{i=1}^{N} {h(\delta'(o_{ij}){V_{\tilde{q}_{ij},k}}) \tilde{w}^{(i,j)}}
\end{align}
where $o_{ij} \triangleq \sigma({w^{(i)T}} y^{(j)})$ is the output of each ReLU activation and $\delta'(o_{ij})$ denotes the derivative of $\delta(\cdot)$ evaluated at $o_{ij}$, which is
\begin{align*}
    \delta'(o_{ij}) = 
    \begin{cases}
    1 & \text{if $o_{ij}$ is chosen by max-pooling}\\
    0 & \text{otherwise}
    \end{cases} 
\end{align*}
Since $o_{ij} \geq 0$ with equality holds for ${w^{(i)T}}y^{(j)} \leq 0$, given a proper pooling window size, it is highly possible that $o_{ij}$ is chosen by the max-pooling if and only if ${w^{(i)T}}y^{(j)} > 0$. It means with high probability, Eq. (\ref{maxpool_der}) is approximated as
\begin{align} \label{maxpool_der2}
s_k(x) \approx \frac{1}{Z_k} \sum_{j=1}^{J} {D_j}^T \sum_{i=1}^{N} {h({V_{\tilde{q}_{ij},k}}) \tilde{w}^{(i,j)}} \mathbb{I}({w^{(i)T}}y^{(j)})
\end{align}
For saliency map and GBP, we know $\tilde{w}^{(i,j)} \mathbb{I}({w^{(i)T}}y^{(j)}) = \tilde{w}^{(i,j)}$ and thus Eq. (\ref{maxpool_der2}) is further reduced to Eq. (\ref{der_lem}), which means
the behaviors of saliency map and GBP remain the same after introducing the max-pooling.
However, with high probability, DeconvNet at the $k$-th logit becomes
\begin{align*}
s^{\text{Deconv}}_k(x) \approx \frac{1}{Z_k} \sum_{j=1}^{J} {D_j}^T \sum_{i=1}^{N} {\sigma({V_{\tilde{q}_{ij},k}}) w^{(i)}} \mathbb{I}({w^{(i)T}}y^{(j)})
\end{align*}
which is exactly the form of GBP in Eq. (\ref{der_lem}). 
Therefore, adding the max-pooling makes the DeconvNet behave like GBP -- doing nothing but image recovery. This also explains and extends the previous intuitive claims in \cite{samek2017evaluating, odena2016deconvolution} that the image-specific information in DeconvNet comes from the max-pooling.

Note that that the approximation from Eq. (\ref{maxpool_der}) to Eq. (\ref{maxpool_der2}) in DeconvNet with the max-pooling is essentially different from the approximations used in GBP. For GBP, the approximate gap can be made arbitrarily small by increasing the hidden layer size $N$, leading to a perfect recovery of the input. However, for DeconvNet, given any pooling window size, there might always exist at least one of the following two contradictory cases: 
it is possible that $a_{ij}$ is chosen  by the max-pooling if ${w^{(i)T}}y^{(j)} \leq 0$, and also possible that $a_{ij}$ is not chosen if ${w^{(i)T}}y^{(j)} > 0$.
This makes DeconvNet  (with max-pooling), in theory, never recover input perfectly, which might explain why the unusual texture-like artifacts appear in the DeconvNet visualizations.





\subsubsection{Deep CNNs}

The analysis for a three-layer CNN can be  generalized to the multi-layer (or deeper) case. For clarity, we formulate the $k$-th logit of an $L$-layer deep CNN in a matrix form:
\begin{align*}
f_k(x) = {\Gamma^{(L)T}_k} \sigma \left({\Gamma^{(L-1)T}} \cdots \sigma \left( {\Gamma^{(1)T}} x \right)\right)
\end{align*}
where $\Gamma^{(l)} \in \mathbb{R}^{d_{l} \times d_{l+1}}$ denotes either the convolutional or fully-connected operator matrix in the $l$-th layer and $\Gamma^{(L)}_k$ is the $k$-th column  of $\Gamma^{(L)}$. Denote by $o^{(l)}$ the output of ReLU activations in the $l$-th layer, i.e. 
$o^{(l)} = \sigma\left( {\Gamma^{(l)T}} o^{(l-1)} \right), \; \forall l \in \{1, \cdots, L-1\}$ with $o^{(0)} \triangleq x$.
Then backpropagation-based visualizations at the $k$-th logit in an $L$-layer deep CNN can be formulated as
\begin{align} \label{sal_deep}
\begin{split}
    s_k(x) &= \frac{1}{Z_k}\frac{\partial \tilde{o}^{(1)}}{\partial x} \cdot h(\hat{V}_{\cdot,k}^{(1)})\\
    & \mathop = \limits^{(a)} \frac{1}{Z_k} \sum_{j=1}^{J} {D_j}^T \sum_{i=1}^{N} {h({\hat{V}_{q_{ij},k}^{(1)}}) \tilde{w}^{(i,j)}}
\end{split}
\end{align}
with $\forall l \in \{1, \cdots, L-1\}$,
\begin{align*} 
\begin{split}
    \hat{V}_{\cdot,k}^{(l)} = \frac{\partial {\tilde{o}}^{(l+1)}}{\partial o^{(l)}} \cdot h\left(\frac{\partial {\tilde{o}}^{(l+2)}}{\partial o^{(l+1)}}  \cdots  h\left( \frac{\partial {\tilde{o}}^{(L-1)}}{\partial o^{(L-2)}} h\left(\Gamma^{(L)}_k \right) \right) \right)
\end{split}
\end{align*}
where in $(a)$ we rewrite $s_k(x)$ in an expanded form,
${\tilde{o}}^{(l)} \triangleq g\left( {\Gamma^{(l)T}} o^{(l-1)} \right)$, $w^{(i)}$ is the $i$-th filter encoded in $\Gamma^{(1)}$ and $N$ is the number of filters in the first convolutional layer.
Also, $h(\cdot)$, $g(\cdot)$ and ${\tilde{w}}^{(i,j)}$ are defined in Eq. (\ref{two_funcs}) and Lemma \ref{lem1}. 

First, 
the approximate property of $\hat{V}_{\cdot, k}^{(1)}$ in the random deep CNN is given in the following proposition.
\begin{proposition} \label{prop1}
    For a random deep CNN where weights are \textit{i.i.d.} Gaussians with zero mean, we can also approximate every entry of $\hat{V}_{\cdot, k}^{(1)}$ as \textit{i.i.d.} Gaussian with zero mean. 
\end{proposition}


\textit{Proof.} See Appendix D. \hfill $\square$

Based on Proposition \ref{prop1}, we can see that the statistical properties of $\hat{V}_{q_{ij}, k}^{(1)}$ in Eq. (\ref{sal_deep}) are approximately the same with those of $V_{q_{ij},k}$ in Eq. (\ref{der_lem}), which means the analysis of backpropagation-based visualizations in a shallow three-layer CNN also applies to the deep CNN case. Therefore, the behaviors of these visualizations will barely change when increasing the depth of neural networks.


\begin{figure} [!t]
    \centering
    \begin{subfigure}[b]{0.25\textwidth}
		\centering
		\includegraphics[width=0.9\textwidth]{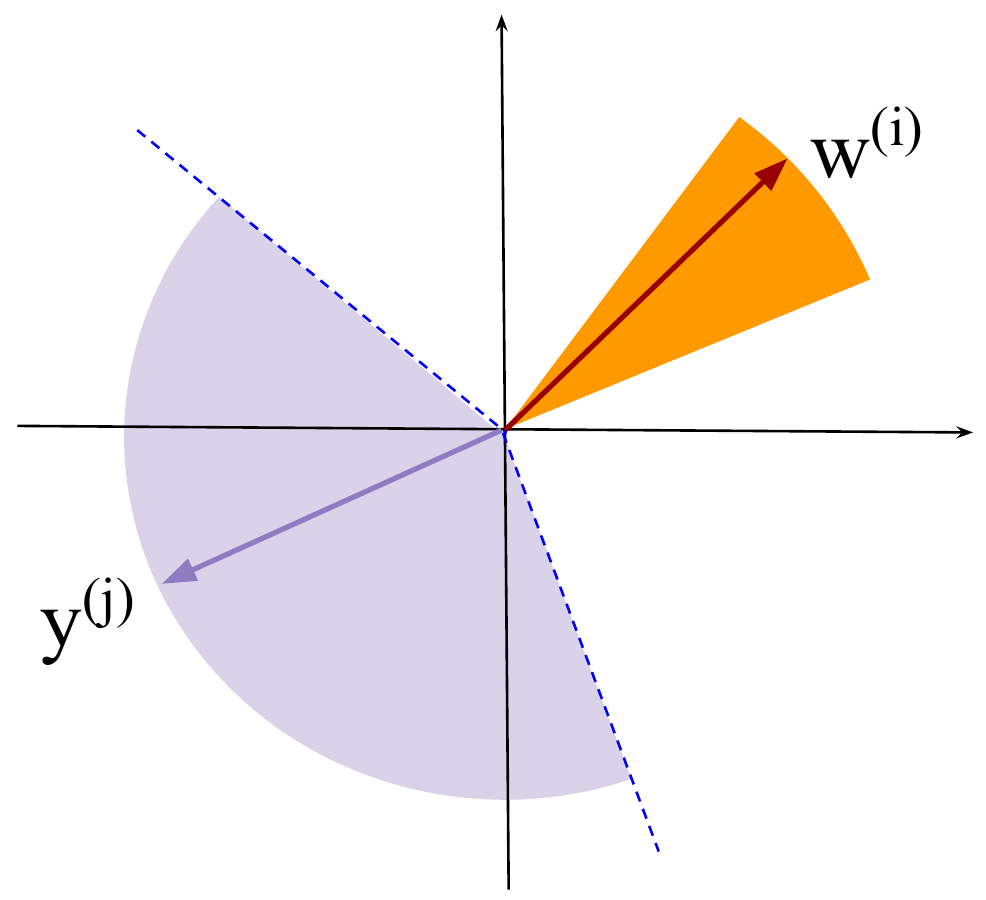}
		\caption{\footnotesize A toy example}
	\end{subfigure}
	
	\begin{subfigure}[b]{0.23\textwidth}
		\centering
		\includegraphics[width=\textwidth]{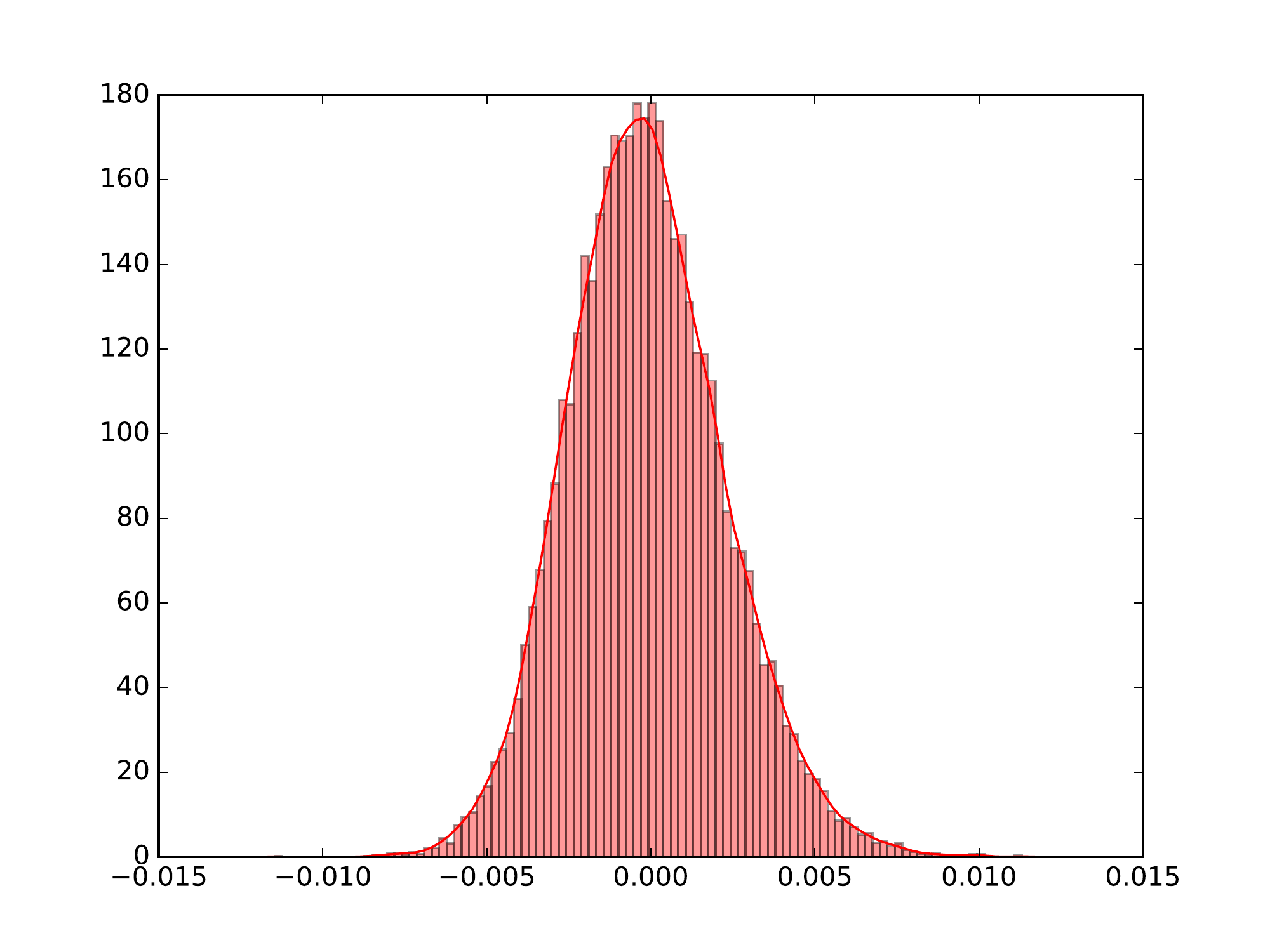}
		\caption{\footnotesize neuron-26 in trained fc1}
	\end{subfigure}
	\begin{subfigure}[b]{0.23\textwidth}
		\centering
		\includegraphics[width=\textwidth]{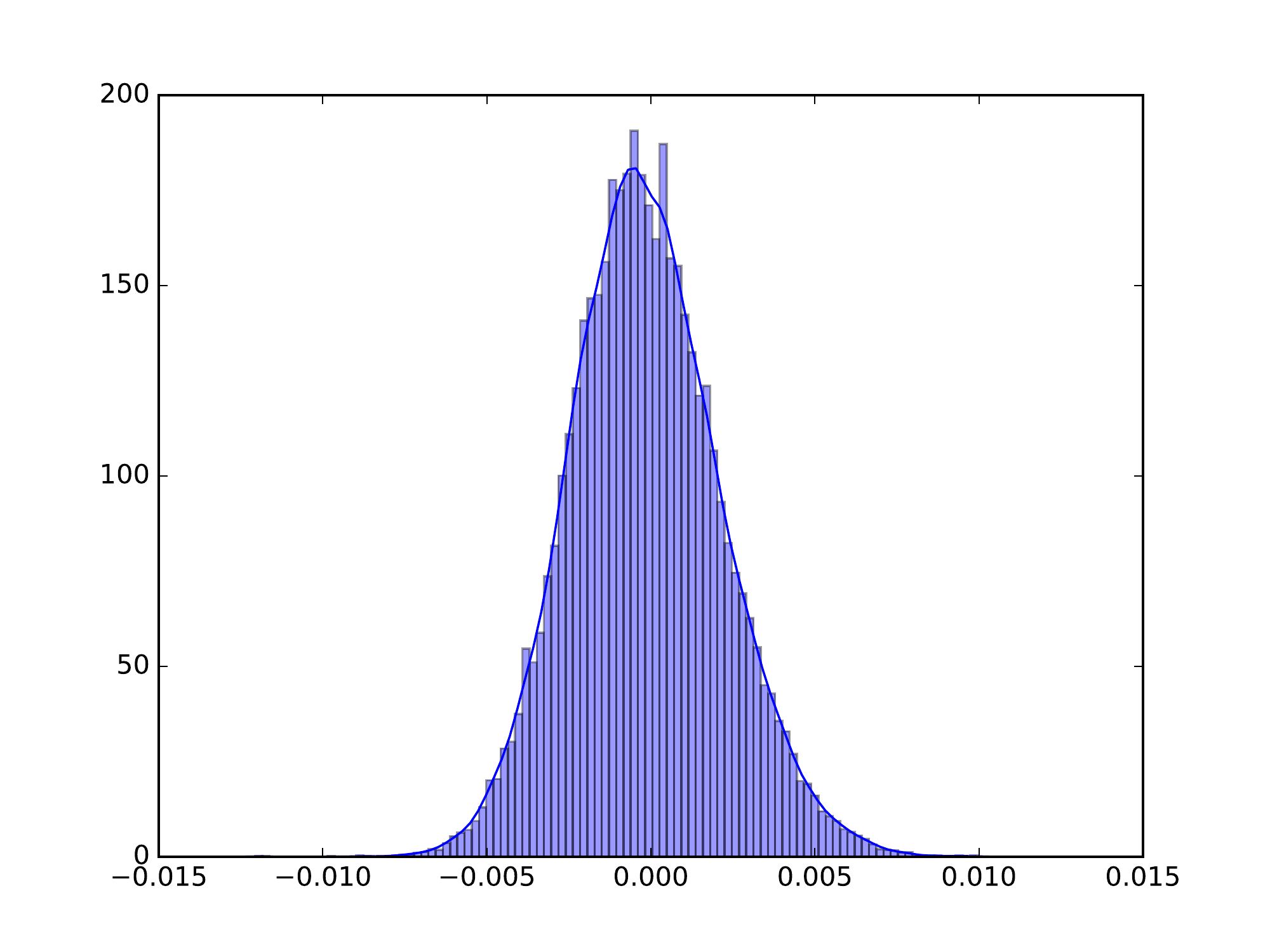}
		\caption{\footnotesize neuron-44 in trained fc1}
	\end{subfigure}
	
	\caption{ (a) shows a two-dimensional toy example where $w^{(i)}$'s are all in a cone (the orange area) and all the $y^{(j)}$'s in another cone (the grey area) called ``dead zone'' will be filtered out by the ReLU. (b) and (c) show the histograms of all weights connected to the $26$-th activation and the $44$-th one, respectively, in the layer ``fc1'' of the trained VGG-16 net. Note that we randomly picked up two activations (i.e. 26 and 44 here) for comparison. }
	\label{weight_trained}
\end{figure}

\subsubsection{CNNs with Trained Weights}

The previous analysis for random CNNs does not apply to the trained case directly
since the weights 
here may not be \textit{i.i.d.} Gaussian distributed. 
For saliency map, which uses the true gradient, the trained weights are likely to impose a stronger bias towards some specific subset of the input pixels, and so they can highlight class-relevant pixels rather than producing random noise. For GBP and DeconvNet, the analysis is a little more involved.

On the one hand,
the trained weights $w^{(i)}$ will only lie in a small subspace of the whole image patch space which will create some ``dead zones'', as illustrated in Figure \ref{weight_trained} (a). 
That is, all image patches lying in the ``dead zone'' will be filtered out by the forward ReLU. For example, it is well-known that the trained weights in the first convolutional layer are Gabor-like filters to detect the image patches containing edges \cite{yosinski2014transferable,zeiler2014visualizing}. That is, image patches without edges will probably be filtered out by the first convolutional layer. 
Also, the higher convolutional layers  keep filtering out more image patches with certain patterns (e.g. Figure \ref{part_load_1}). See the supplementary material for a comparison between GBP and a linear edge detector.


On the other hand, as shown in Figure \ref{weight_trained} (b) and (c), the histograms of weights connected to the respective one of any two different neurons in the first fully connected layer (called ``fc1'') of the trained VGG-16 net are very similar to each other. Approximately, they form two very similar Gaussians with a small standard deviation, which means the (modified) gradients at any two different neurons in the layer ``fc1'' with respect to the input image are almost the same. Namely, $\frac{\partial \tilde{o}^{(\text{fc1)}}}{\partial x}$ in Eq. (\ref{sal_deep})  for GBP and DeconvNet (with max-pooling) satisfies
$$\frac{\partial \tilde{o}^{(\text{fc1)}}_m}{\partial x} \approx F_{\text{conv}}(x), \forall m \in \{1,\cdots,M\}$$
where $\tilde{o}^{(\text{fc1)}}_m$ is the $m$-th entry of $\tilde{o}^{(\text{fc1)}}$ and $F_{\text{conv}}(\cdot): \mathbb{R}^d \to \mathbb{R}^d$ denotes the (normalized) overall filtering effect of the convolutional layers and $M$ is the number of neurons in the layer ``fc1''. 
Thus, Eq. (\ref{sal_deep}) for GBP and DeconvNet (with max-pooling) in the trained CNN can be approximated as
\begin{align} \label{der_trained}
    \begin{split}
        s_k(x) &= \frac{1}{Z_k }\frac{\partial \tilde{o}^{(\text{fc1})}}{\partial x} \cdot h(\hat{V}_{\cdot,k}^{(\text{fc1})}) \\
        &= \frac{1}{Z_k} \sum_{m=1}^{M} { \frac{\partial \tilde{o}_{m}^{(\text{fc1})}}{\partial x} \cdot h(\hat{V}_{m,k}^{(\text{fc1})}) } \\
        & \mathop \approx \limits^{(a)} F_{\text{conv}}(x)
    \end{split}
\end{align}
where $(a)$ follows from setting the normalization coefficient to be $Z_k = \frac{1}{\sum_{m=1}^M h(\hat{V}_{m,k}^{(\text{fc1})})}$.

It shows that GBP and DeconvNet (with max-pooling) in a trained CNN are actually doing the partial image recovery, where the trained weights control which image patch could form an \textit{active path} to the class logit. More importantly, this filtering process is not class sensitive (e.g. the edge detector). 
In the end, only these ``active'' image patches are combined in the first fully connected layer to form the final visualization results.
As the right side of (\ref{der_trained}) does not depend on $k$, it illustrates why the GBP and DecovNet visualizations in the trained VGG are not class-sensitive. 





\section{Experiments}
\label{experim}


To verify our theoretical analysis, we conduct a series of experiments on a three-layer CNN, a three-layer fully-connected network (FCN) and a VGG-16 net. For a random network, their weights are all sampled from the truncated Gaussians with a zero-mean and standard deviation $0.1$. Unless stated otherwise, the input is the image ``tabby'' from the ImageNet dataset \cite{deng2009imagenet} with size $224 \times 224 \times 3$. See the supplementary materials for more  results on other images and other neural network such as ResNet \cite{he2016deep}. In the three-layer CNN, the filter size is $7 \times 7 \times 3$, the number of filters is $N=256$, and the stride is $2$. In the three-layer FCN, the hidden layer size is set to $N_h=4096$. By default, the backpropagation-based visualizations are calculated with respect to the maximum class logit.

\begin{figure}[!t]
	\centering
		\centering
		\begin{subfigure}[b]{0.11\textwidth}
			\centering
			\includegraphics[width=\textwidth]{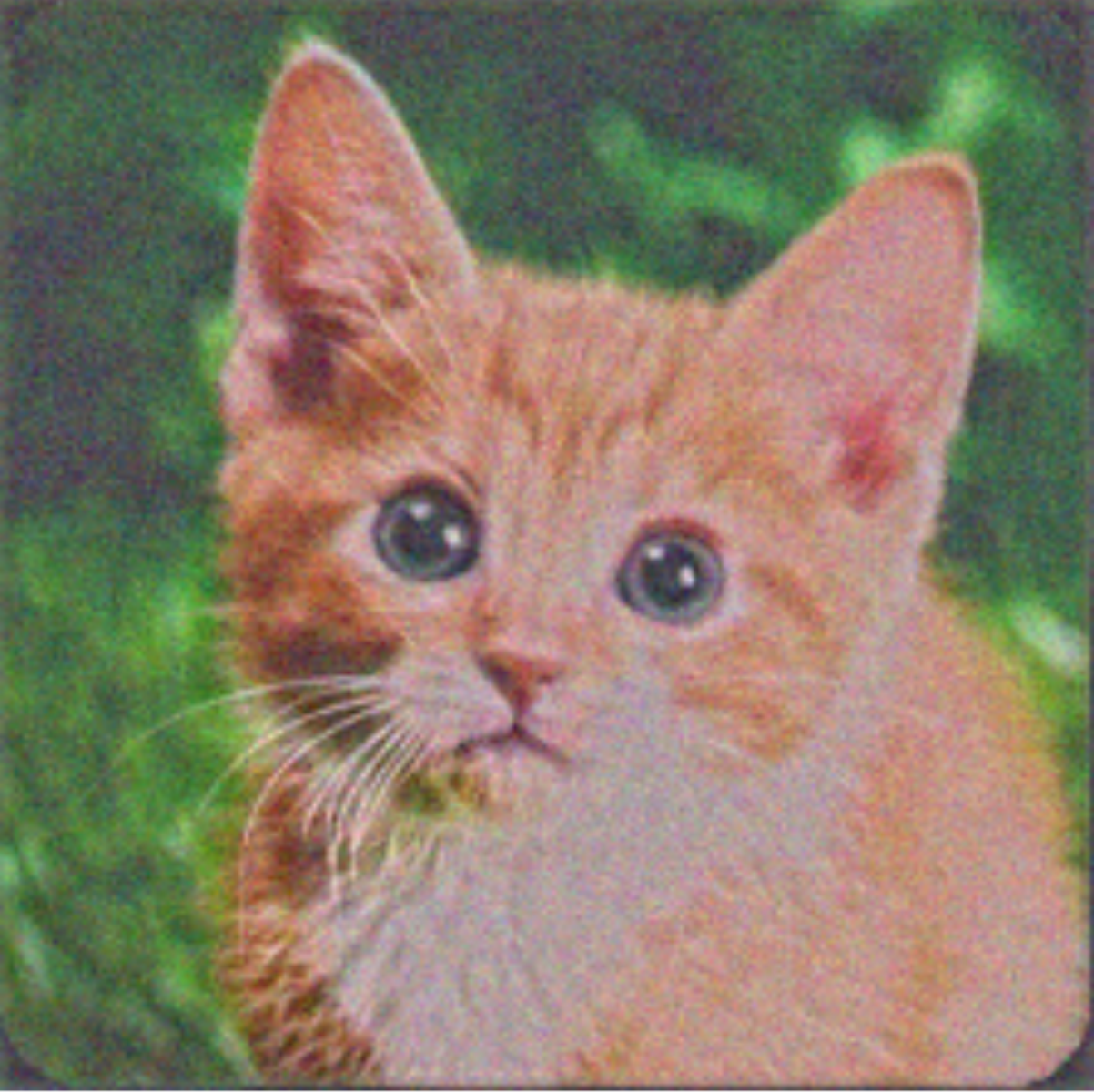}
			\caption*{\footnotesize GBP-CNN}
		\end{subfigure}
		\begin{subfigure}[b]{0.11\textwidth}
			\centering
			\includegraphics[width=\textwidth]{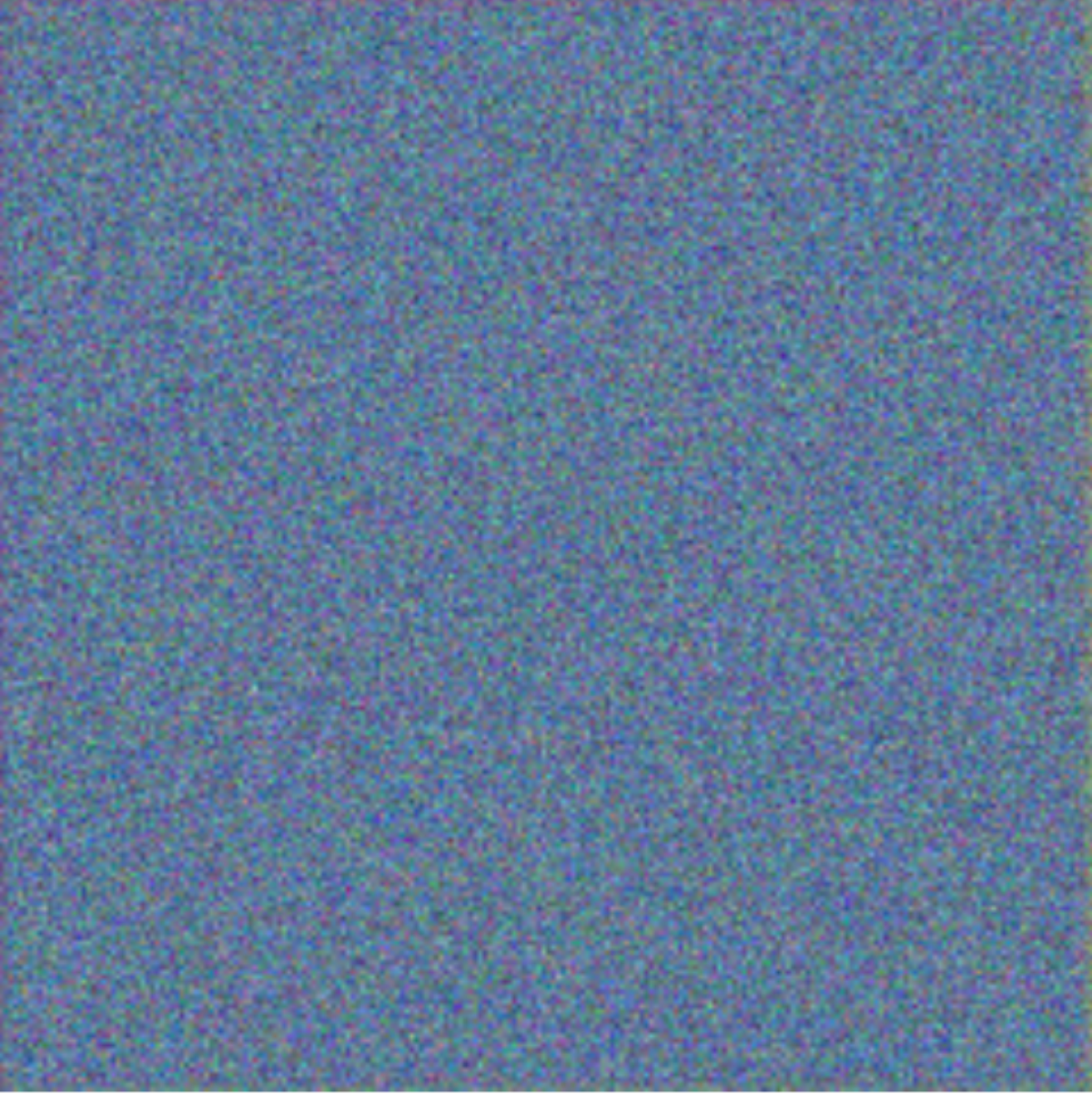}
			\caption*{\footnotesize Deconv-CNN}
		\end{subfigure}
		\begin{subfigure}[b]{0.11\textwidth}
			\centering
			\includegraphics[width=\textwidth]{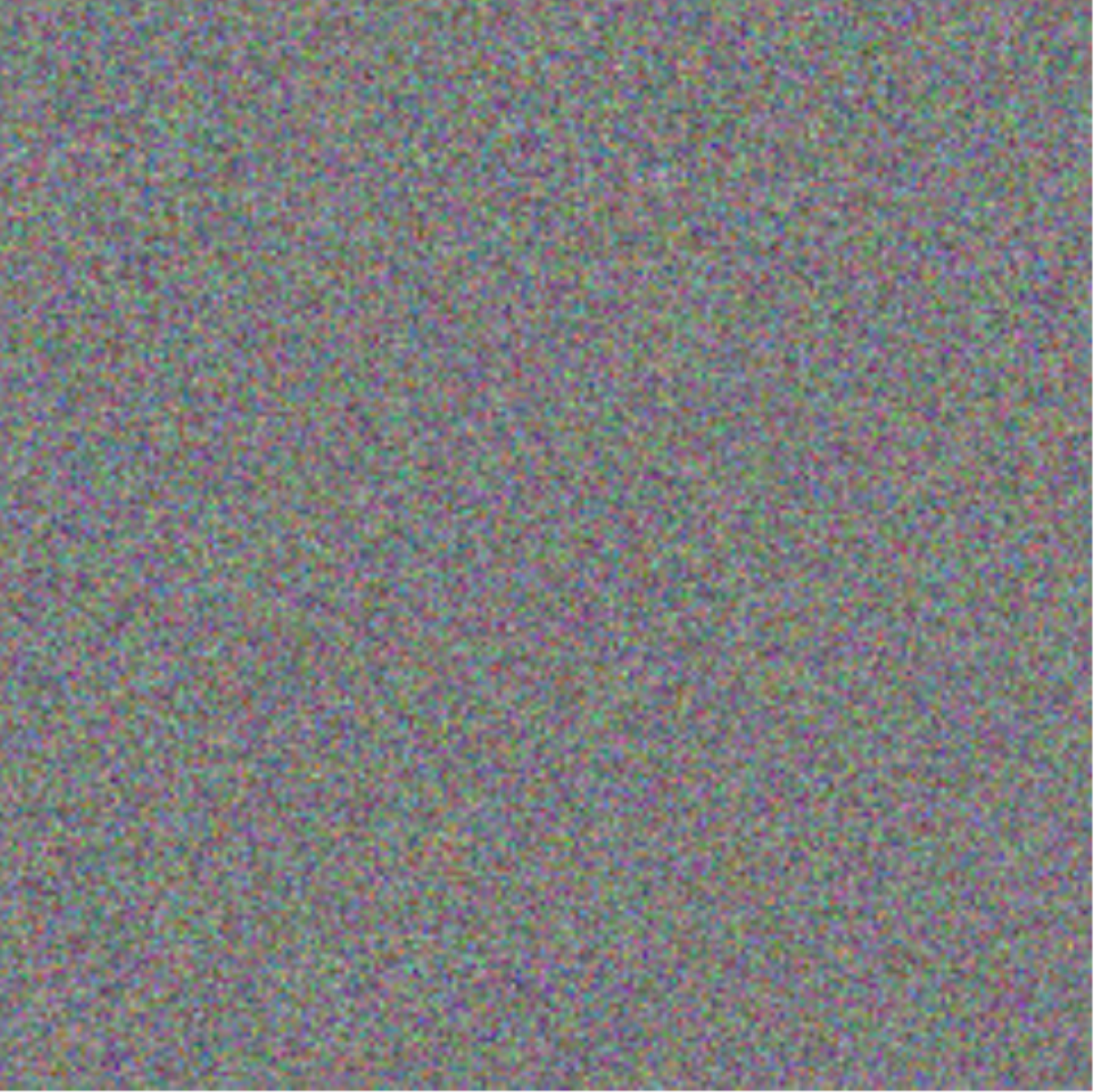}
			\caption*{\footnotesize Sal-CNN}
		\end{subfigure}
		
		\centering
		\begin{subfigure}[b]{0.11\textwidth}
			\centering
			\includegraphics[width=\textwidth]{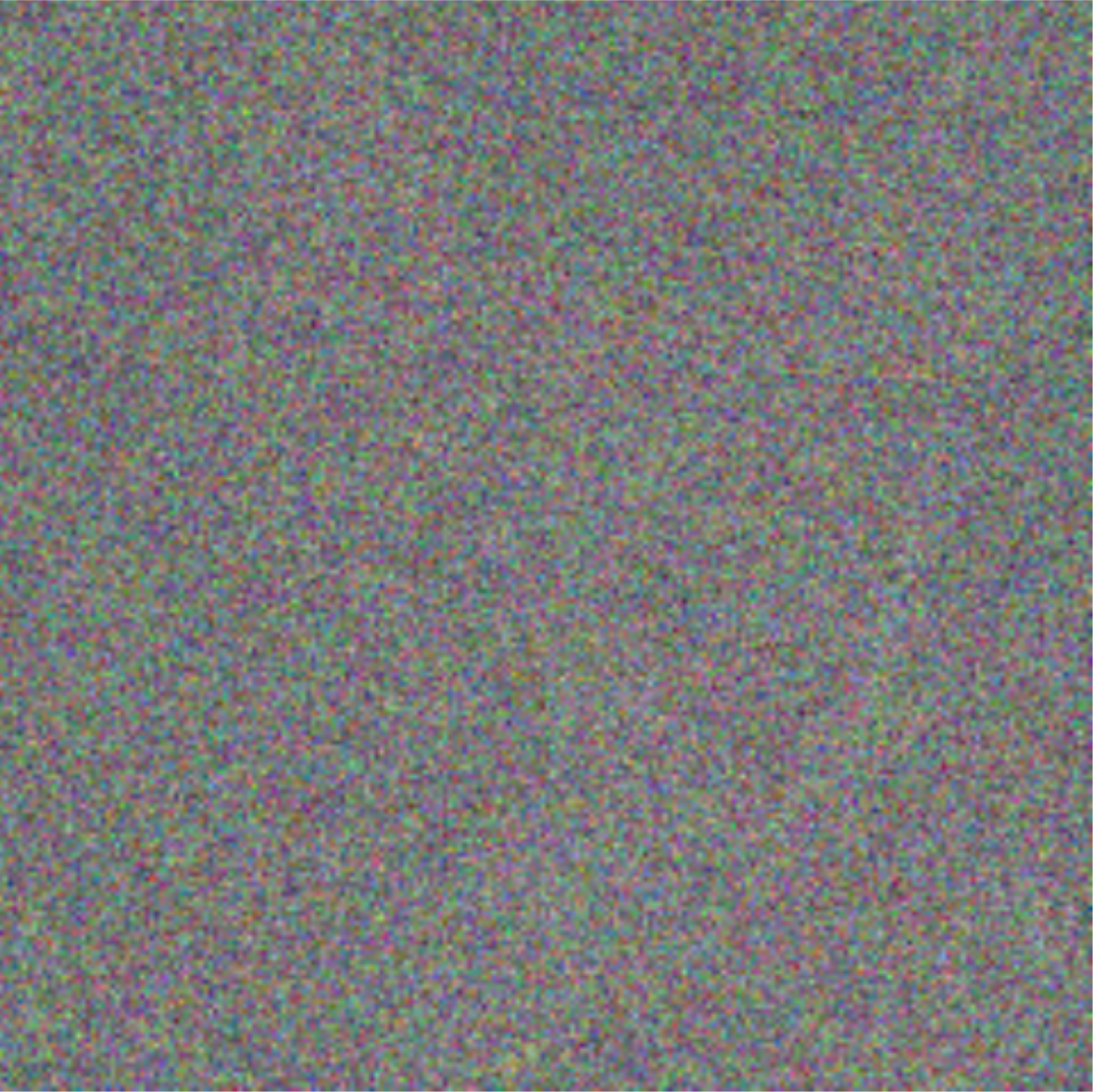}
			\caption*{\footnotesize GBP-FCN}
		\end{subfigure}
		\begin{subfigure}[b]{0.11\textwidth}
			\centering
			\includegraphics[width=\textwidth]{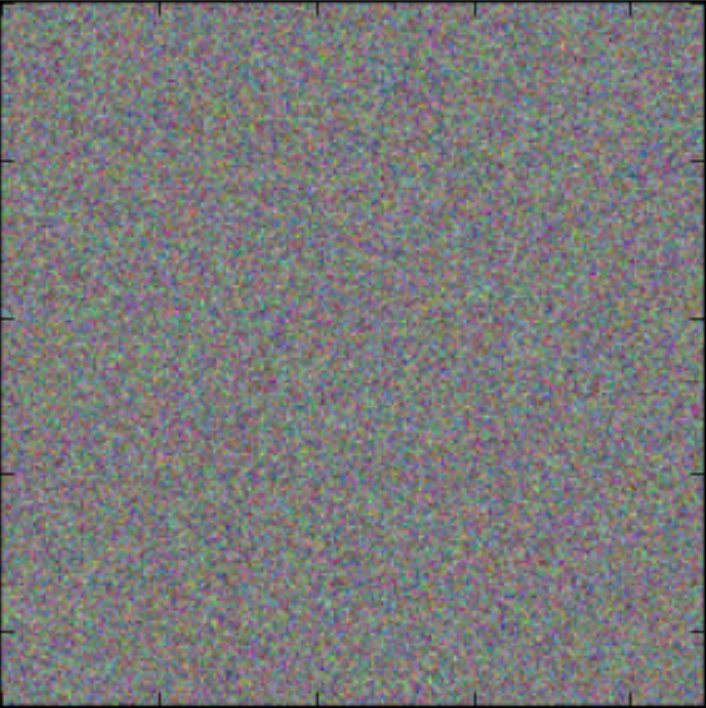}
			\caption*{\footnotesize Deconv-FCN}
		\end{subfigure}
		\begin{subfigure}[b]{0.11\textwidth}
			\centering
			\includegraphics[width=\textwidth]{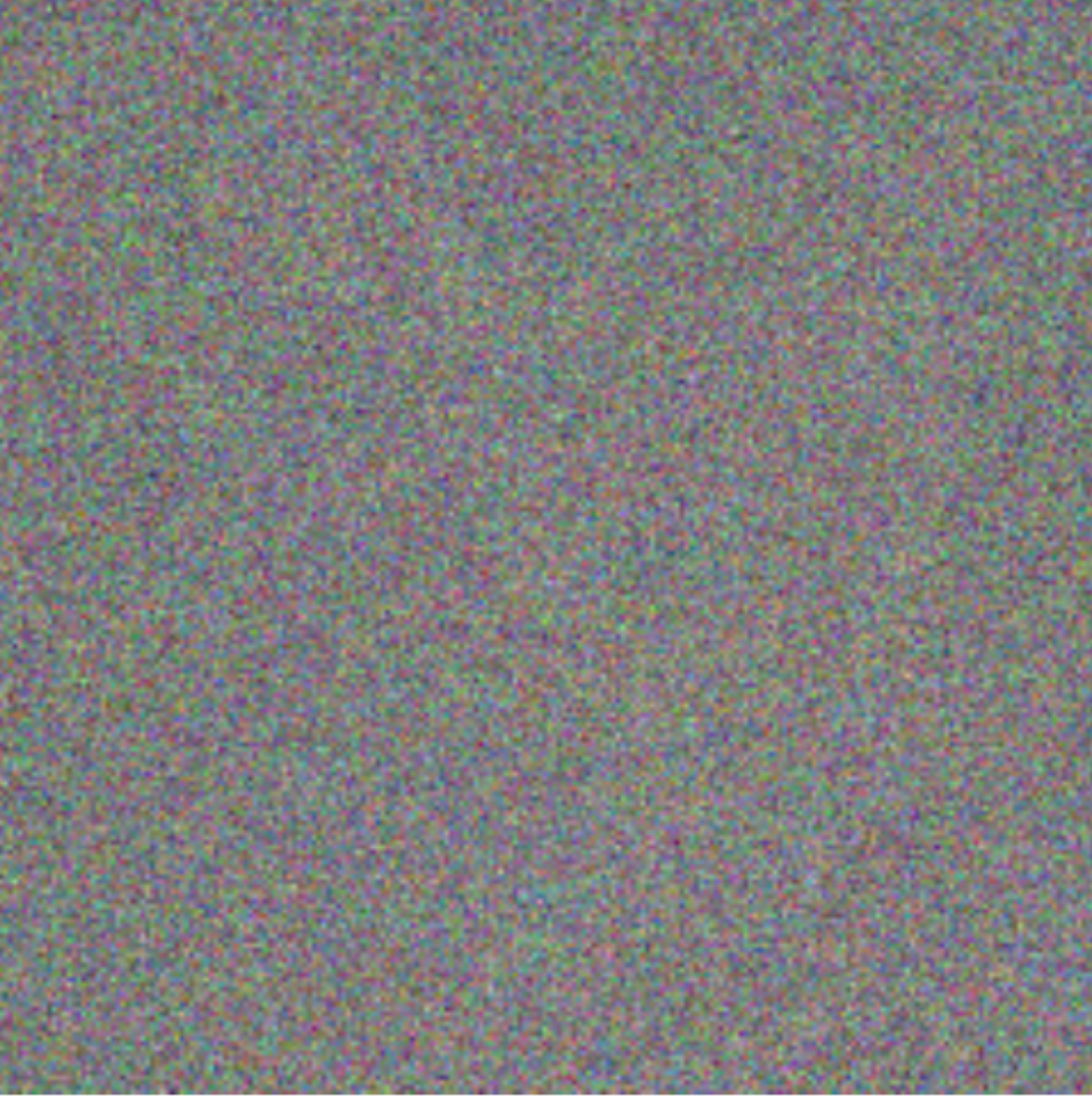}
			\caption*{\footnotesize Sal-FCN}
		\end{subfigure}
	\caption{Backpropagation-based visualizations in  a random three-layer CNN (top row) and a random three-layer FCN (bottom row) given the input image ``tabby''. From left to right, each column represents GBP, DeconvNet and saliency map, respectively. Only GBP visualization in the CNN is human-interpretable.}\label{verify_cnn_fcn}
\end{figure}

\subsection{Impact of Local Connections}

Figure \ref{verify_cnn_fcn} shows the backpropagation-based visualizations on a random three-layer CNN and a random three-layer FCN, respectively. We can see only GBP in the CNN can produce a human-interpretable visualization, while DeconvNet and saliency map in the CNN get random noise, which verifies our theoretical analysis in the section \ref{three_layer_cnn}. In contrast, as local connections do not exist in the FCN and the input size (e.g. $224\times 224\times3$) is extremely large, all the backpropagation-based methods (including GBP) in the FCN generate random noise. Particularly for GBP, the number of hidden neurons $N_h=4096$ is still not large enough to 
recover the image.

\begin{figure}[!t]
	\centering
	\begin{subfigure}[b]{0.11\textwidth}
		\centering
		\includegraphics[width=\textwidth]{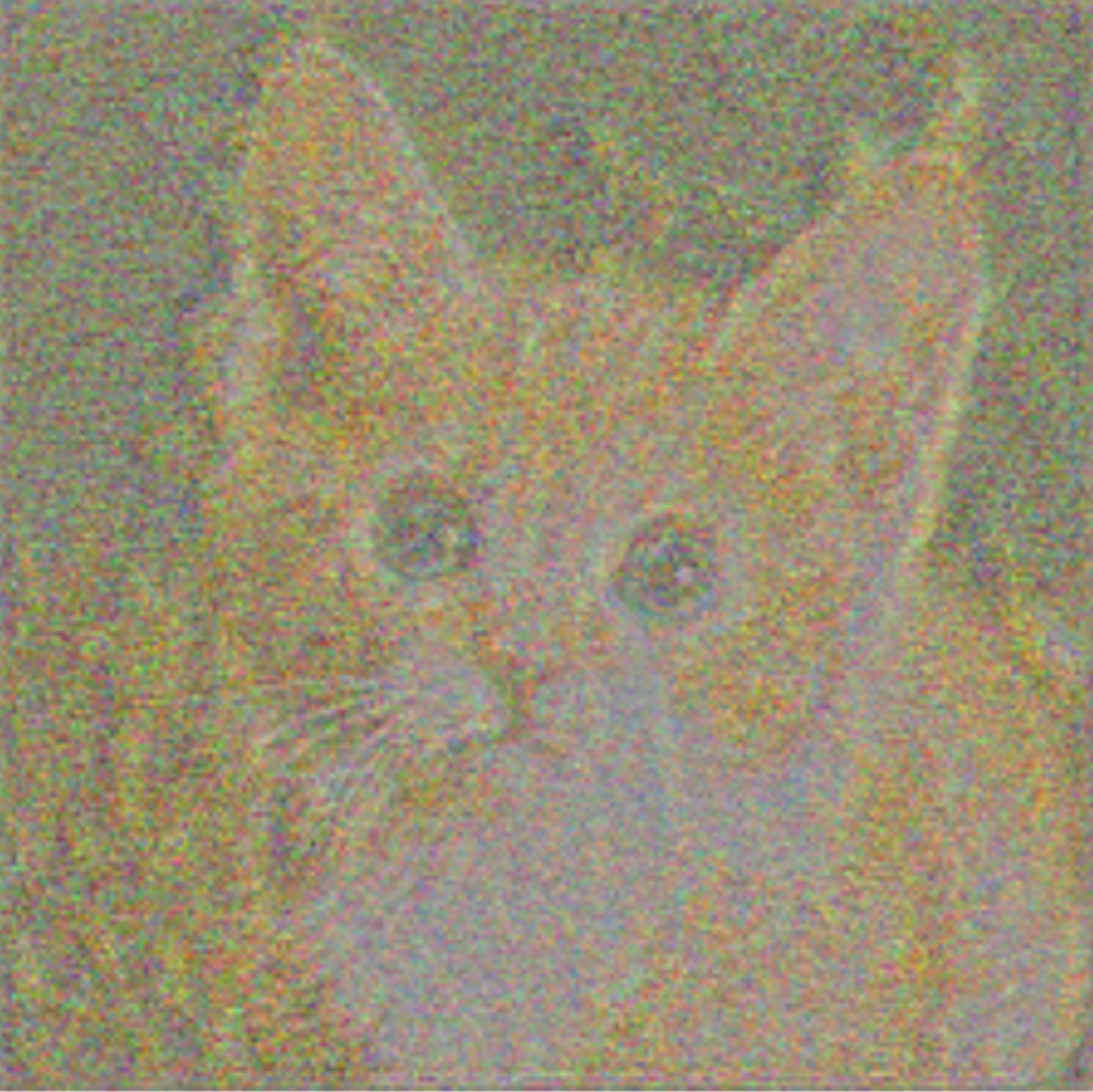}
		\caption*{\footnotesize $N=8$}
	\end{subfigure}
	\centering
	\begin{subfigure}[b]{0.11\textwidth}
		\centering
		\includegraphics[width=\textwidth]{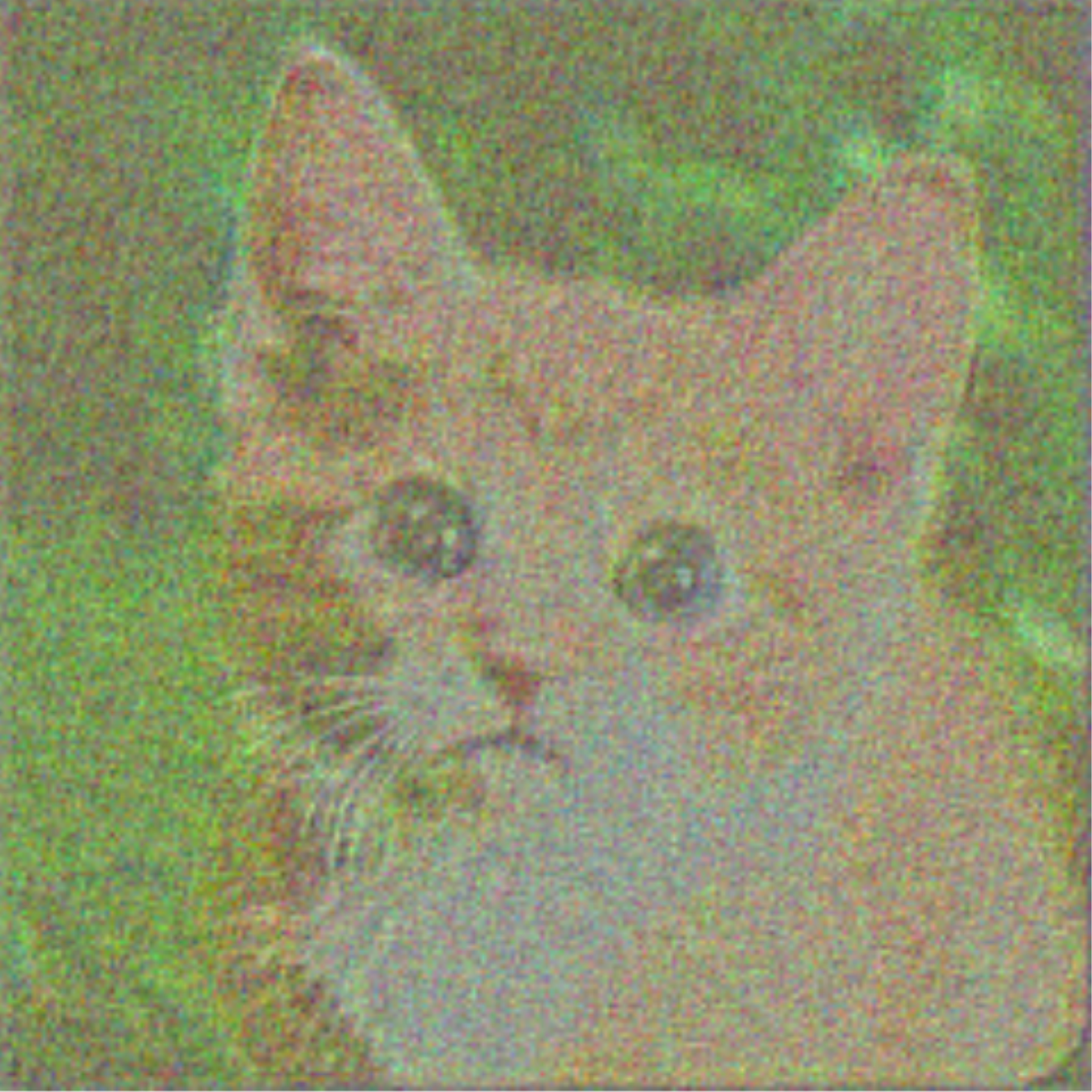}
		\caption*{\footnotesize $N=16$}
	\end{subfigure}
	\centering
	\begin{subfigure}[b]{0.11\textwidth}
		\centering
		\includegraphics[width=\textwidth]{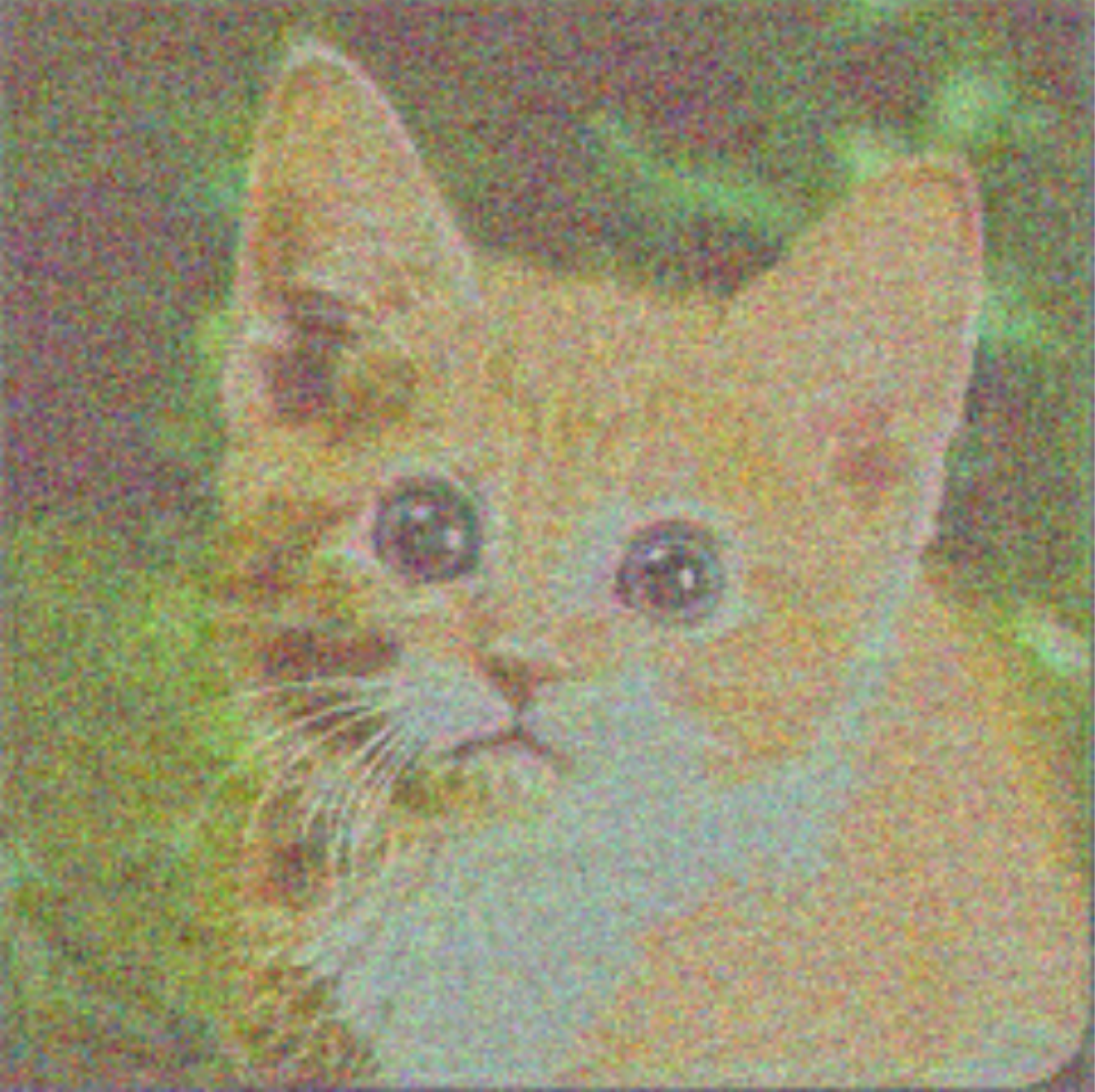}
		\caption*{\footnotesize $N=32$}
	\end{subfigure}
	\begin{subfigure}[b]{0.11\textwidth}
		\centering
		\includegraphics[width=\textwidth]{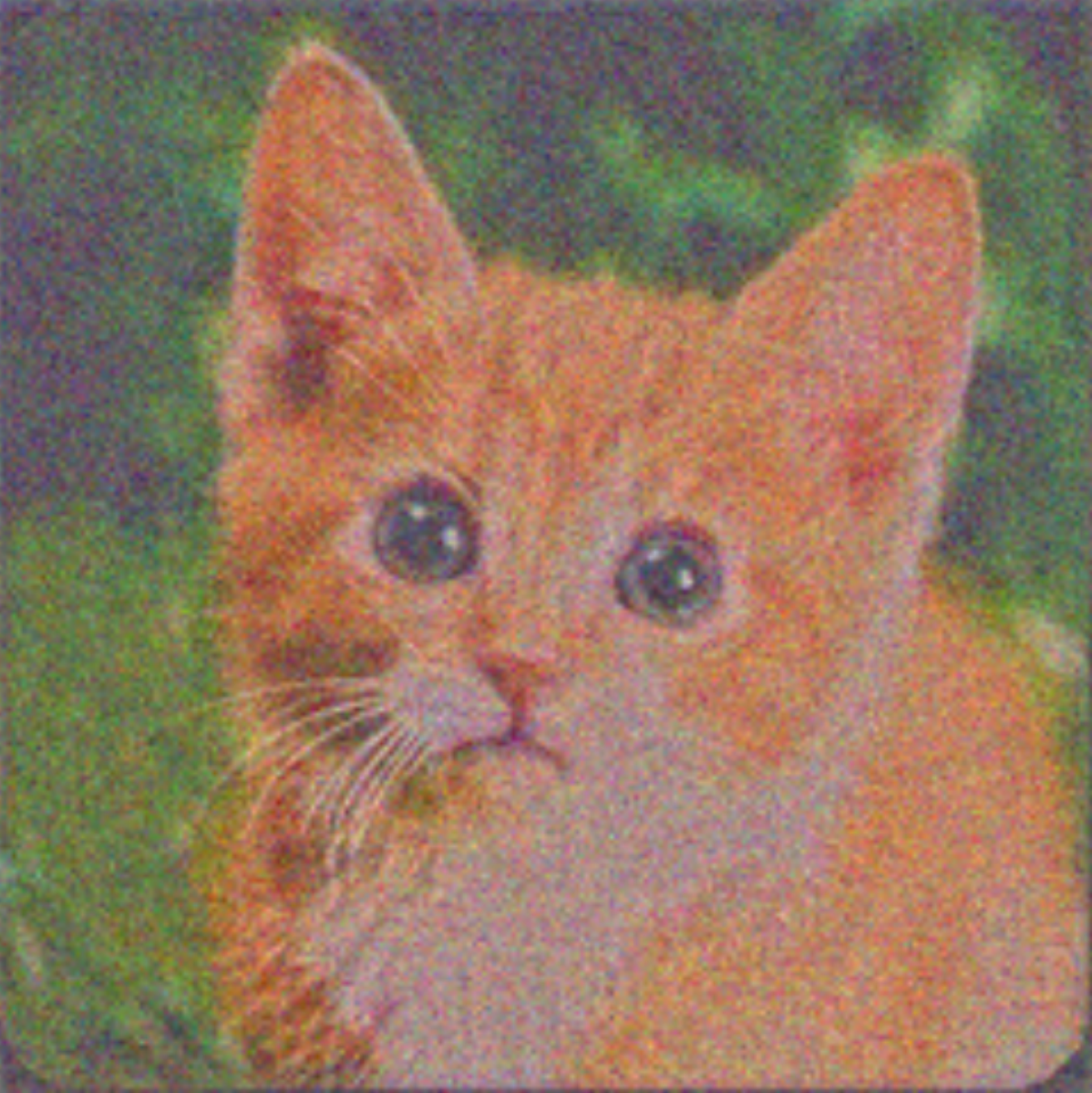}
		\caption*{\footnotesize $N=64$}
	\end{subfigure}

    \begin{subfigure}[b]{0.11\textwidth}
		\centering
		\includegraphics[width=\textwidth]{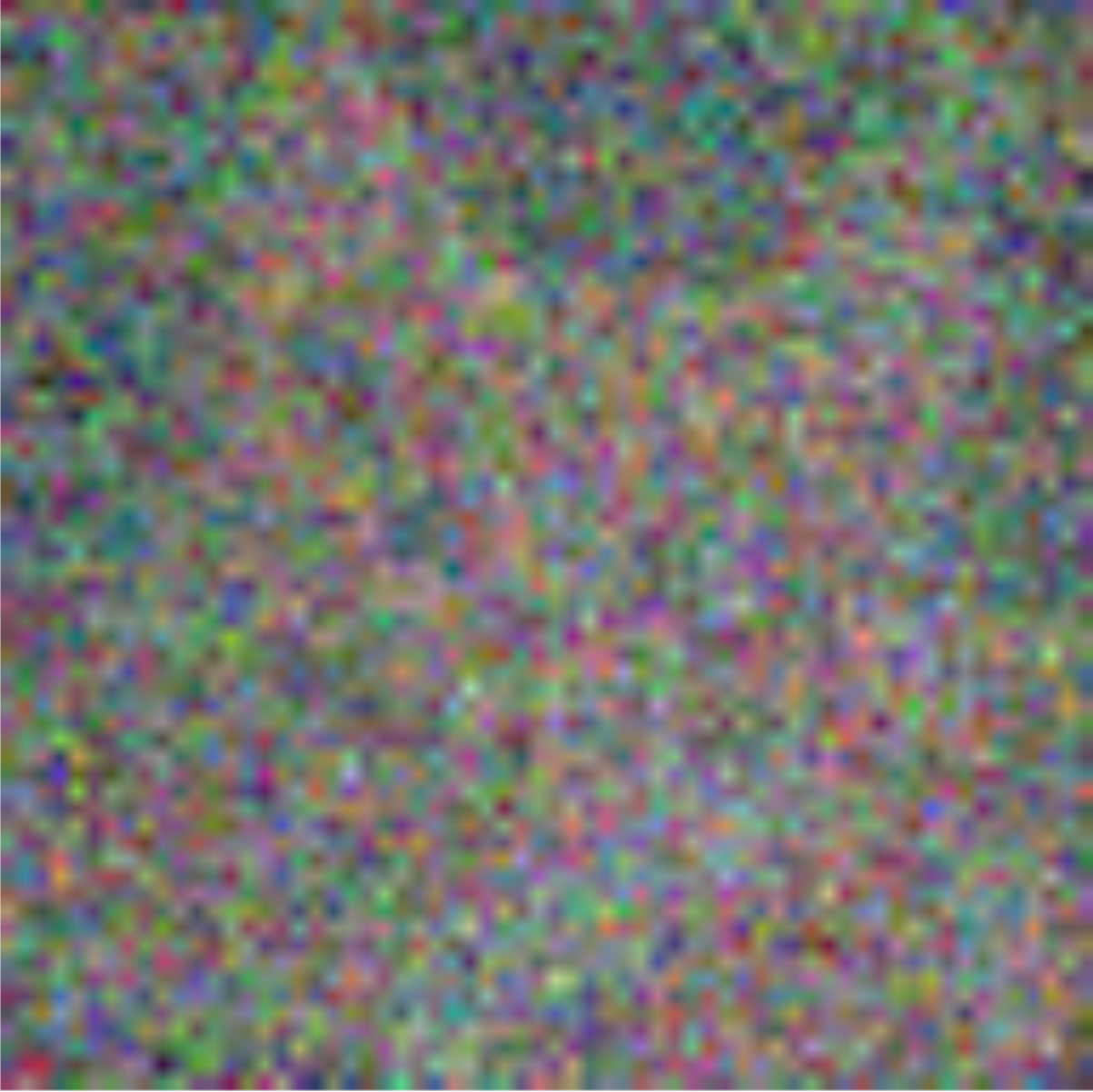}
		\caption*{\footnotesize $N_h=5000$}
	\end{subfigure}
	\centering
	\begin{subfigure}[b]{0.11\textwidth}
		\centering
		\includegraphics[width=\textwidth]{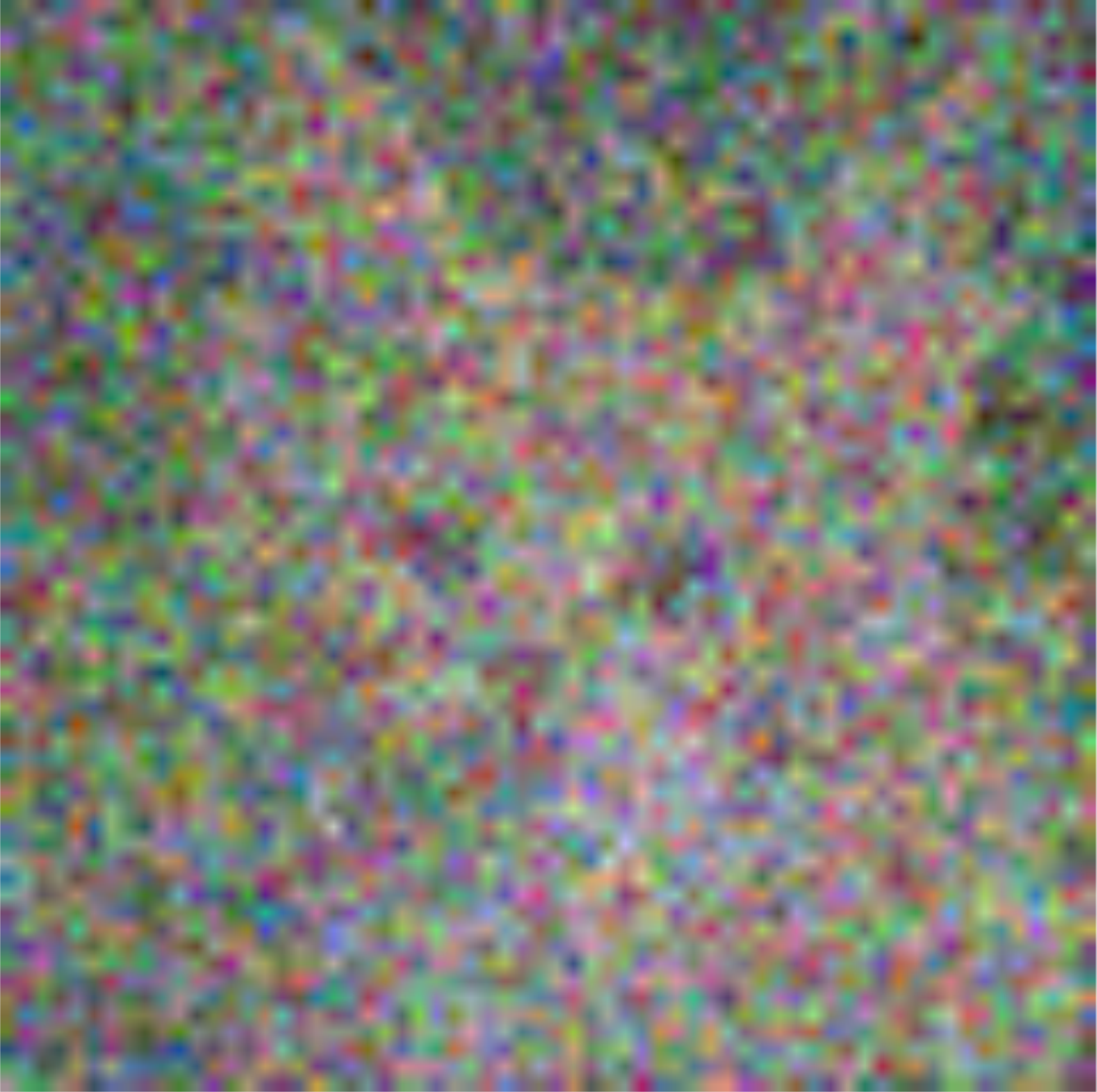}
		\caption*{\footnotesize $N_h=10000$}
	\end{subfigure}
	\centering
	\begin{subfigure}[b]{0.11\textwidth}
		\centering
		\includegraphics[width=\textwidth]{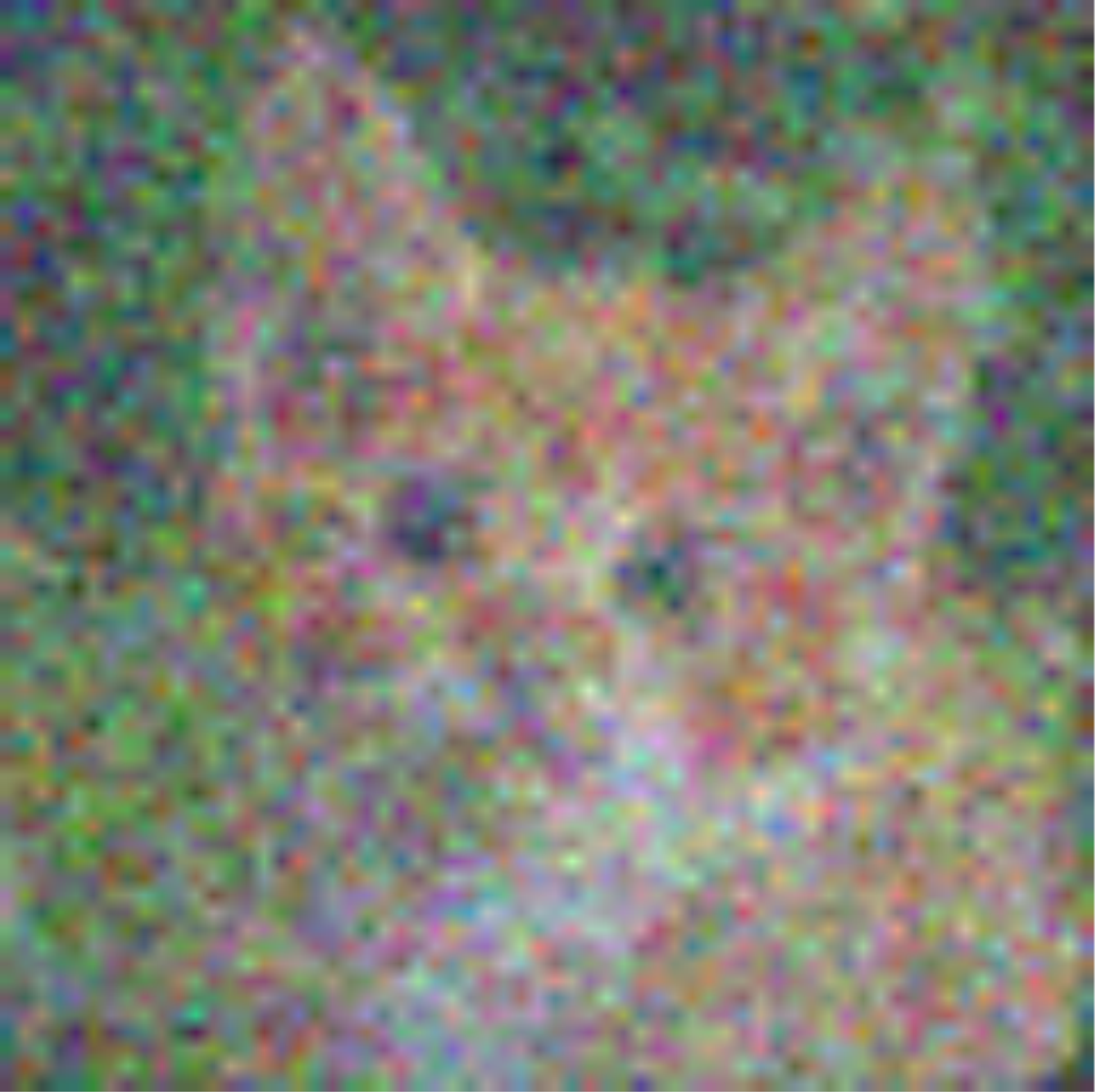}
		\caption*{\footnotesize $N_h=40000$}
	\end{subfigure}
	\begin{subfigure}[b]{0.11\textwidth}
		\centering
		\includegraphics[width=\textwidth]{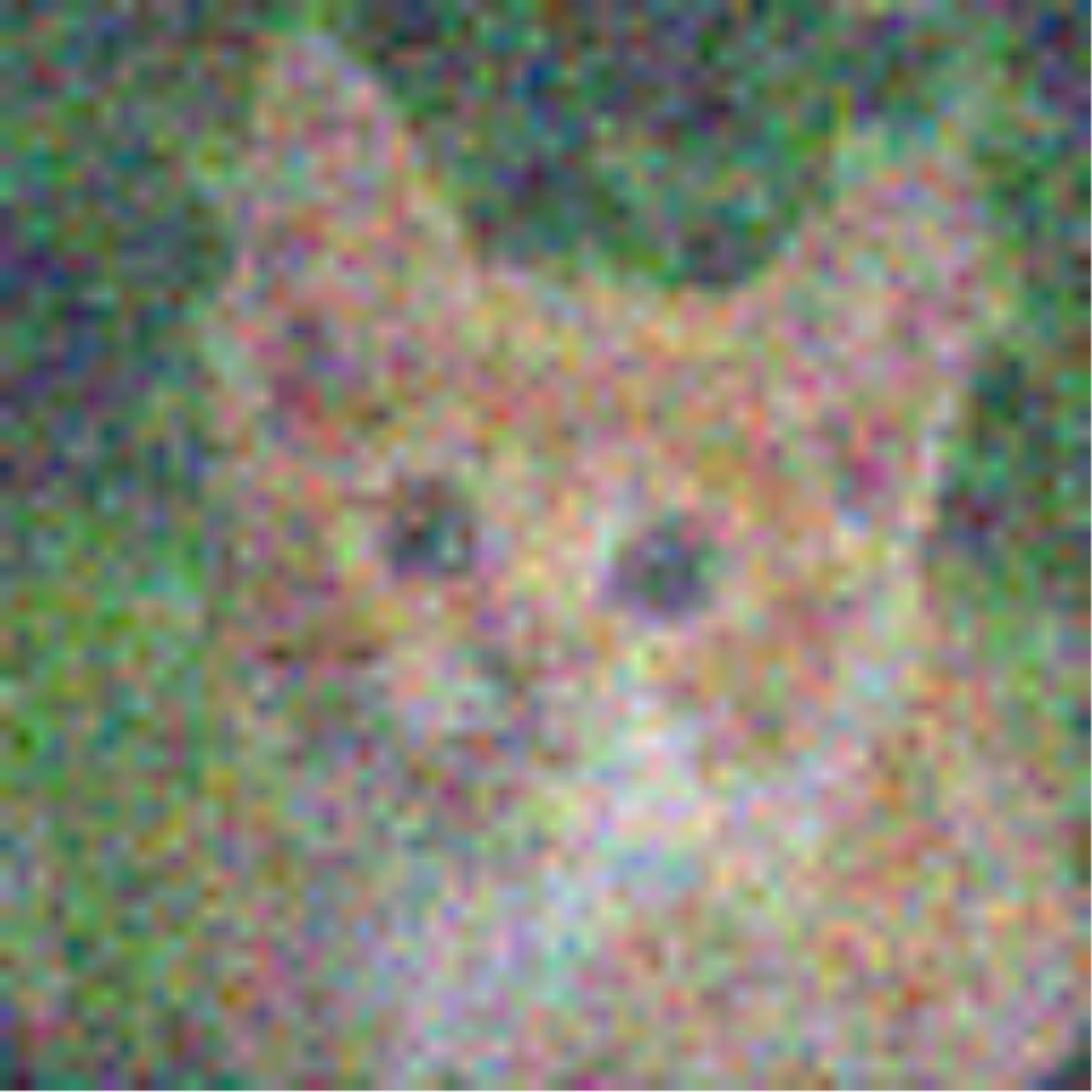}
		\caption*{\footnotesize $N_h=70000$}
	\end{subfigure}
	\caption{GBP visualizations given the input image ``tabby'' in a three-layer CNN (top row) by varying the number of filters $N$ and in a three-layer FCN (bottom row) by varying the number of hidden neurons $N_h$. }\label{gbp_fi}
\end{figure}

To further highlight the impact of local connections in the visual quality of GBP, we vary the number of filters $N$ in the CNN and the number of hidden neurons $N_h$ in the FCN, respectively, while keep other parameters fixed. The results are given in Figure \ref{gbp_fi}. Note that in the FCN, we have downsampled the input image to be of size $64 \times 64 \times 3$ due to computational limitations. We can see that as the number of filters $N$ increases (resp. the hidden layer size $N_h$), the visual quality of GBP in the CNN (resp. in the FCN) becomes better. Interestingly, even by setting $N_h=70000$, which is definitely unrealistic, the FCN cannot achieve a comparable performance to the CNN with $N=64$. Therefore, it confirms that the local connections in the CNN really contribute to the good visual quality of GBP. 


\subsection{Impact of Max-Pooling and Network Depth}

\begin{figure}[!t]
	\centering
	\begin{subfigure}[b]{0.11\textwidth}
		\centering
		\includegraphics[width=\textwidth]{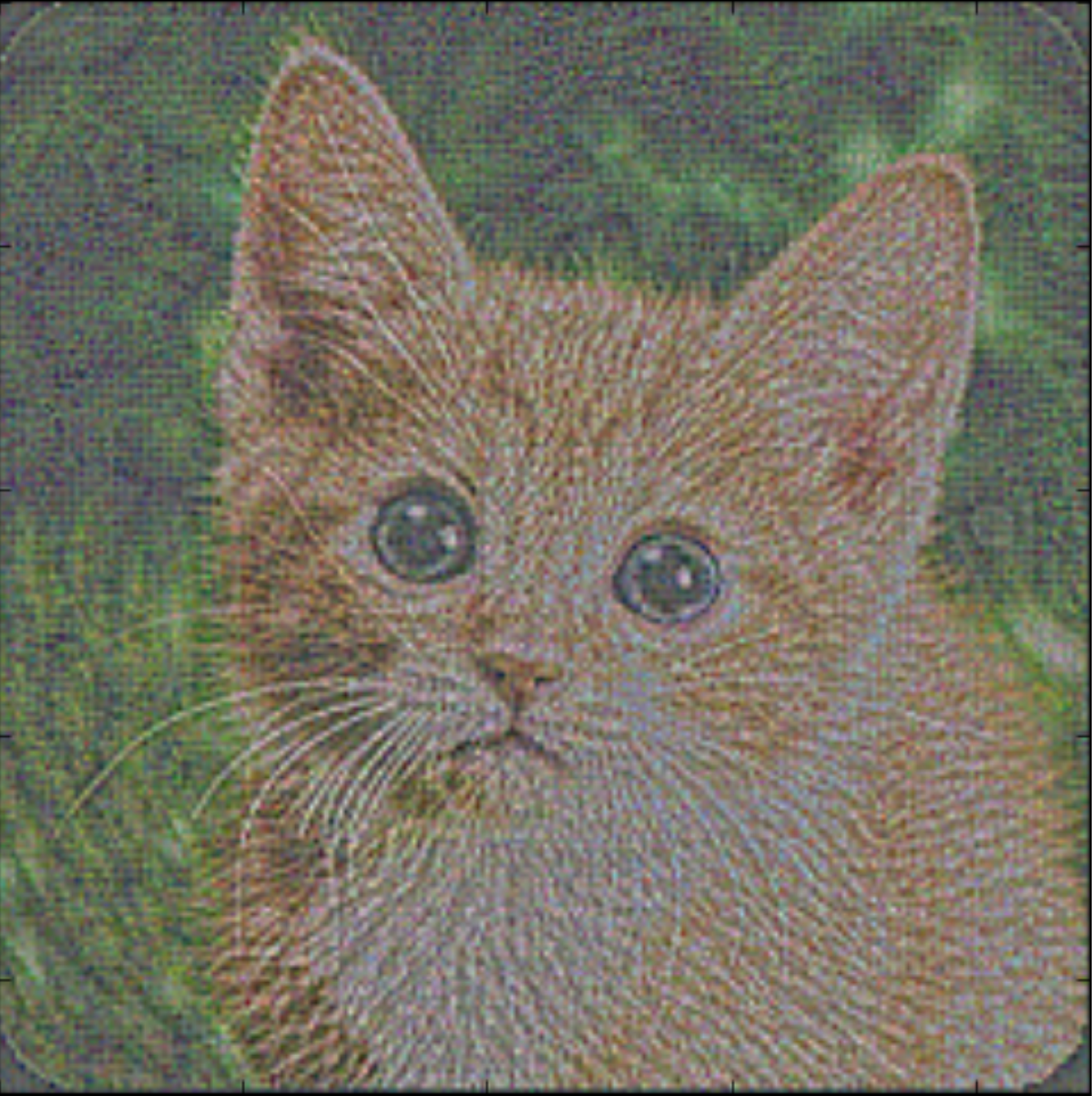}
		\caption*{\footnotesize GBP-pool}
	\end{subfigure}
	\begin{subfigure}[b]{0.11\textwidth}
		\centering
		\includegraphics[width=\textwidth]{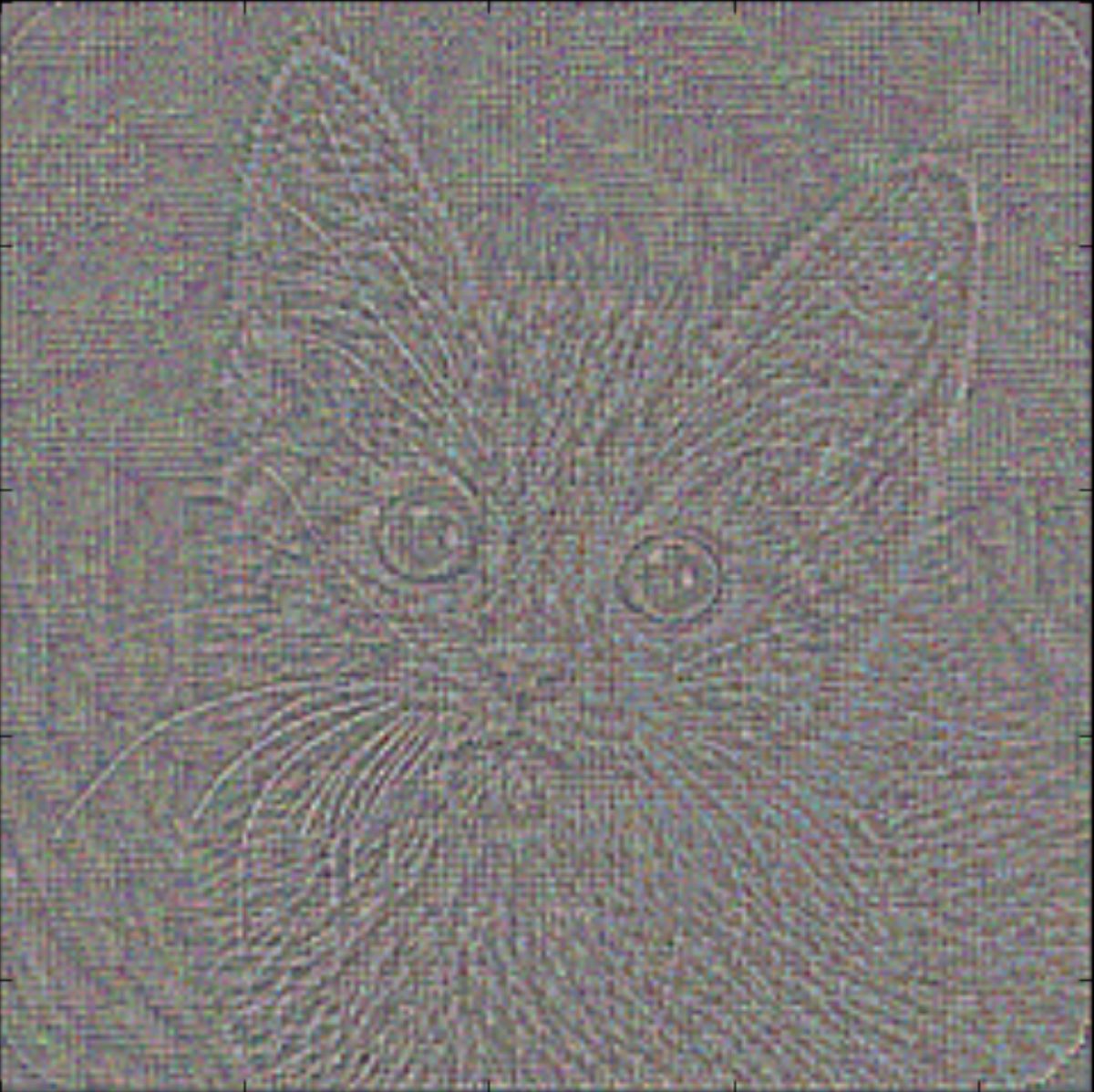}
		\caption*{\footnotesize Deconv-pool}
	\end{subfigure}
	\begin{subfigure}[b]{0.11\textwidth}
		\centering
		\includegraphics[width=\textwidth]{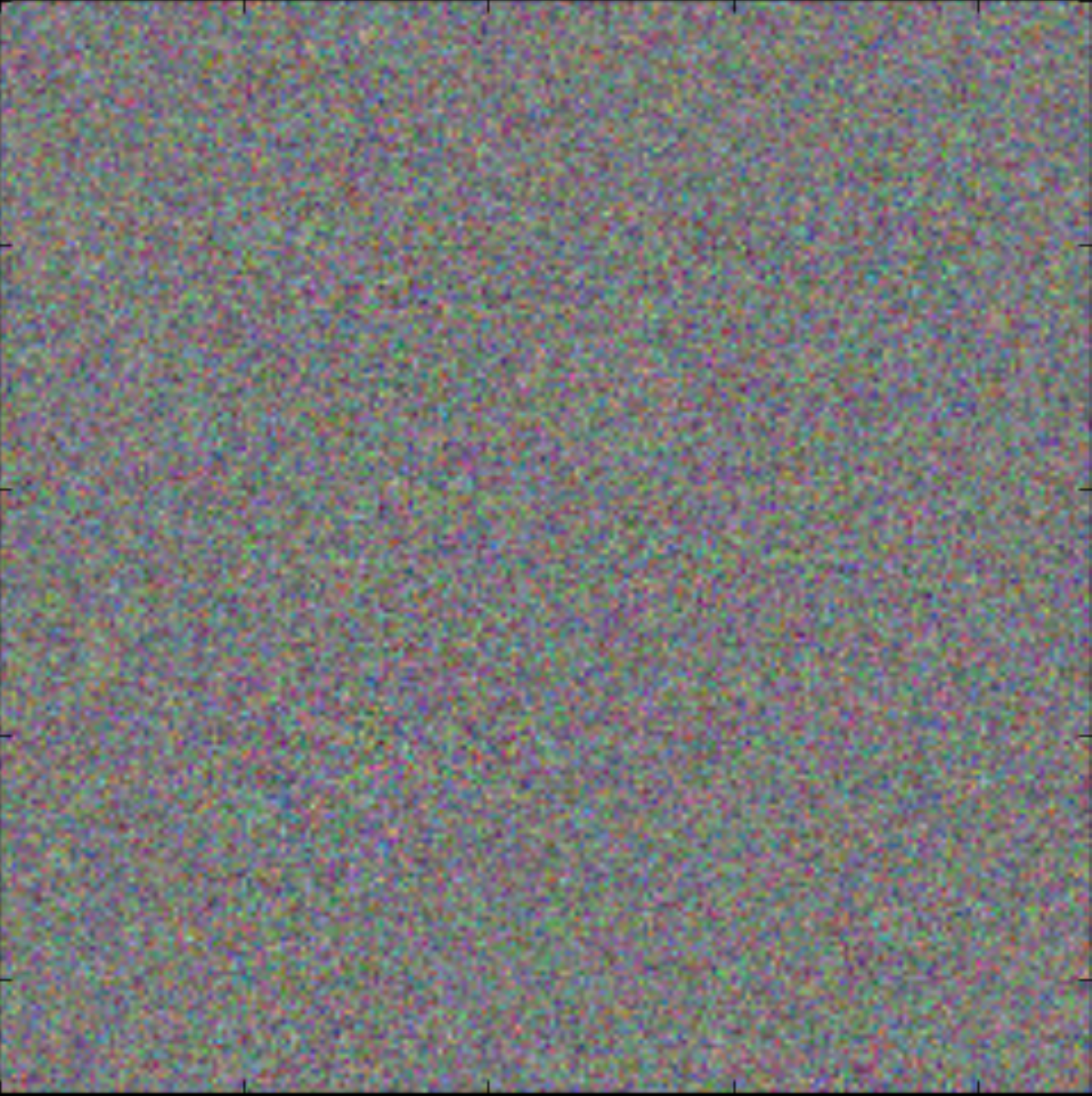}
		\caption*{\footnotesize Sal-pool}
	\end{subfigure}
	
	\begin{subfigure}[b]{0.11\textwidth}
		\centering
		\includegraphics[width=\textwidth]{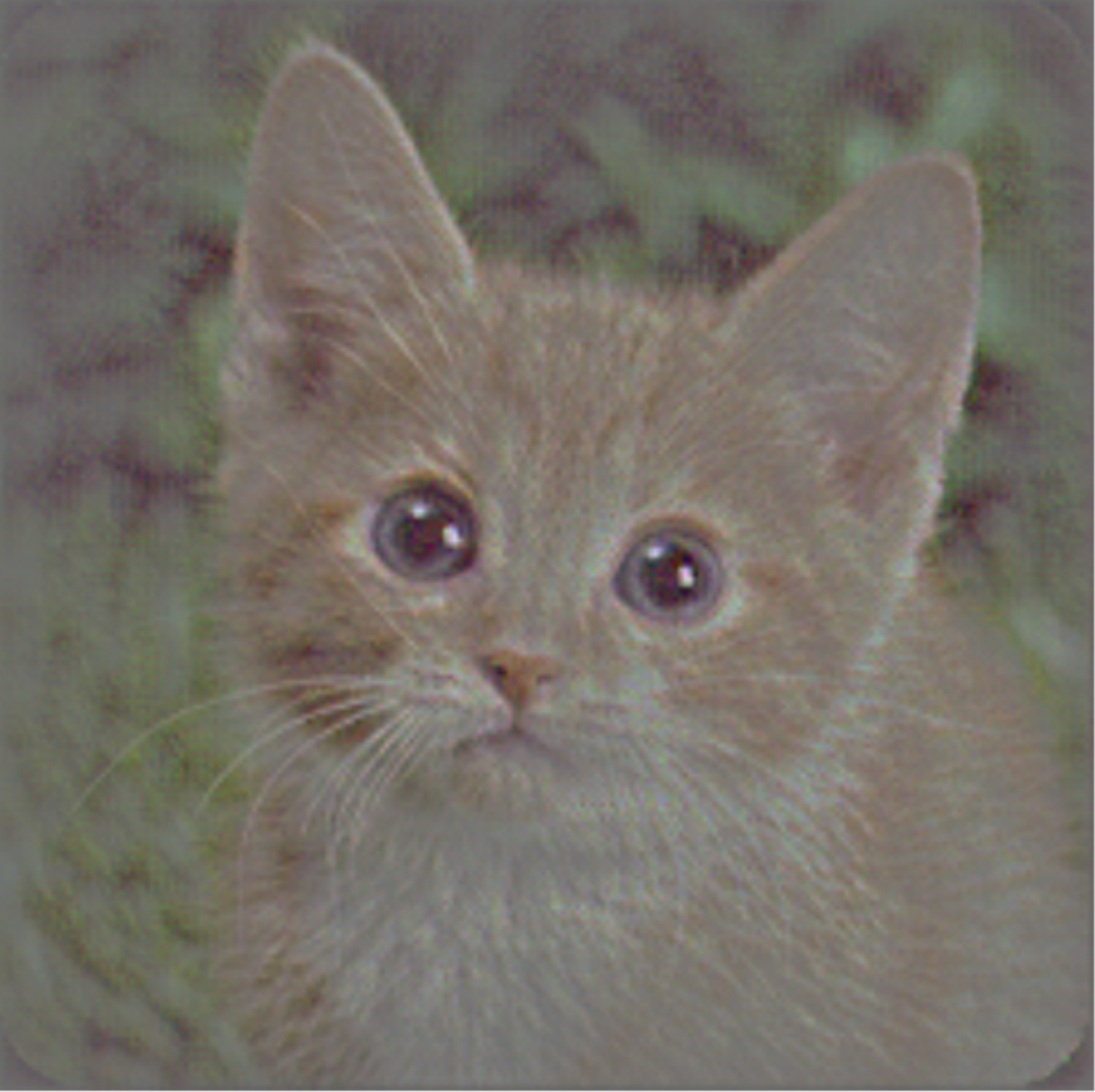}
		\caption*{\footnotesize GBP-VGG}
	\end{subfigure}
    \begin{subfigure}[b]{0.11\textwidth}
		\centering
		\includegraphics[width=\textwidth]{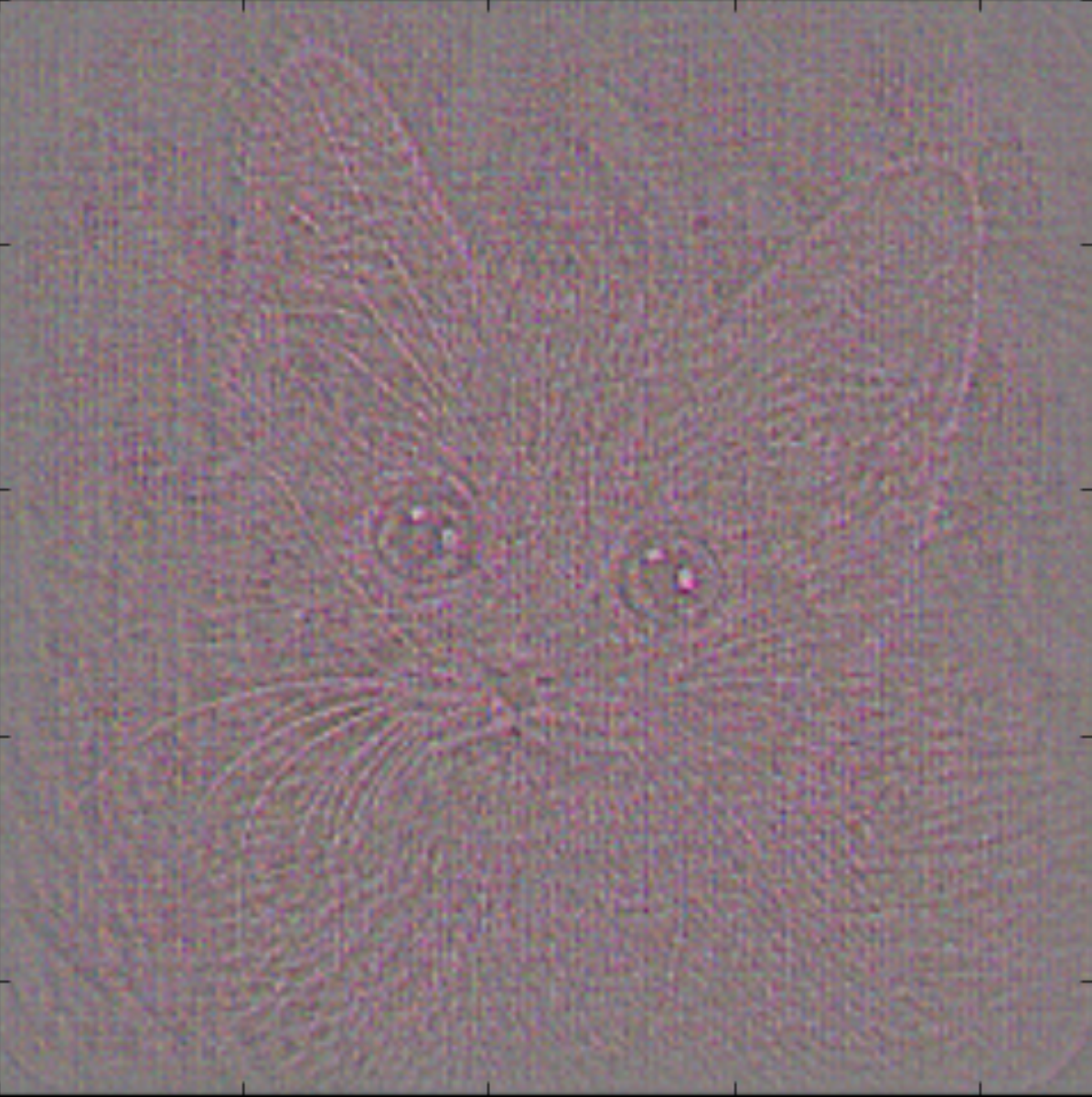}
		\caption*{\footnotesize Deconv-VGG}
	\end{subfigure}
    \begin{subfigure}[b]{0.11\textwidth}
		\centering
		\includegraphics[width=\textwidth]{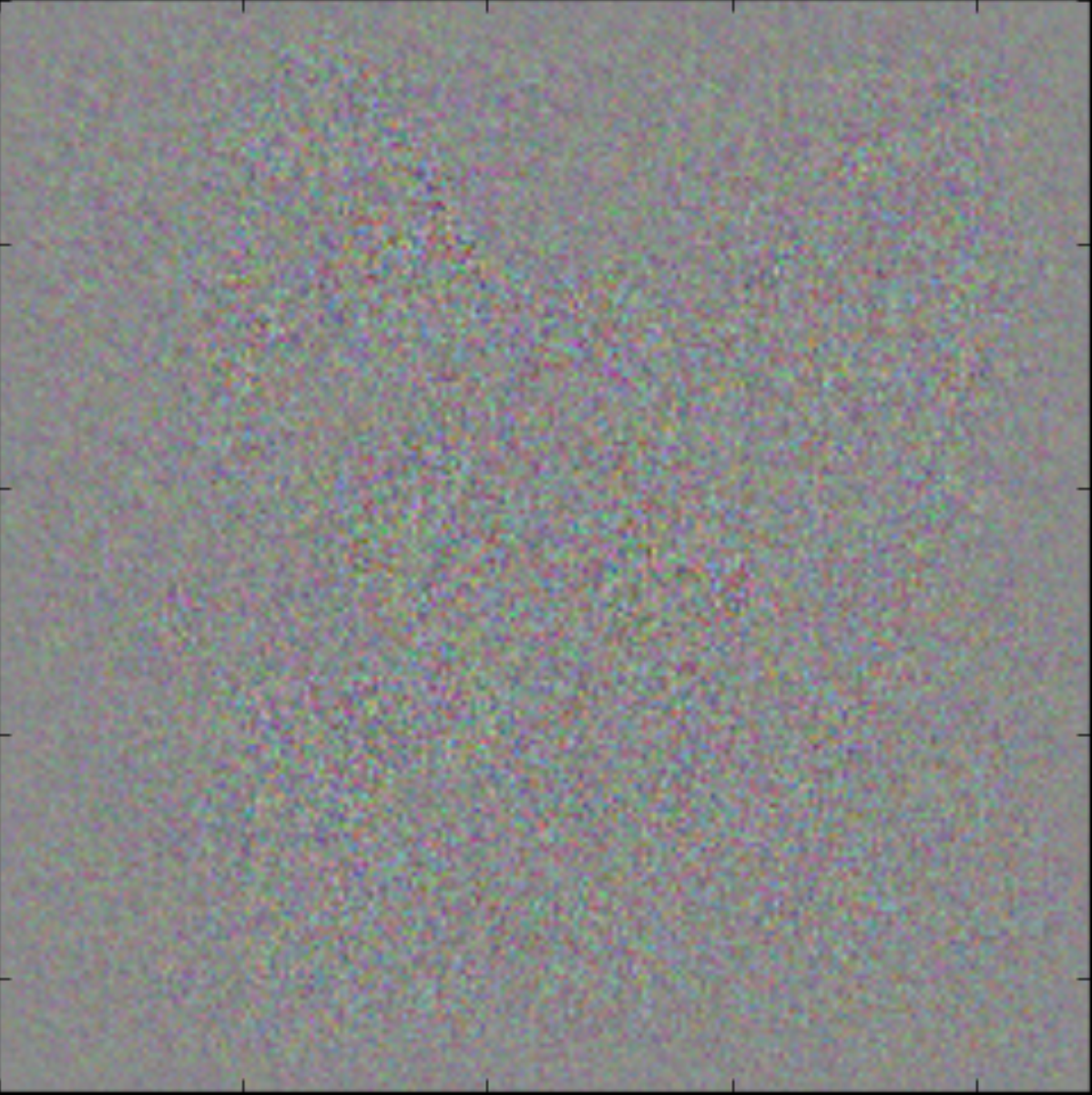}
		\caption*{\footnotesize Sal-VGG}
	\end{subfigure}

	\caption{Backpropagation-based visualizations given the input image ``tabby'' in a random three layer CNN with the max-pooling (top row) and in a random VGG-16 net (bottom row). Now DeconvNet visualization also becomes human-interpretable.}\label{fig_max_pool}
\end{figure}

To show the impact of the max-pooling in backpropagation-based visualizations, we then add a max-pooling layer in the above random three-layer CNN while keeping other parameters fixed, and the results are given in Figure \ref{fig_max_pool} (top row). As compared with the visualizations in Figure \ref{verify_cnn_fcn} (top row), neither GBP or saliency map is impacted by the max-pooling, whereas the DeconvNet visualization has now become human interpretable instead of being the random noise as before. It confirms that the max-pooling is critical in helping DeconvNet produce human-interpretable visualizations via image recovery, as predicted by our theoretical analysis in the section \ref{theory_maxpool}. 


	

To show the impact of network depth, we also apply backpropagation-based visualizations in a random VGG-16 net, which also includes the max-pooling but is much deeper than the three-layer CNN. Figure \ref{fig_max_pool} (bottom row) shows that only saliency map generates random noise while both GBP and DeconvNet could produce human-interpretable visualizations. Though there are subtle visual differences between the top row and bottom row of Figure \ref{fig_max_pool}, the behaviors of backpropagation-based methods are basically unchanged after increasing the network depth. In addition, both GBP and DeconvNet reconstruct every fine-grained detail of the input image in the random VGG , which is different from the trained VGG in Figure \ref{pre_vgg_vis} where only those ``active'' image patches are preserved. 


\subsection{Average $l_2$ Distance Statistics}

\begin{figure} [!t]
    \centering
    \includegraphics[width=0.75\linewidth]{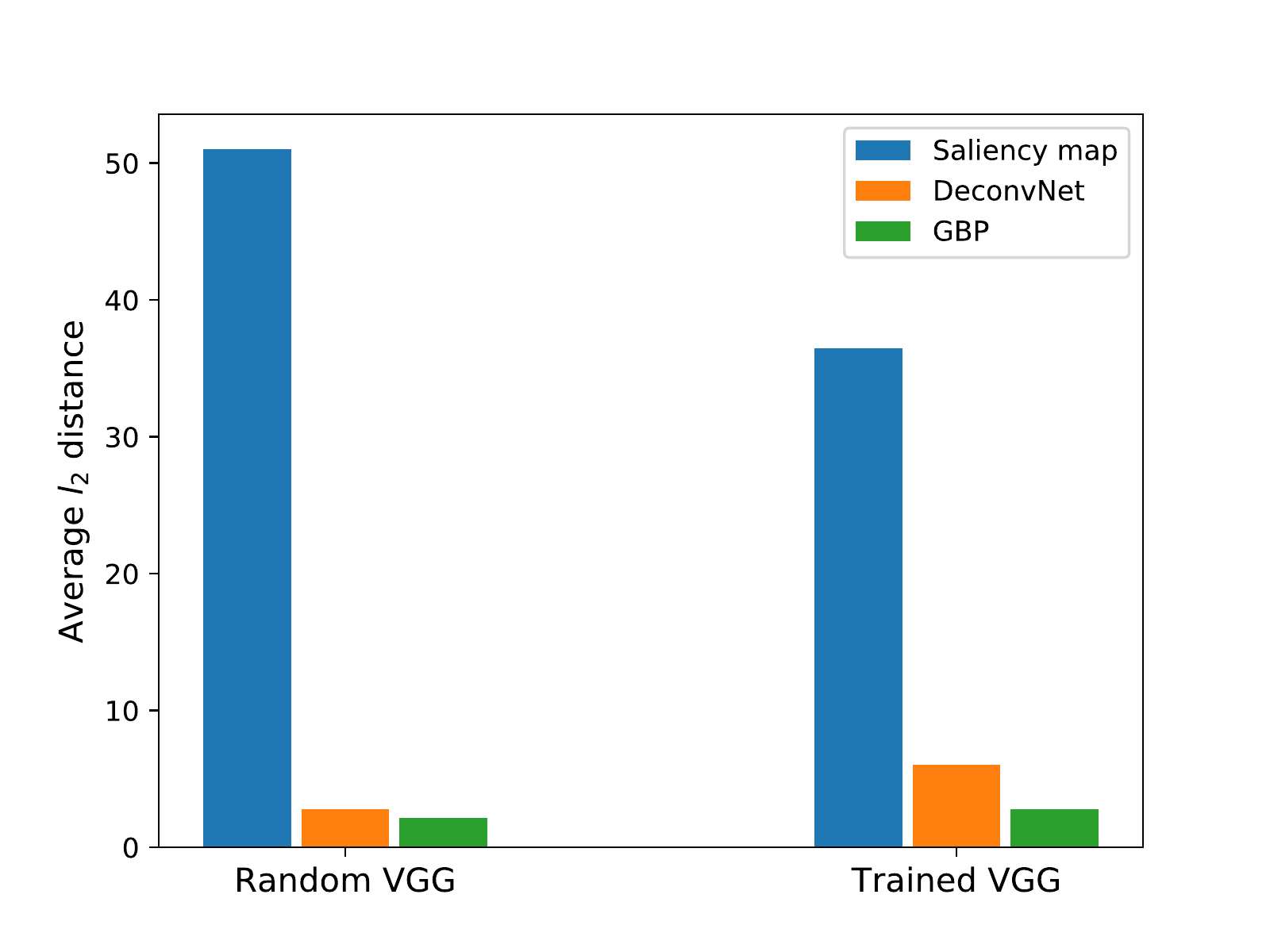}
    \caption{Average $l_2$ distance statistics. For each input, we randomly choose two class logits to get corresponding visualizations and calculate their $l_2$ distances. The above is an average $l_2$ distance by using 10K images of the ImageNet for each backpropagation-based method in both random and trained VGG-16 net.}
    \label{stats_vgg}
\end{figure}

To quantitatively describe how backpropagation-based visualizations change with respect to different class logits, we also provide the average $l_2$ distance statistics as shown in Figure \ref{stats_vgg}. Our results are obtained by first calculating the $l_2$ distance of two visualization results given two different class logits for each input image and then taking an average of those $l_2$ distances based on 10K images from the ImageNet test set. The process is repeated for all backpropagation-based methods in both random and trained cases. As we can see, the average $l_2$ distance of saliency map is much larger than that of both GBP and DeconvNet in either a random VGG or a trained VGG, which clearly demonstrates that saliency map is class-sensitive but GBP and DeconvNet are not. Interestingly, in the trained VGG-16 net, the average $l_2$ distance of DeconvNet is slightly larger than that of GBP. It shows that the class insensitivity is exchanged for further improvement of visual quality.

\subsection{Adversarial Attack on VGG}

\begin{figure}[!t]
	\centering
		\begin{subfigure}[b]{0.11\textwidth}
			\centering
			\includegraphics[width=\textwidth]{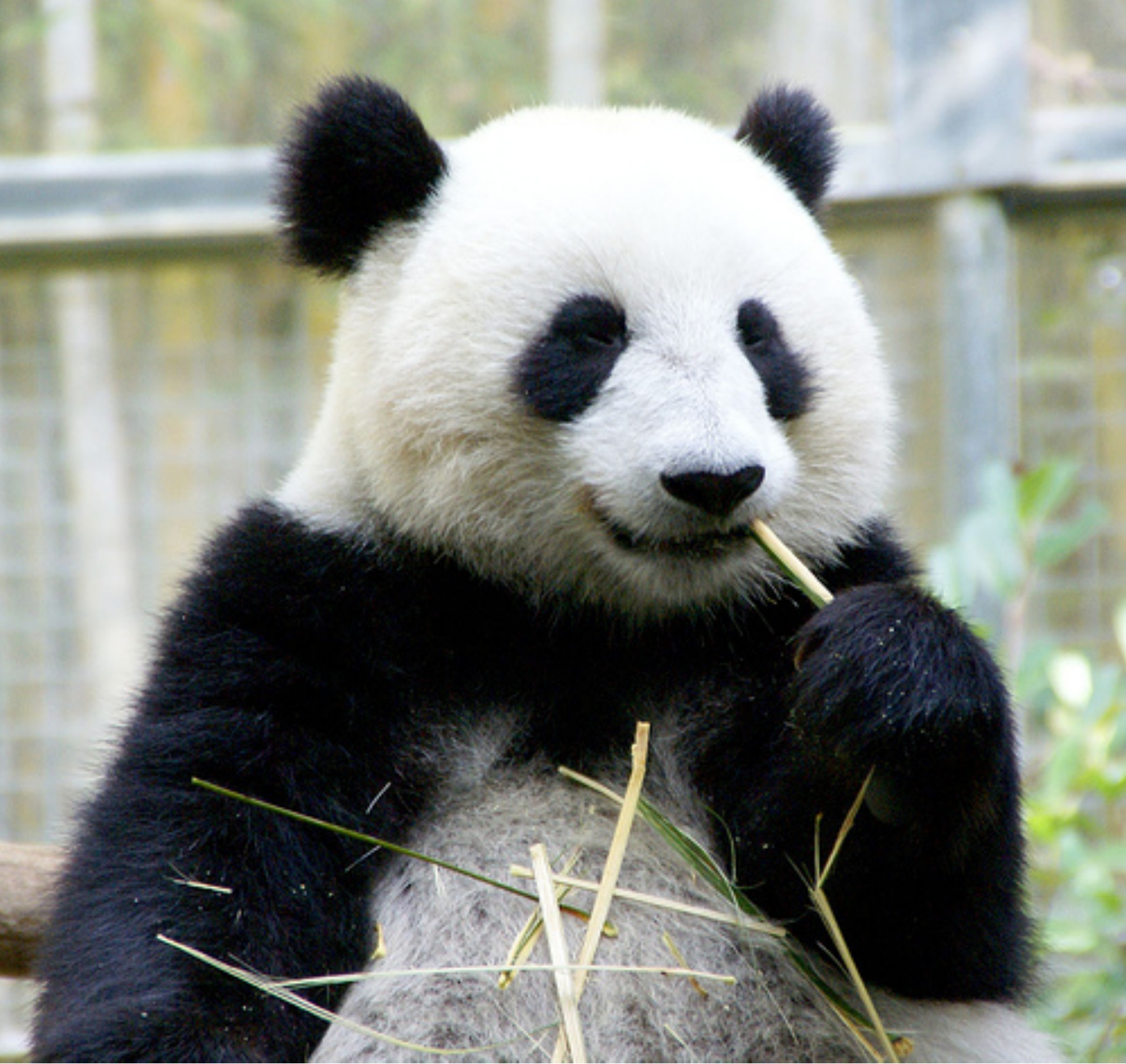}
			\caption*{\footnotesize{panda}}
		\end{subfigure}
		\begin{subfigure}[b]{0.11\textwidth}
			\centering
			\includegraphics[width=\textwidth]{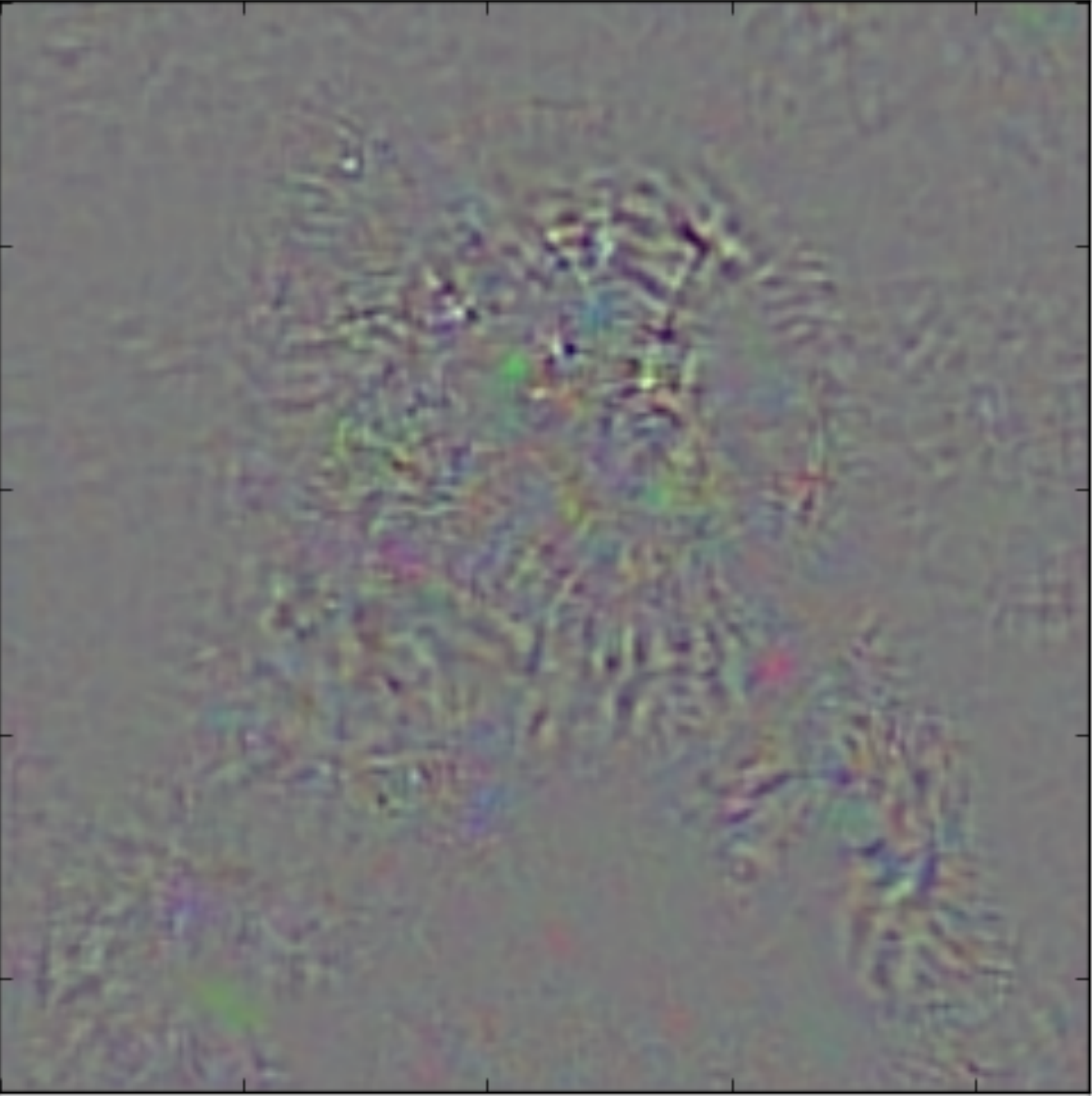}
			\caption*{saliency map}
		\end{subfigure}
		\begin{subfigure}[b]{0.11\textwidth}
			\centering
			\includegraphics[width=\textwidth]{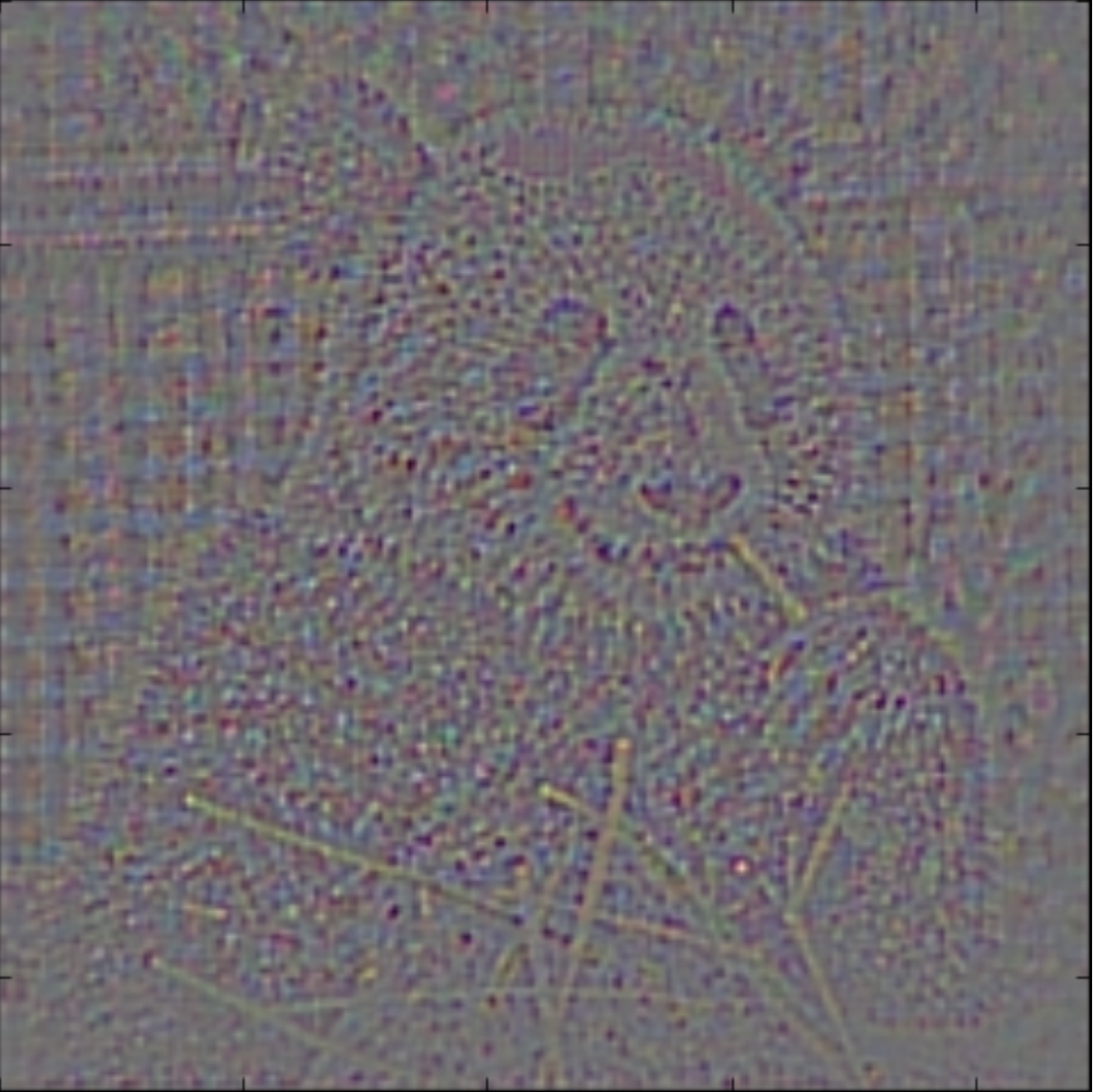}
			\caption*{DeconvNet}
		\end{subfigure}
		\begin{subfigure}[b]{0.11\textwidth}
			\centering
			\includegraphics[width=\textwidth]{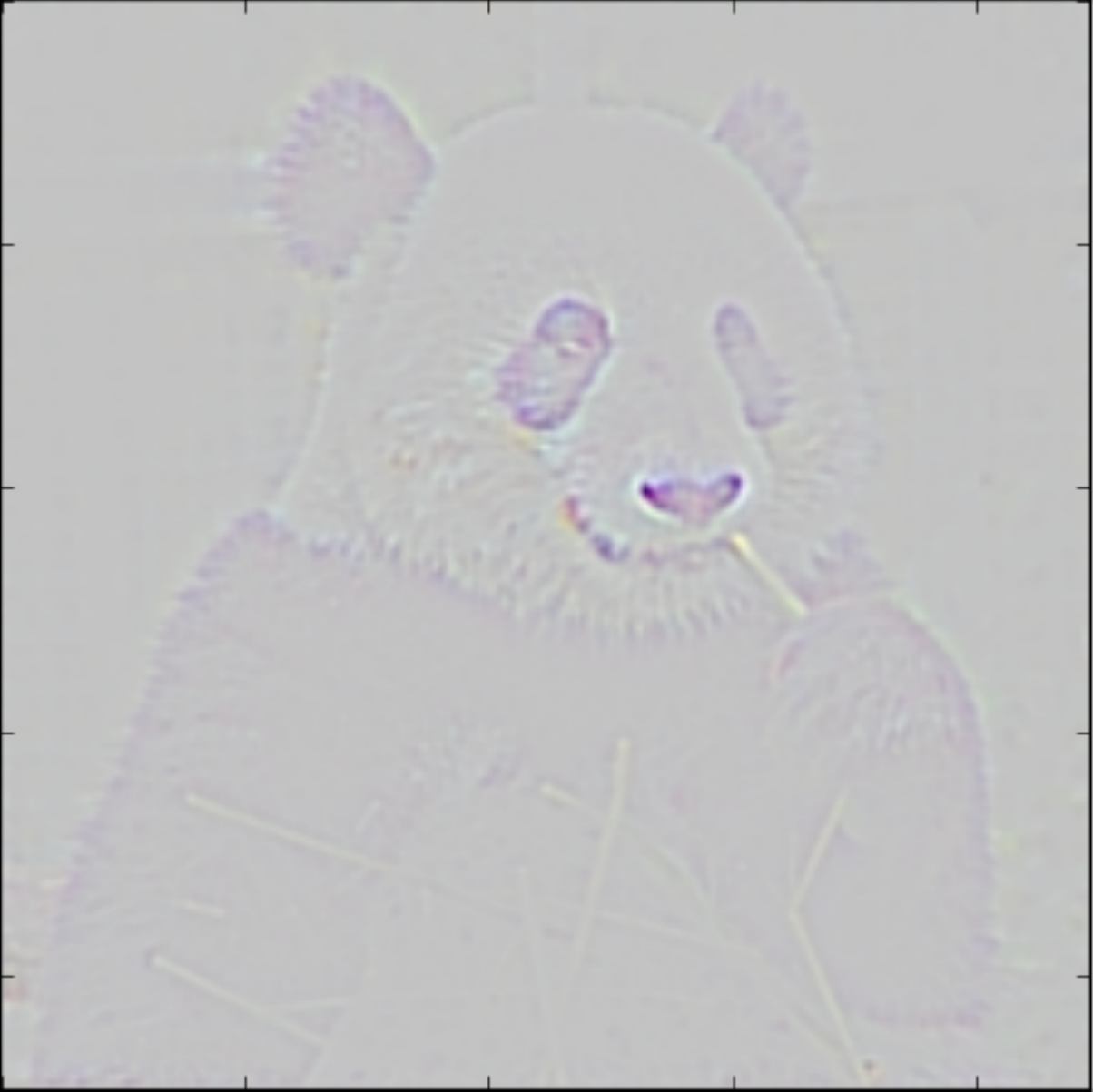}
			\caption*{GBP}
		\end{subfigure}
		
		\begin{subfigure}[b]{0.11\textwidth}
			\centering
			\includegraphics[width=\textwidth]{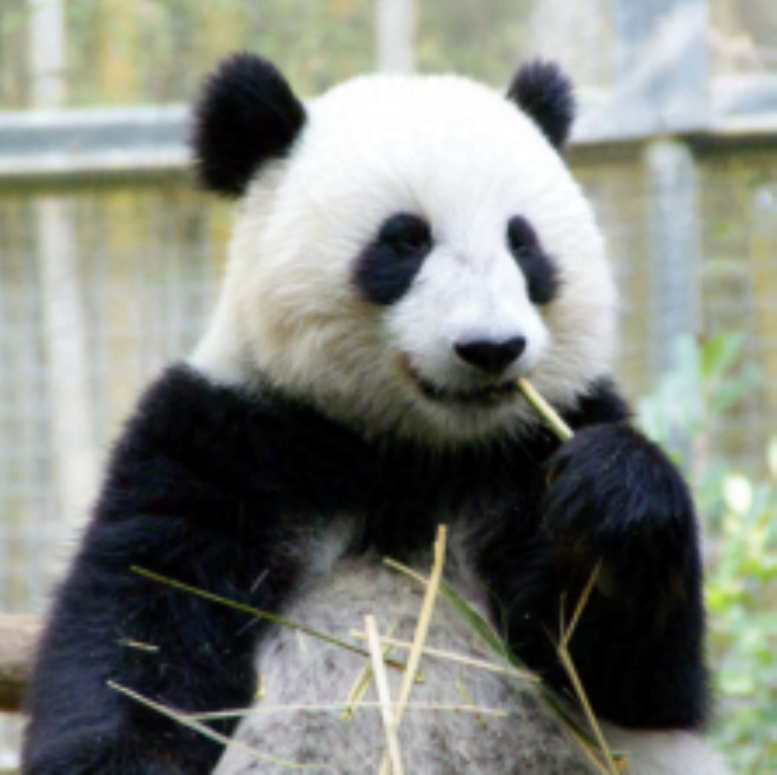}
			\caption*{busby}
		\end{subfigure}
		\begin{subfigure}[b]{0.11\textwidth}
			\centering
			\includegraphics[width=\textwidth]{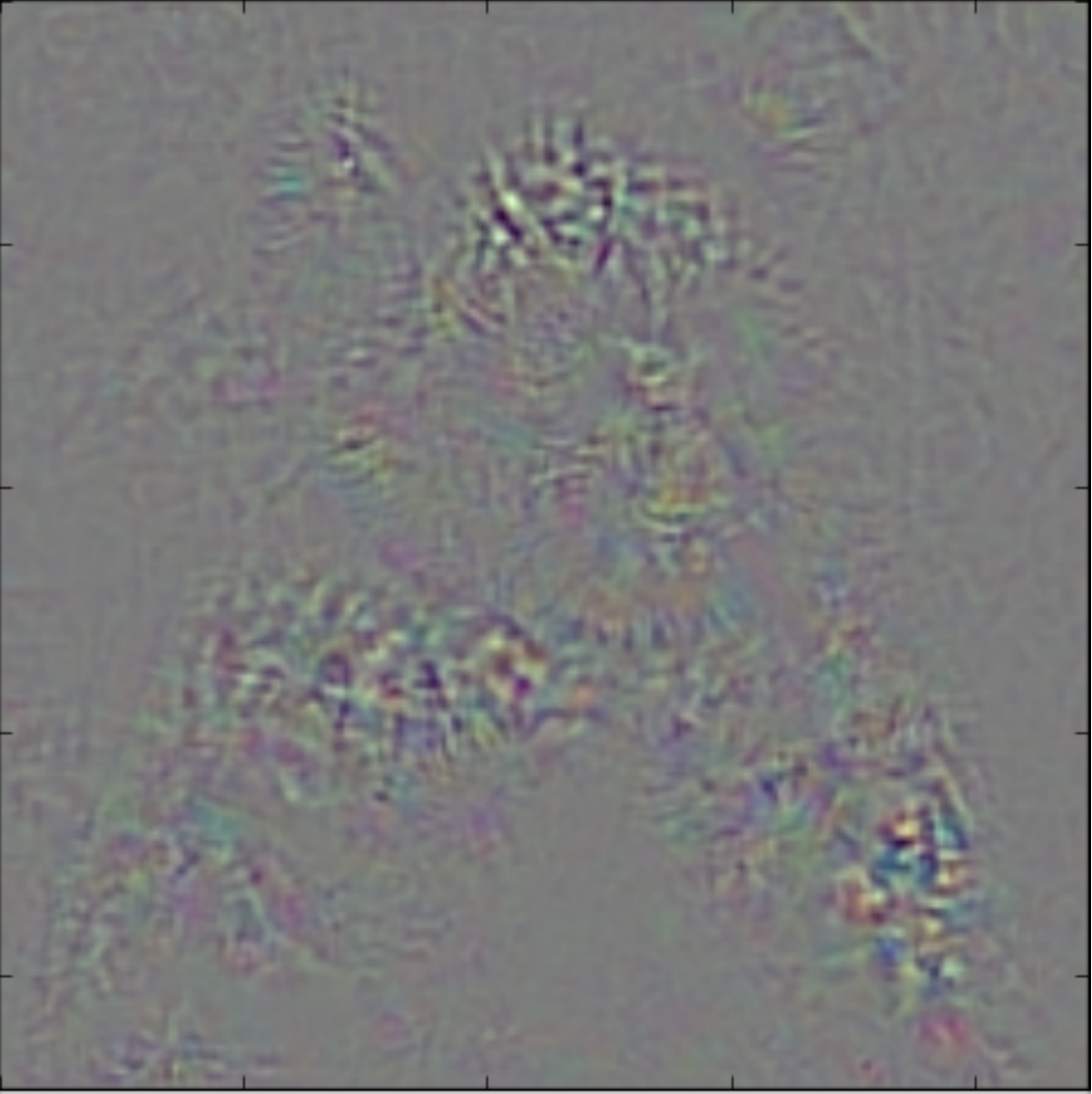}
			\caption*{saliency map}
		\end{subfigure}
		\begin{subfigure}[b]{0.11\textwidth}
			\centering
			\includegraphics[width=\textwidth]{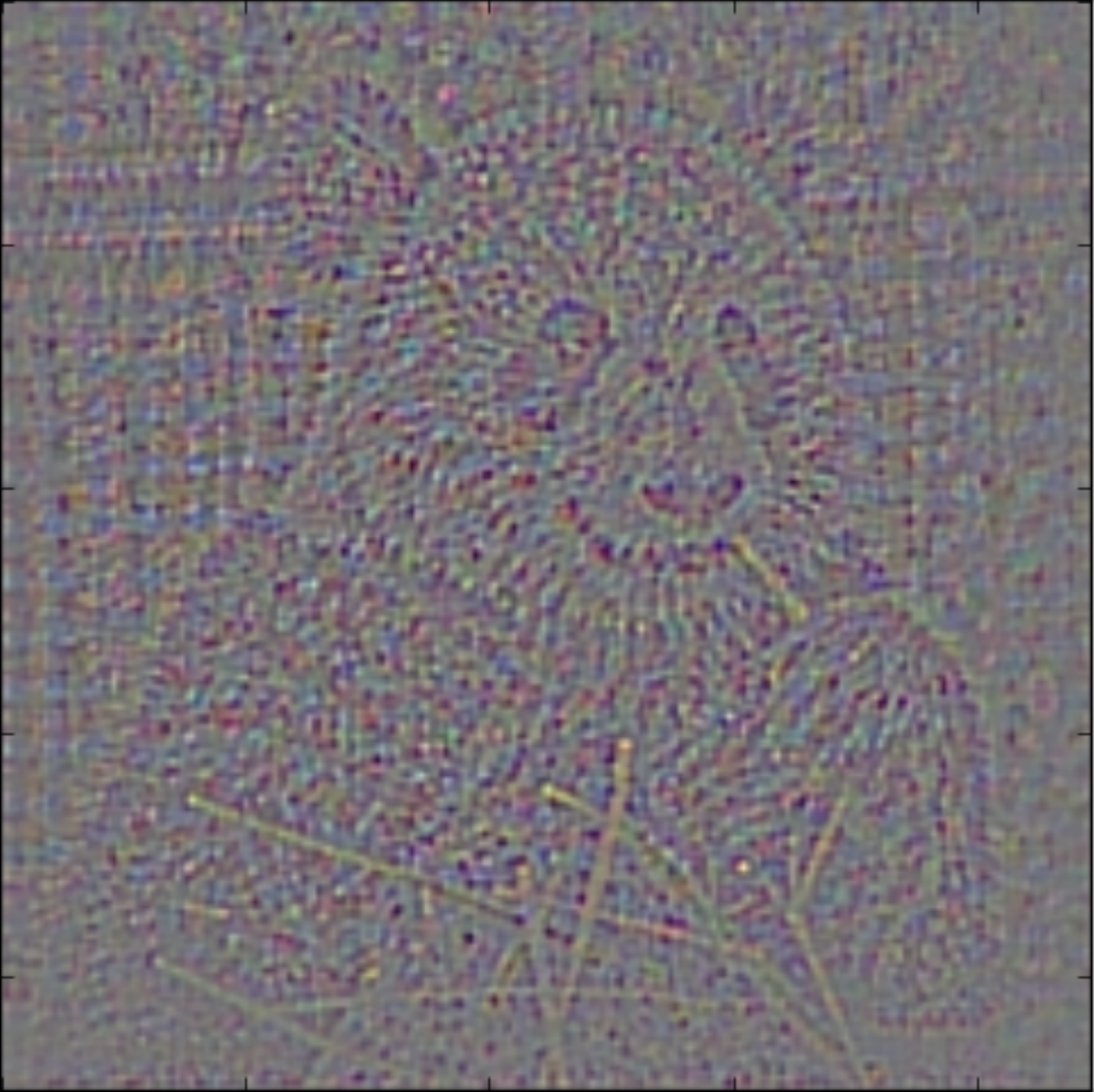}
			\caption*{DeconvNet}
		\end{subfigure}
		\begin{subfigure}[b]{0.11\textwidth}
			\centering
			\includegraphics[width=\textwidth]{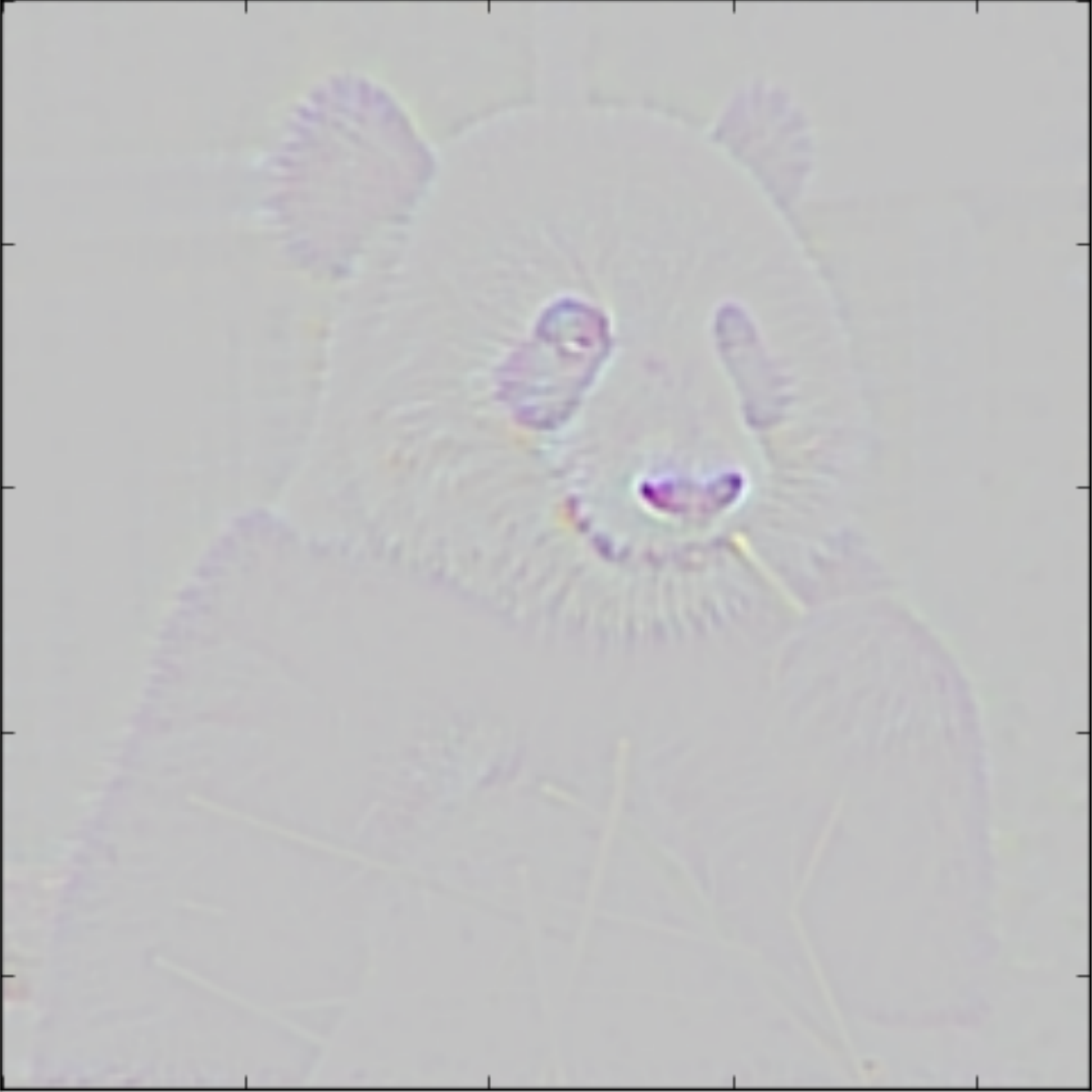}
			\caption*{GBP}
		\end{subfigure}

	\caption{Top row: the image ``panda'' and its backpropagation-based visualizations. Bottom row: the adversarial example misclassified as ``busby'' and its backpropagation-based visualizations. Both experiments are applied in the trained VGG-16 net.}\label{adv_ex}
	
\end{figure}

Adversarial attack provides another way of directly testing whether visualizations are class-sensitive or doing image recovery. 
The class-sensitive visualizations should change drastically as both the predicted class label and ReLU states of intermediate layers have changed, while the visualizations doing image recovery should change little as only a tiny adversarial perturbation is added into the input image.
In this experiment, we first generate an adversarial example ``busby'' via the fast gradient sign method (FGSM) \cite{goodfellow2014explaining} by feeding the image ``panda'' into the pre-trained VGG-16 net.
Next, we apply the backpropagation-based visualizations to the original image ``panda'' and its adversary ``busby'' in the trained VGG-16 net. 
As shown in Figure \ref{adv_ex}, the saliency map visualization changes significantly whereas the GBP and DeconvNet visualizations remain almost unchanged after replacing ``panda'' by its adversary ``busby''. 
Therefore, it further confirms that saliency map is class-sensitive in that it highlights important pixels in making classification decisions. However, 
GBP and DeconvNet are doing nothing but (partial) image recovery.

\subsection{VGG with Partly Trained Weights}

There exist some differences for backpropagation-based
visualizations, GBP and DeconvNet in particular, between the random and trained cases. 
We take GBP as an example here to investigate the contributions of different layers in the trained VGG-16 net to these visual differences. 

\begin{figure}[!t]
	\centering
		\begin{subfigure}[b]{0.11\textwidth}
			\centering
			\includegraphics[width=\textwidth]{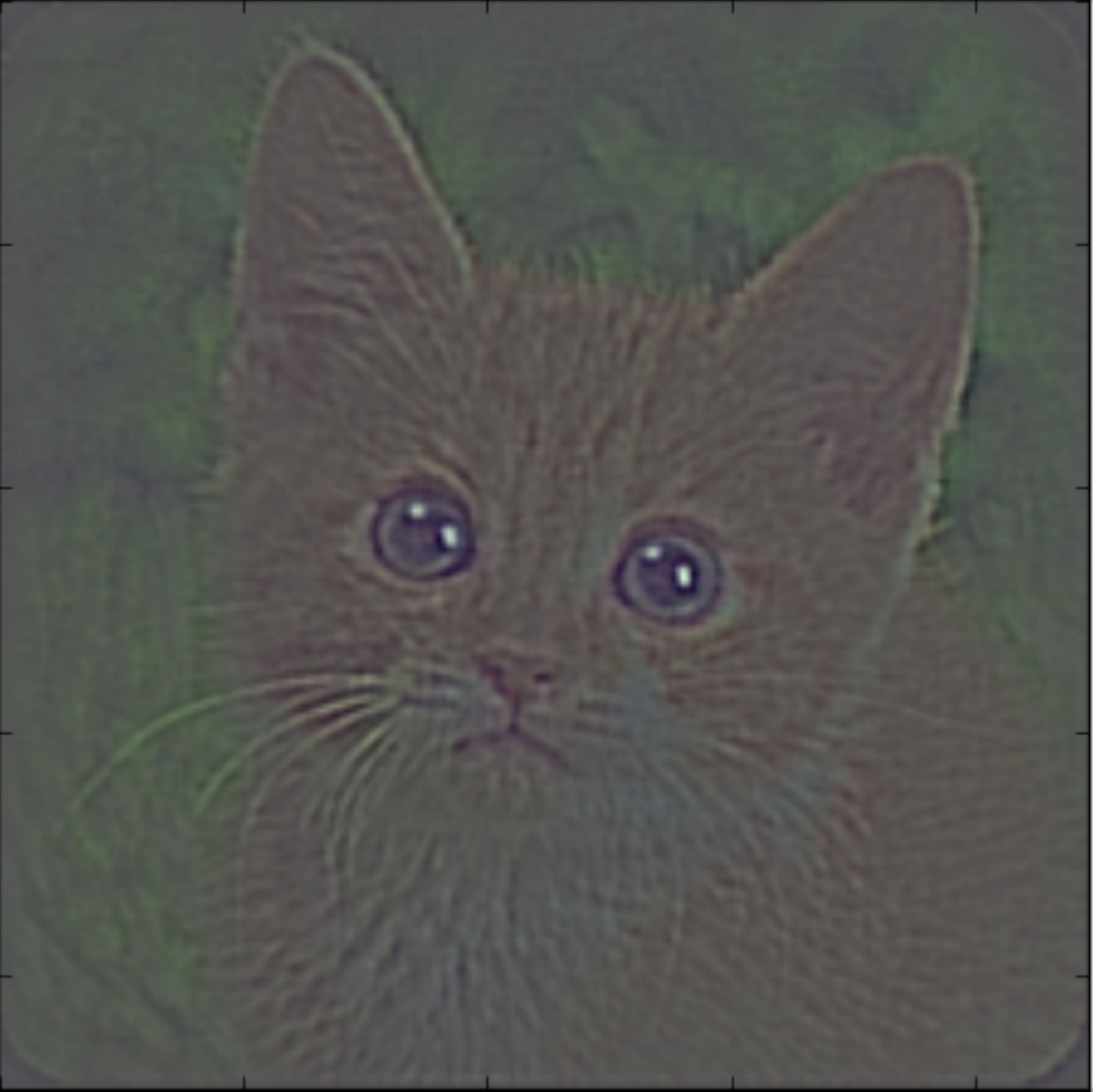}
			\caption*{Conv1-1*}
		\end{subfigure}
		\begin{subfigure}[b]{0.11\textwidth}
			\centering
			\includegraphics[width=\textwidth]{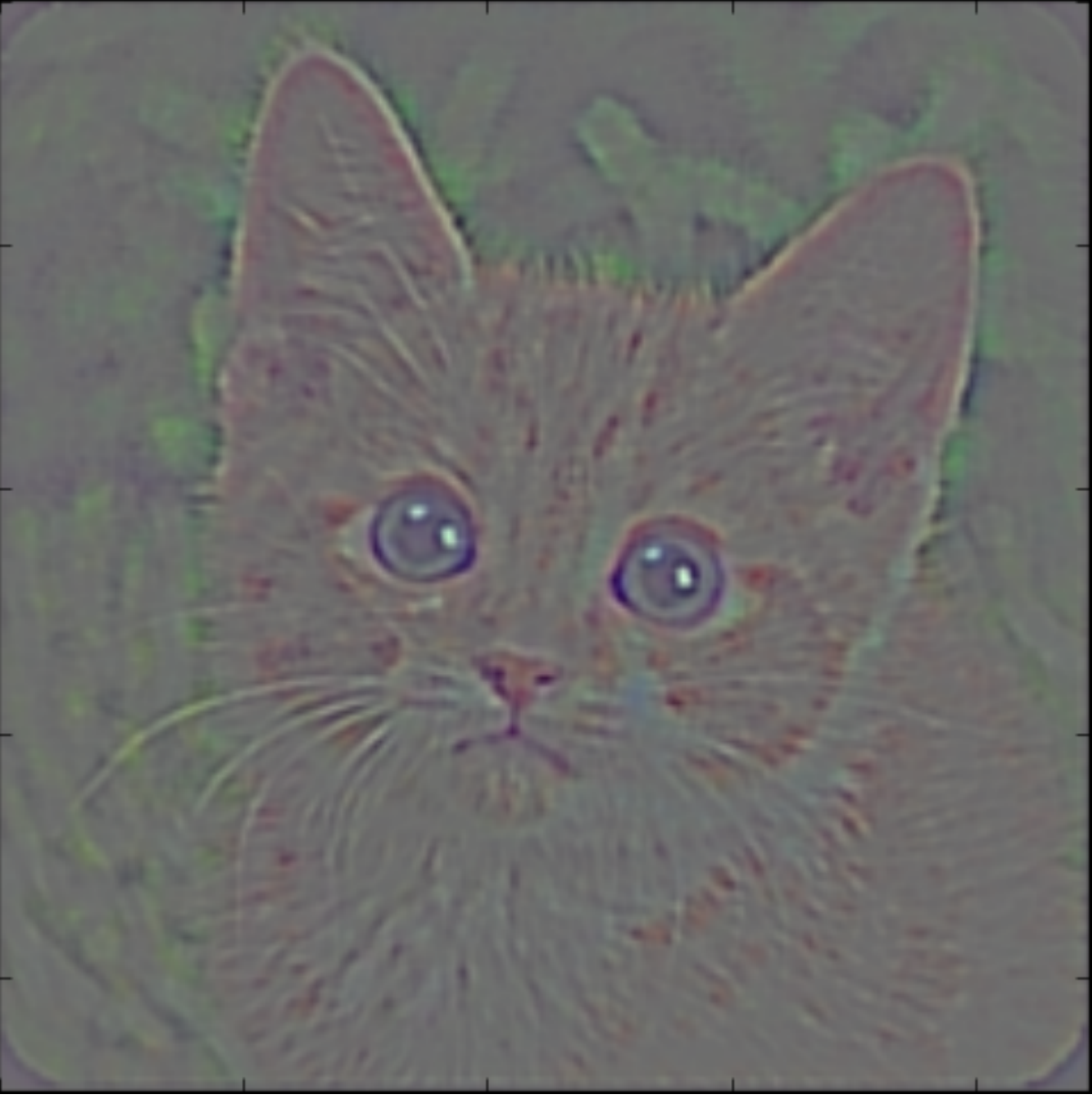}
			\caption*{Conv3-1*}
		\end{subfigure}
		\begin{subfigure}[b]{0.11\textwidth}
			\centering
			\includegraphics[width=\textwidth]{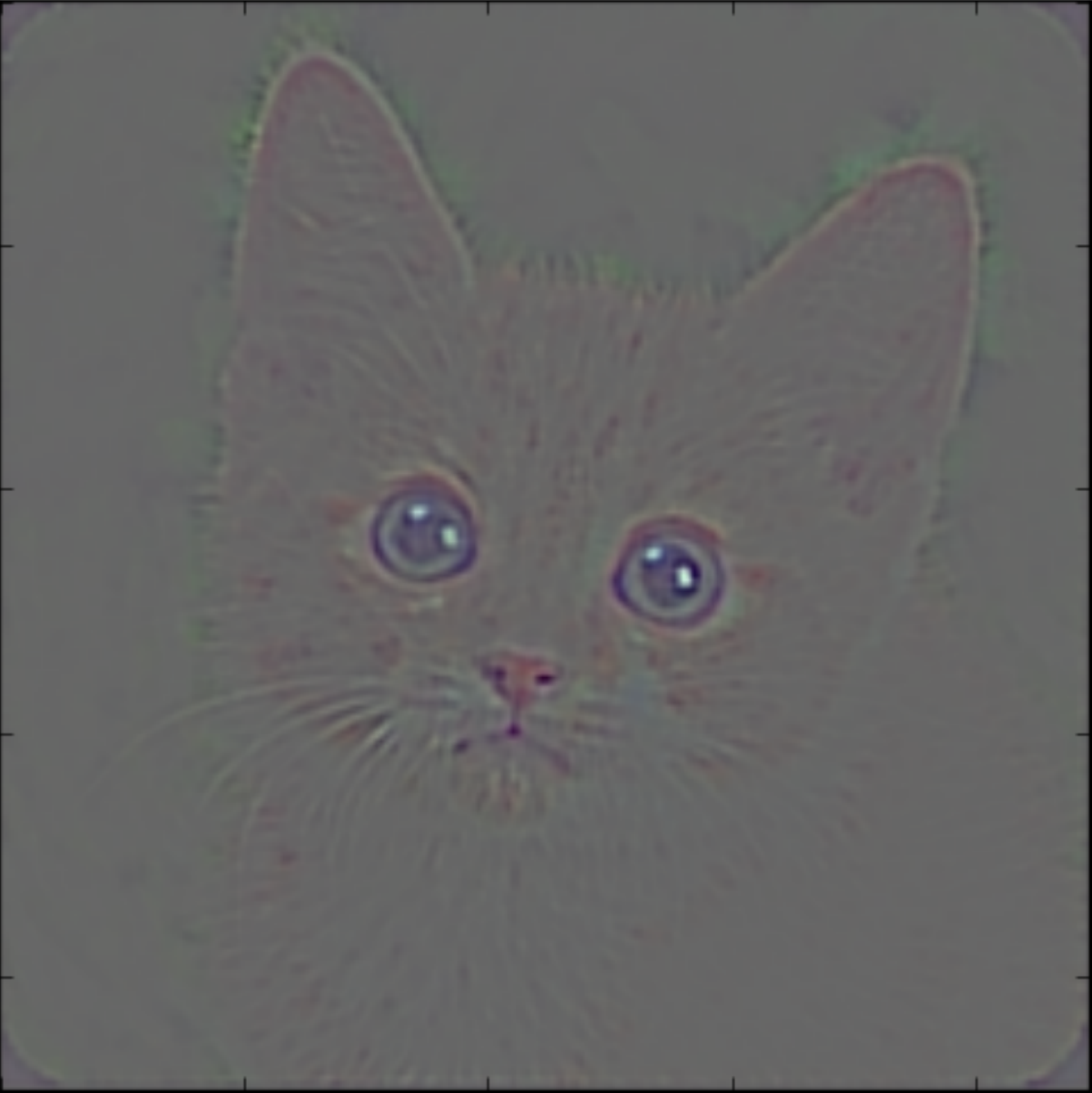}
			\caption*{Conv5-1*}
		\end{subfigure}
		\begin{subfigure}[b]{0.11\textwidth}
			\centering
			\includegraphics[width=\textwidth]{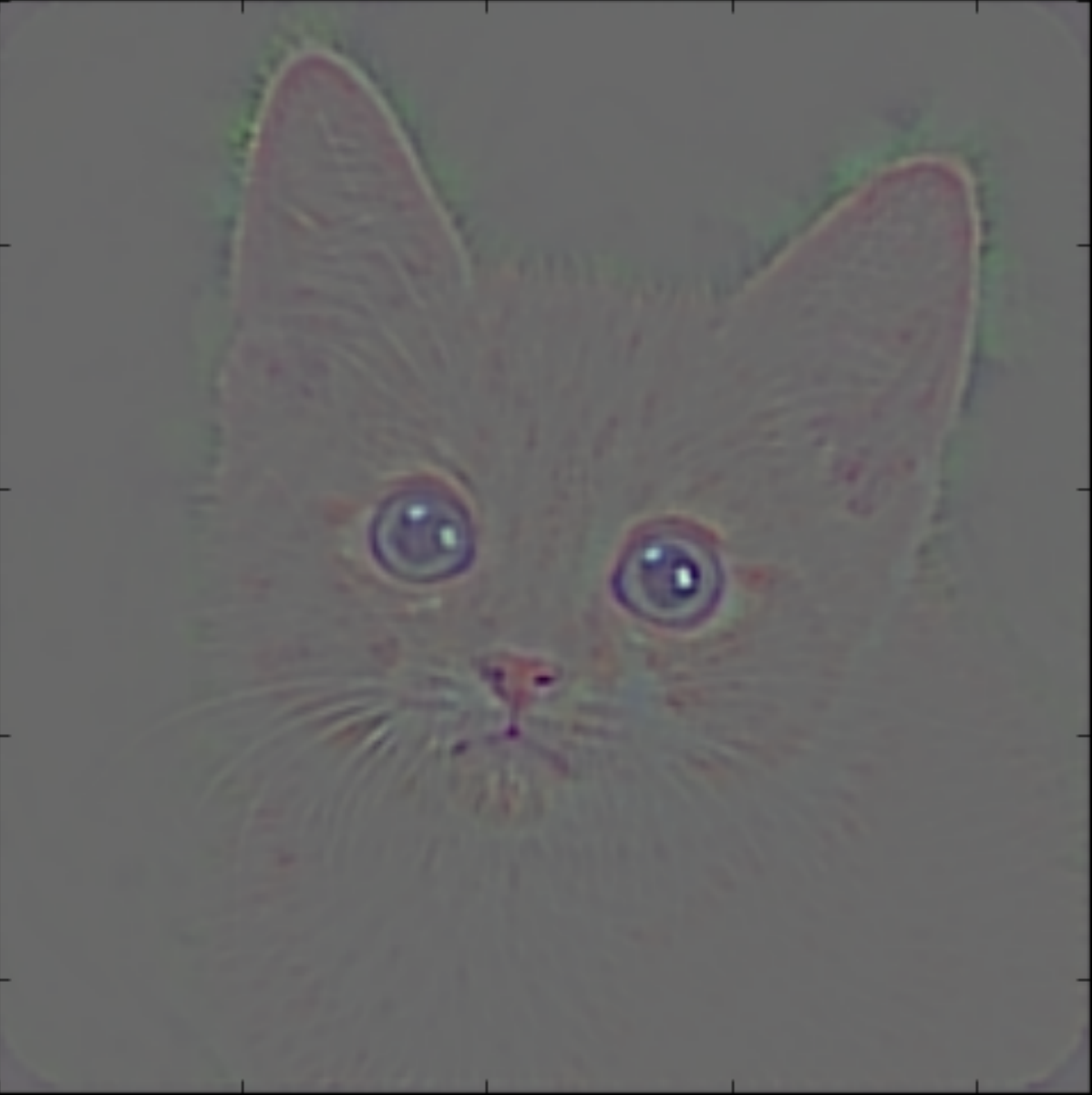}
			\caption*{FC3*}
		\end{subfigure}
		
		\begin{subfigure}[b]{0.11\textwidth}
			\centering
			\includegraphics[width=\textwidth]{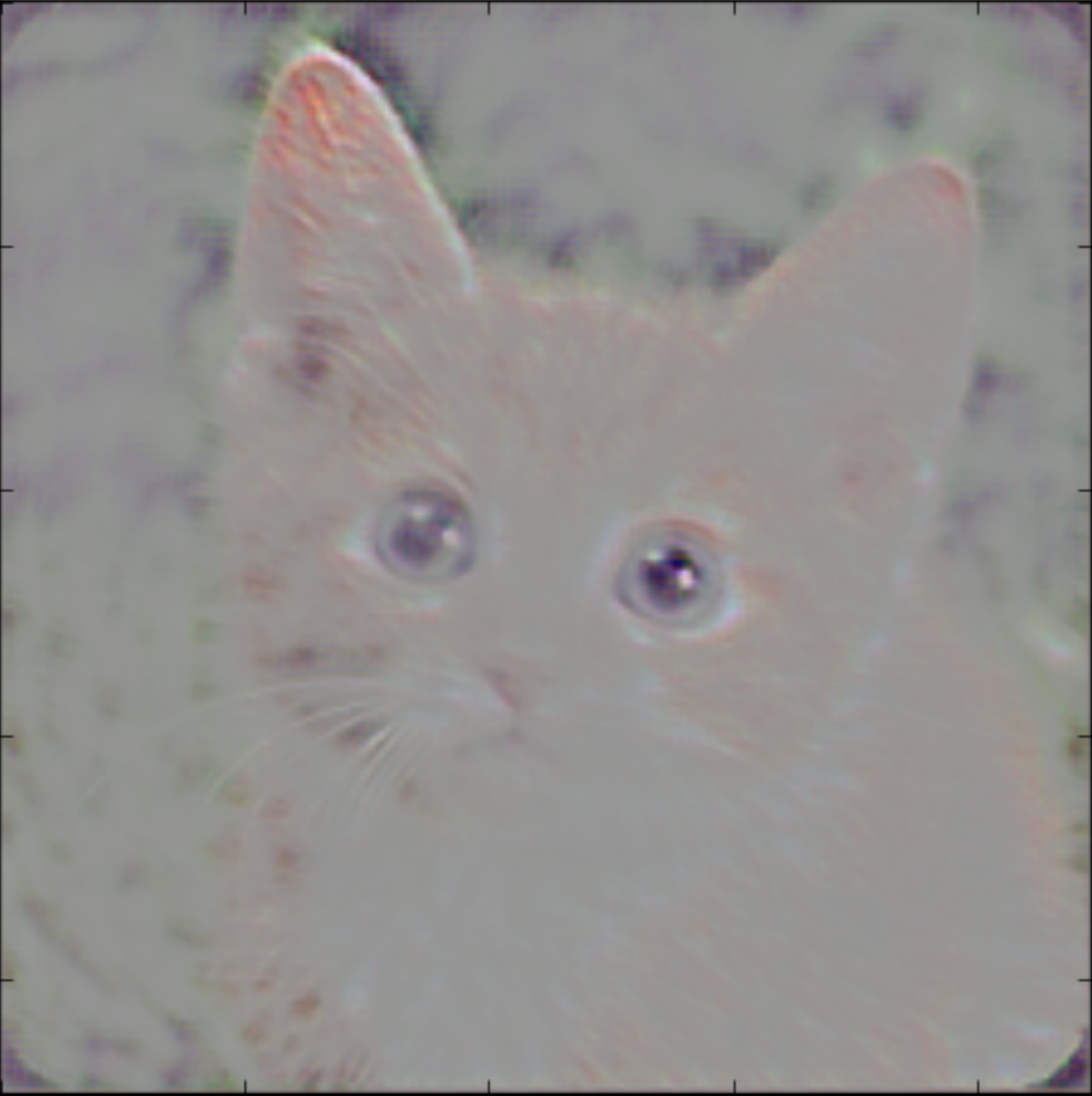}
			\caption*{Conv1-1$^\diamond$}
		\end{subfigure}
		\begin{subfigure}[b]{0.11\textwidth}
			\centering
			\includegraphics[width=\textwidth]{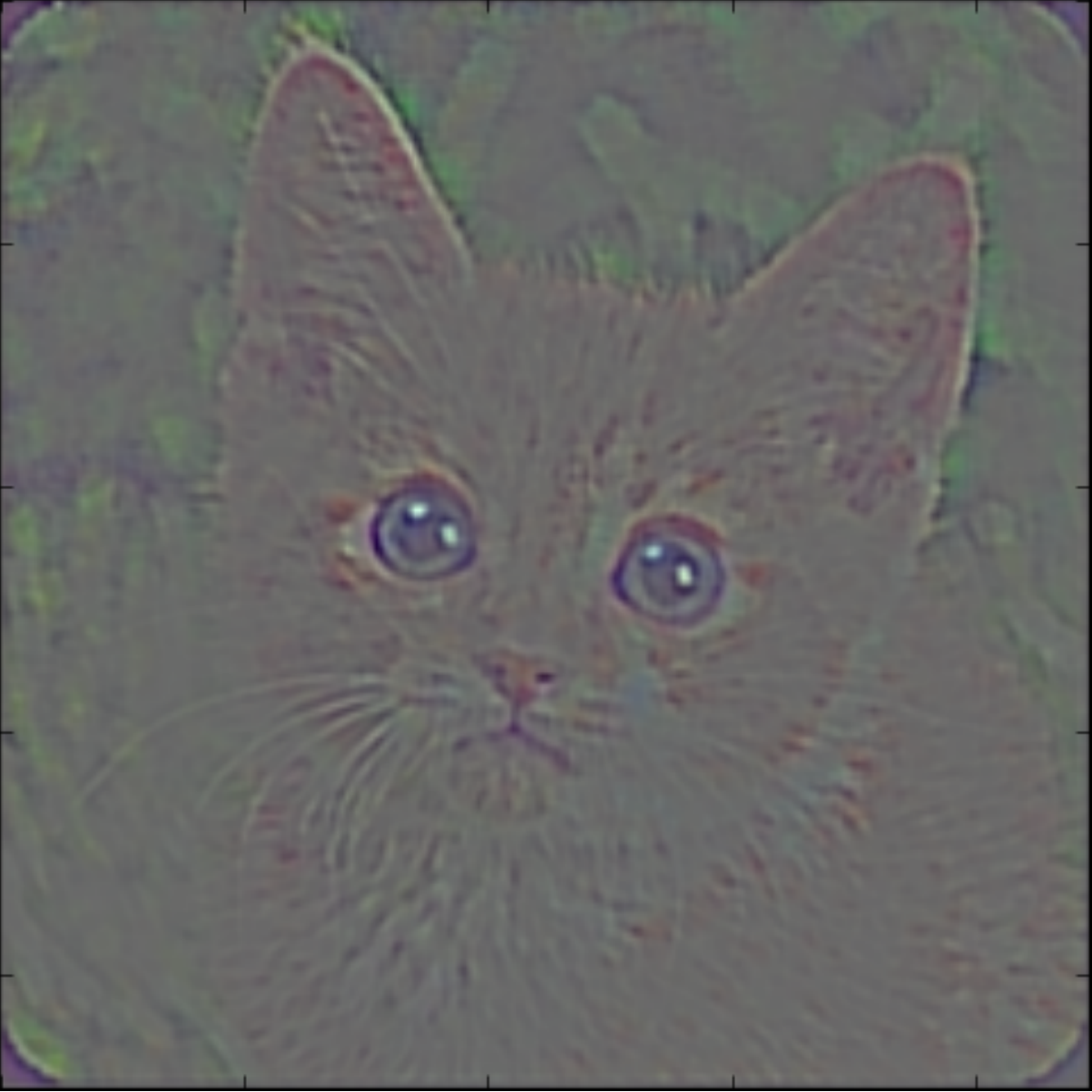}
			\caption*{Conv3-1$^\diamond$}
		\end{subfigure}
		\begin{subfigure}[b]{0.11\textwidth}
			\centering
			\includegraphics[width=\textwidth]{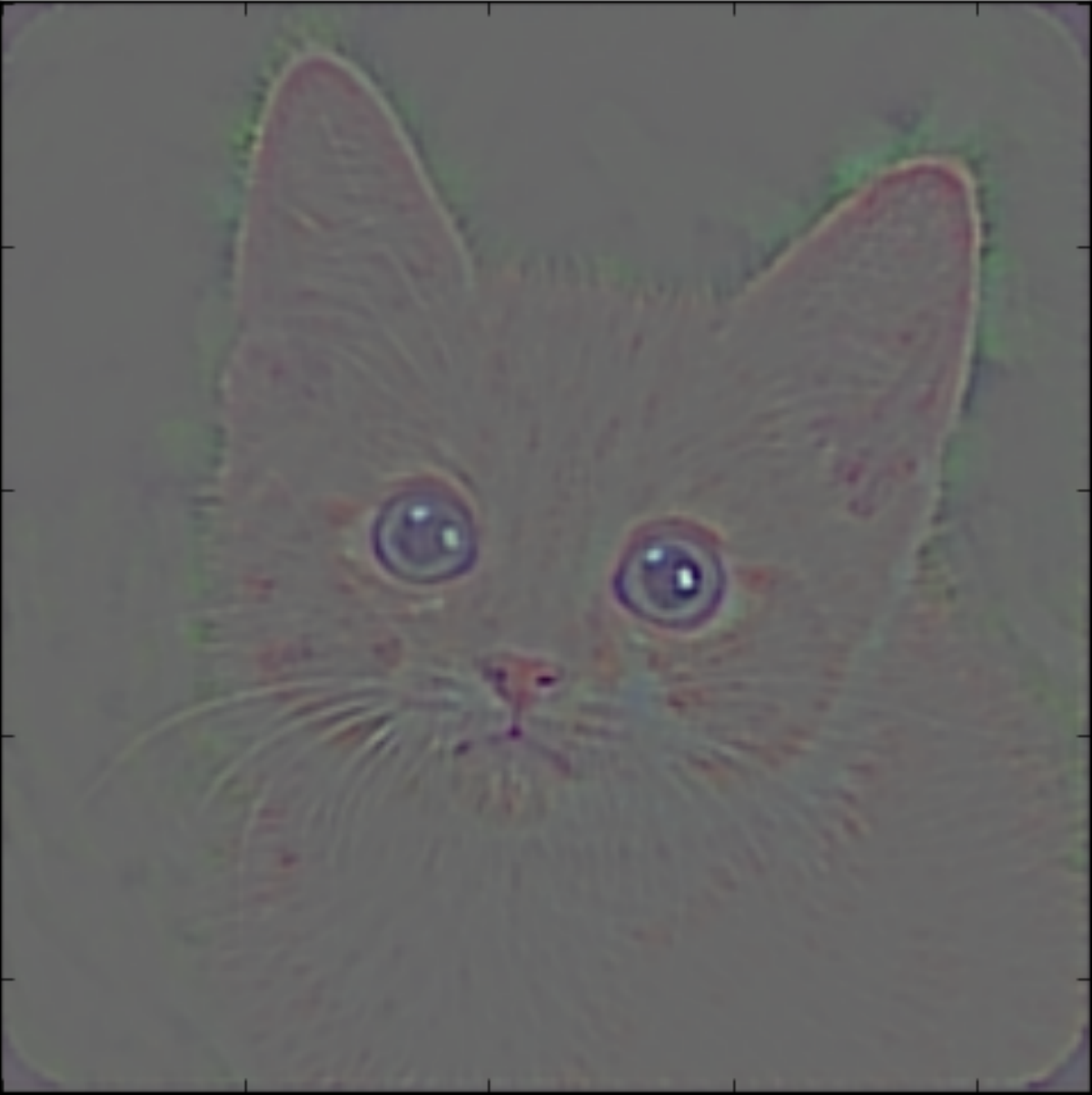}
			\caption*{Conv5-1$^\diamond$}
		\end{subfigure}
		\begin{subfigure}[b]{0.11\textwidth}
			\centering
			\includegraphics[width=\textwidth]{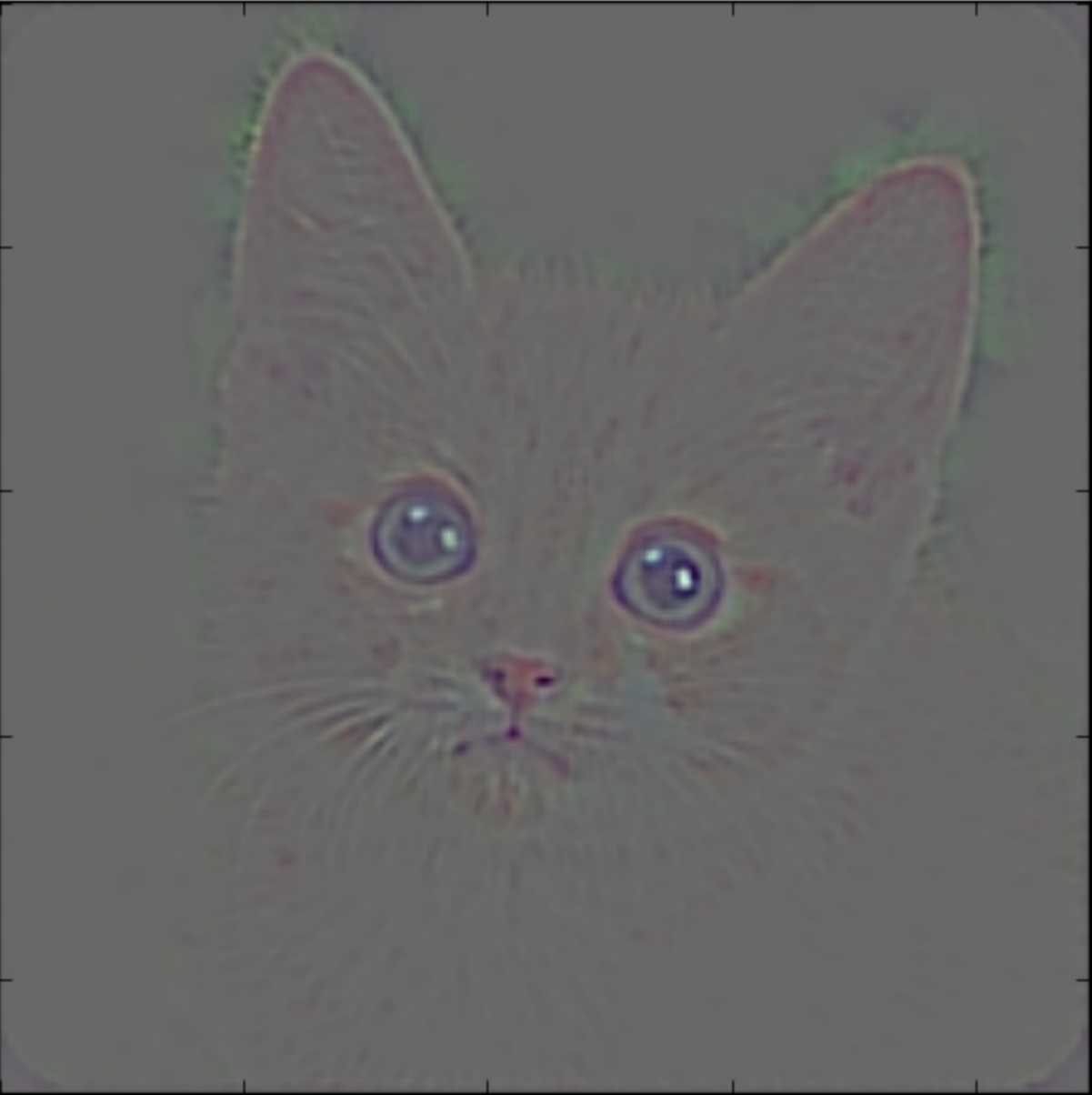}
			\caption*{FC3$^\diamond$}
		\end{subfigure}
	\caption{Top row: load trained weights  \textbf{up to} the indexed layer and leave the later layers to be randomly initialized (marked by star sign). Bottom row: load trained weights  \textbf{except for} the indexed layer is randomly initialized instead (marked by diamond sign).
	}\label{part_load_1}
	
\end{figure}

First, to isolate the impact of later layers, we load the trained weights up to a given layer and leave later layers randomly initialized. As shown in
Figure \ref{part_load_1} (top row), from ``Conv1-1*'' to ``Conv5-1*'' GBP keeps filtering out more image patches as the number of trained convolutional layers increases. 
However, from ``Conv5-1*'' to ``FC3*'' (i.e., the fully-trained case) GBP behaves almost the same, no matter weights in the dense layers are random or trained. 
Therefore, it is the trained weights in the convolutional layers rather than those in the dense layers that account for filtering out image patches. Also, it further confirms that GBP is class-insensitive. 
Furthermore, to reveal the impact of each layer, we load the trained weights for the whole VGG-16 net except for a given layer which is randomly initialized instead. The results are shown in 
Figure \ref{part_load_1} (bottom row). We can see that the GBP visualization is blurry for ``Conv1-1$^\diamond$'', clean with much background information for ``Conv3-1$^\diamond$'' and clean without background information for ``Conv5-1$^\diamond$'', respectively. It means that
the earlier convolutional layer has more important impact in the GBP visualization than the later convolutional layer.  


\section{Conclusions}
\label{others}

In this paper, we proposed a theoretical explanation for backpropagation-based visualizations, where we started from a random three-layer CNN and later generalized it to more realistic cases. 
We showed that unlike saliency map, both GBP and DeconvNet are essentially doing (partial) image recovery, which verified their class-insensitive properties. We revealed that it is the backward ReLU, used by both GBP and DeconvNet, along with the local connections in CNNs, that is responsible for human-interpretable visualizations. We also explained how DeconvNet also relies on the max-pooling to recover the input. Our analysis was supported by extensive experiments. 
Finally, we hope our analysis can 
provide useful insights into developing better visualization methods for deep neural networks. A future direction is to understand how the GBP visualizations in the trained CNNs filter out image patches layer by layer.



\section*{Acknowledgements}

Thanks to the anonymous reviewers for useful comments. WN, YZ and AB were supported by IARPA via DoI/IBC contract D16PC00003. Also, we thank Leon Sixt for pointing out an error in the proof of Theorem 1.

\bibliography{GBP_draft}

\begin{thebibliography}{25}
\providecommand{\natexlab}[1]{#1}
\providecommand{\url}[1]{\texttt{#1}}
\expandafter\ifx\csname urlstyle\endcsname\relax
  \providecommand{\doi}[1]{doi: #1}\else
  \providecommand{\doi}{doi: \begingroup \urlstyle{rm}\Url}\fi

\bibitem[Deng et~al.(2009)Deng, Dong, Socher, Li, Li, and
  Fei-Fei]{deng2009imagenet}
Deng, J., Dong, W., Socher, R., Li, L.-J., Li, K., and Fei-Fei, L.
\newblock Imagenet: A large-scale hierarchical image database.
\newblock In \emph{Computer Vision and Pattern Recognition, 2009. CVPR 2009.
  IEEE Conference on}, pp.\  248--255. IEEE, 2009.

\bibitem[Dosovitskiy \& Brox(2016)Dosovitskiy and
  Brox]{dosovitskiy2016inverting}
Dosovitskiy, A. and Brox, T.
\newblock Inverting visual representations with convolutional networks.
\newblock In \emph{Proceedings of the IEEE Conference on Computer Vision and
  Pattern Recognition}, pp.\  4829--4837, 2016.

\bibitem[Fong \& Vedaldi(2017)Fong and Vedaldi]{fong2017interpretable}
Fong, R.~C. and Vedaldi, A.
\newblock Interpretable explanations of black boxes by meaningful perturbation.
\newblock In \emph{Proceedings of the IEEE Conference on Computer Vision and
  Pattern Recognition}, pp.\  3429--3437, 2017.

\bibitem[Goodfellow et~al.(2014)Goodfellow, Shlens, and
  Szegedy]{goodfellow2014explaining}
Goodfellow, I.~J., Shlens, J., and Szegedy, C.
\newblock Explaining and harnessing adversarial examples.
\newblock \emph{arXiv preprint arXiv:1412.6572}, 2014.

\bibitem[Gunning(2017)]{gunning2017explainable}
Gunning, D.
\newblock Explainable artificial intelligence (xai).
\newblock \emph{Defense Advanced Research Projects Agency (DARPA), nd Web},
  2017.

\bibitem[He et~al.(2016)He, Zhang, Ren, and Sun]{he2016deep}
He, K., Zhang, X., Ren, S., and Sun, J.
\newblock Deep residual learning for image recognition.
\newblock In \emph{Proceedings of the IEEE conference on computer vision and
  pattern recognition}, pp.\  770--778, 2016.

\bibitem[Johnson et~al.(2016)Johnson, Alahi, and
  Fei-Fei]{johnson2016perceptual}
Johnson, J., Alahi, A., and Fei-Fei, L.
\newblock Perceptual losses for real-time style transfer and super-resolution.
\newblock In \emph{European Conference on Computer Vision}, pp.\  694--711.
  Springer, 2016.

\bibitem[Kindermans et~al.(2017)Kindermans, Sch{\"u}tt, Alber, M{\"u}ller, and
  D{\"a}hne]{kindermans2017patternnet}
Kindermans, P.-J., Sch{\"u}tt, K.~T., Alber, M., M{\"u}ller, K.-R., and
  D{\"a}hne, S.
\newblock Patternnet and patternlrp--improving the interpretability of neural
  networks.
\newblock \emph{arXiv preprint arXiv:1705.05598}, 2017.

\bibitem[Kraus et~al.(2016)Kraus, Ba, and Frey]{kraus2016classifying}
Kraus, O.~Z., Ba, J.~L., and Frey, B.~J.
\newblock Classifying and segmenting microscopy images with deep multiple
  instance learning.
\newblock \emph{Bioinformatics}, 32\penalty0 (12):\penalty0 i52--i59, 2016.

\bibitem[Krizhevsky et~al.(2012)Krizhevsky, Sutskever, and
  Hinton]{krizhevsky2012imagenet}
Krizhevsky, A., Sutskever, I., and Hinton, G.~E.
\newblock Imagenet classification with deep convolutional neural networks.
\newblock In \emph{Advances in neural information processing systems}, pp.\
  1097--1105, 2012.

\bibitem[Lugosi \& Mendelson(2017)Lugosi and Mendelson]{lugosi2017sub}
Lugosi, G. and Mendelson, S.
\newblock Sub-gaussian estimators of the mean of a random vector.
\newblock \emph{arXiv preprint arXiv:1702.00482}, 2017.

\bibitem[Mahendran \& Vedaldi(2016)Mahendran and Vedaldi]{mahendran2016salient}
Mahendran, A. and Vedaldi, A.
\newblock Salient deconvolutional networks.
\newblock In \emph{European Conference on Computer Vision}, pp.\  120--135.
  Springer, 2016.

\bibitem[Nguyen et~al.(2016)Nguyen, Dosovitskiy, Yosinski, Brox, and
  Clune]{nguyen2016synthesizing}
Nguyen, A., Dosovitskiy, A., Yosinski, J., Brox, T., and Clune, J.
\newblock Synthesizing the preferred inputs for neurons in neural networks via
  deep generator networks.
\newblock In \emph{Advances in Neural Information Processing Systems}, pp.\
  3387--3395, 2016.

\bibitem[Odena et~al.(2016)Odena, Dumoulin, and Olah]{odena2016deconvolution}
Odena, A., Dumoulin, V., and Olah, C.
\newblock Deconvolution and checkerboard artifacts.
\newblock \emph{Distill}, 2016.
\newblock \doi{10.23915/distill.00003}.
\newblock URL \url{http://distill.pub/2016/deconv-checkerboard}.

\bibitem[Samek et~al.(2017)Samek, Binder, Montavon, Lapuschkin, and
  M{\"u}ller]{samek2017evaluating}
Samek, W., Binder, A., Montavon, G., Lapuschkin, S., and M{\"u}ller, K.-R.
\newblock Evaluating the visualization of what a deep neural network has
  learned.
\newblock \emph{IEEE transactions on neural networks and learning systems},
  28\penalty0 (11):\penalty0 2660--2673, 2017.

\bibitem[Selvaraju et~al.(2016)Selvaraju, Das, Vedantam, Cogswell, Parikh, and
  Batra]{selvaraju2016grad}
Selvaraju, R.~R., Das, A., Vedantam, R., Cogswell, M., Parikh, D., and Batra,
  D.
\newblock Grad-cam: Why did you say that? visual explanations from deep
  networks via gradient-based localization.
\newblock \emph{arXiv preprint arXiv:1610.02391}, 2016.

\bibitem[Shrikumar et~al.(2017)Shrikumar, Greenside, and
  Kundaje]{shrikumar2017learning}
Shrikumar, A., Greenside, P., and Kundaje, A.
\newblock Learning important features through propagating activation
  differences.
\newblock \emph{arXiv preprint arXiv:1704.02685}, 2017.

\bibitem[Silver et~al.(2016)Silver, Huang, Maddison, Guez, Sifre, Van
  Den~Driessche, Schrittwieser, Antonoglou, Panneershelvam, Lanctot,
  et~al.]{silver2016mastering}
Silver, D., Huang, A., Maddison, C.~J., Guez, A., Sifre, L., Van Den~Driessche,
  G., Schrittwieser, J., Antonoglou, I., Panneershelvam, V., Lanctot, M.,
  et~al.
\newblock Mastering the game of go with deep neural networks and tree search.
\newblock \emph{Nature}, 529\penalty0 (7587):\penalty0 484--489, 2016.

\bibitem[Simonyan \& Zisserman(2014)Simonyan and Zisserman]{simonyan2014very}
Simonyan, K. and Zisserman, A.
\newblock Very deep convolutional networks for large-scale image recognition.
\newblock \emph{arXiv preprint arXiv:1409.1556}, 2014.

\bibitem[Simonyan et~al.(2013)Simonyan, Vedaldi, and
  Zisserman]{simonyan2013deep}
Simonyan, K., Vedaldi, A., and Zisserman, A.
\newblock Deep inside convolutional networks: Visualising image classification
  models and saliency maps.
\newblock \emph{arXiv preprint arXiv:1312.6034}, 2013.

\bibitem[Springenberg et~al.(2014)Springenberg, Dosovitskiy, Brox, and
  Riedmiller]{springenberg2014striving}
Springenberg, J.~T., Dosovitskiy, A., Brox, T., and Riedmiller, M.
\newblock Striving for simplicity: The all convolutional net.
\newblock \emph{arXiv preprint arXiv:1412.6806}, 2014.

\bibitem[Sutskever et~al.(2014)Sutskever, Vinyals, and
  Le]{sutskever2014sequence}
Sutskever, I., Vinyals, O., and Le, Q.~V.
\newblock Sequence to sequence learning with neural networks.
\newblock In \emph{Advances in neural information processing systems}, pp.\
  3104--3112, 2014.

\bibitem[Szegedy et~al.(2013)Szegedy, Zaremba, Sutskever, Bruna, Erhan,
  Goodfellow, and Fergus]{szegedy2013intriguing}
Szegedy, C., Zaremba, W., Sutskever, I., Bruna, J., Erhan, D., Goodfellow, I.,
  and Fergus, R.
\newblock Intriguing properties of neural networks.
\newblock \emph{arXiv preprint arXiv:1312.6199}, 2013.

\bibitem[Yosinski et~al.(2014)Yosinski, Clune, Bengio, and
  Lipson]{yosinski2014transferable}
Yosinski, J., Clune, J., Bengio, Y., and Lipson, H.
\newblock How transferable are features in deep neural networks?
\newblock In \emph{Advances in neural information processing systems}, pp.\
  3320--3328, 2014.

\bibitem[Zeiler \& Fergus(2014)Zeiler and Fergus]{zeiler2014visualizing}
Zeiler, M.~D. and Fergus, R.
\newblock Visualizing and understanding convolutional networks.
\newblock In \emph{European conference on computer vision}, pp.\  818--833.
  Springer, 2014.

\end{thebibliography}
\bibliographystyle{icml2018}


\newpage

\section*{\Large Appendix}

\vspace{3mm}

\appendix 

\vspace{-2mm}
\section{Proof of Lemma 1}

\emph{Proof:} Saliency map includes only forward ReLUs without backward ReLUs, whereas DeconvNet includes only backward ReLUs without forward ReLUs. GBP has both types of ReLUs. 
Also, the norm of all the visualization results will be normalized to be in the range of $[0,1]$.
Thus, by taking the (modified) derivative of $f_k(x)$ in Eq. (3) with respect to $x$ and applying the proper normalization, these backpropagation-based visualizations for the $k$-th logit can be unified as
\begin{align} \label{der_sk}
\begin{split}
& s_k(x) \mathop  = \limits^{\left( a \right)} \frac{1}{Z_k} \sum_{i=1}^{N}\sum_{j=1}^{J}{h({V_{q_{ij},k}})\frac{\partial}{\partial x}{g({w^{(i)T}} y^{(j)})}} \\
	&\mathop  = \limits^{\left( b \right)} \frac{1}{Z_k} \sum_{j=1}^{J} {D_j}^T \sum_{i=1}^{N} {h({V_{q_{ij},k}}) \frac{\partial}{\partial y^{(j)}} g({w^{(i)T}} y^{(j)})} \\
	&\mathop = \limits^{(c)} \frac{1}{Z_k} \sum_{j=1}^{J} {D_j}^T \sum_{i=1}^{N} {h({V_{q_{ij},k}}) \tilde{w}^{(i,j)}}
\end{split}
\end{align}
where $Z_k$ is the normalization coefficient to ensure $\|s_k(x)\| \in [0,1]$, $(a)$ follows from the formal definitions of backpropagation-based visualization for a ReLU activation in Eq. (1) with $h(\cdot), g(\cdot)$ being given by Eq. (2), $(b)$ is from applying $y^{(j)} = D_j x$ and swapping the two sums, 
and $(c)$ is from taking the derivative of $g(\cdot)$ in the three cases with
\begin{align*} 
    \tilde{w}^{(i,j)} = \begin{cases}
    w^{(i)} & \text{for DeconvNet}\\
    w^{(i)}\mathbb{I}\left({w^{(i)T}} y^{(j)}\right) & \text{for saliency map and GBP} 
\end{cases}
\end{align*}
as required.
\hfill $\square$

\vspace{-2mm}
\section{Proof of Theorem 1}

\emph{Proof:}
In a random neural network where every entry of both $V$ and $W$ is assumed to be independently Gaussian distributed with a zero mean and variance $c^2$, we have 
$V_{q_{ij},k} \sim \mathcal{N}(0, c^2)$ and $w^{(i)} \sim \mathcal{N}(0, c^2 I)$ $\forall i \in \{1, \cdots, N\}, j \in \{1,\cdots,J\}$. 
For GBP, in order to ensure $\|s_k(x)\| \in [0,1]$ we first set $Z_k = \tilde{Z}_k N$.
Assuming the number of filters $N$ is sufficiently large (e.g. VGG-16 net usually has $N=256$), then from Eq. (\ref{der_sk}) we have 
\begin{align*} 
\begin{split}
	s_k(x) & = \frac{1}{\tilde{Z}_k} \sum_{j=1}^{J} {D_j}^T \frac{1}{N} \sum_{i=1}^{N} {h({V_{q_{ij},k}}) \tilde{w}^{(i,j)}} \\
	& \mathop  \approx \limits^{\left( a \right)} \frac{1}{\tilde{Z}_k} \sum_{j=1}^{J} {D_j}^T \mathbb{E}\left[ {h({V_{q_{ij},k}}) \tilde{w}^{(i,j)}} \right] \\
	& \mathop  = \limits^{\left( b \right)} \frac{1}{\tilde{Z}_k} \sum_{j=1}^{J} {D_j}^T \mathbb{E}\left[ h({V_{q_{ij},k}}) \right] \mathbb{E} \left[ { \tilde{w}^{(i,j)}} \right]
\end{split}
\end{align*}
where $(a)$ follows from the asymptotic approximation of sample mean to the expectation and $(b)$ follows from the fact that $V_{q_{ij},k}$ and $w^{(i)}$ are independent. 

For GBP, we have $h({V_{q_{ij},k}}) = \sigma({V_{q_{ij},k}})$. Since we know ${V_{q_{ij},k}} \sim \mathcal{N} (0, c^2)$, then $h({V_{q_{ij},k}})$ follows one-dimensional rectified Gaussian distribution, and by its definition we can easily get $\mathbb{E}\left[ h({V_{q_{ij},k}}) \right] = \sqrt{\frac{1}{2\pi}}c$. Also, from the definition of $\tilde{w}^{(i,j)}$ for GBP, we know $\tilde{w}^{(i,j)}$ follows a $p$-dimensional rectified Gaussian distribution and its p.d.f. is 
\begin{align} \label{w_pdf}
   p(w) = \frac{1}{{(2\pi c^2)}^{\frac{p}{2}}} e^{-\frac{w^T w}{2c^2}} u({y^{(j)T}} w) + \frac{1}{2} p(w_A) \delta_A(w)
\end{align}
where the manifold $A \triangleq \{ w | {y^{(j)T}} w = 0 \}$, $p(w_A)$ is a $(p-1)$-dimensional Gaussian distribution projected from the $p$-dimensional Gaussian distribution onto the manifold $A$, with $w_A \in \mathbb{R}^{p-1}$ being the corresponding projected vector of $w$, and $u(t)$ and $\delta_A(w)$ are the unit step function and dirac delta function, respectively, i.e., 
\begin{align*}
    u(t) = \begin{cases} 1, & t > 0 \\ 0, & t \leq 0 \end{cases} \quad \text{and} \quad 
    \delta_A(w) = \begin{cases} +\infty, & w \in A \\ 0, & w \notin A \end{cases}
\end{align*}
By definition, we have 
\begin{align*}
    \int p(w_A) \delta_A(w) dw \triangleq \int p(w_A) dw_A = 1
\end{align*}
such that it satisfies $\int p(w) dw = 1$.
Accordingly, its expectation is given by
\begin{align} \label{w_exp}
\begin{split}
& \mathbb{E} \left[ { \tilde{w}^{(i,j)}} \right] \\
&= \int_{{y^{(j)T}} w > 0} \frac{w}{{(2\pi c^2)}^{\frac{p}{2}}} e^{-\frac{w^T w}{2c^2}} dw  + \int \frac{w}{2} p(w_A) \delta_A(w) dw \\
& \mathop = \limits^{(a)}  \int_{\phi_p > 0} U \phi \cdot \frac{1}{{(2\pi c^2)}^{\frac{p}{2}}} e^{-\frac{\phi^T \phi}{2c^2}} |U| d \phi \\
& \qquad\qquad + \int \frac{1}{2} U\phi p(\phi_{A_p}) \delta_{A_p}(\phi) |U| d \phi\\
& \mathop = \limits^{(b)}  U \int_{\phi_p > 0} \phi \cdot \frac{1}{{(2\pi c^2)}^{\frac{p}{2}}} e^{-\frac{\phi^T \phi}{2c^2}} d \phi \\
& \qquad\qquad + \frac{1}{2} U \int \phi p(\phi_{A_p}) \delta_{A_p}(\phi) d \phi \\
\end{split}
\end{align}
where $(a)$ follows from the change of variables $w= U \phi$ and $U$ is an unitary matrix satisfying the condition that $U^T \cdot \frac{y^{(j)}}{\| y^{(j)}\|_2} = e^{(p)}$ and $e^{(p)}$ is an unit vector with only the $p$-th entry being 1. That is, $\frac{y^{(j)}}{\| y^{(j)}\|_2}$ is the $p$-th column of $U$. Thus, ${y^{(j)T}} w = {y^{(j)T}} U \phi = e^{(p)T} \phi {\| y^{(j)}\|_2} = \phi_p {\| y^{(j)}\|_2}$ with $\phi_p$ being the $p$-th entry of $\phi$, which means ${y^{(j)T}} w > 0$ is equivalent to $\phi_p > 0$. Also, by the change of variables in the integral, we have $dw = |U| d \phi$ where $|\cdot|$ denotes the determinant of a matrix. Accordingly, we define a new manifold $A_p = \{\phi | \phi_p = 0\}$.
$(b)$ follows from $|U| = 1$ by the definition of an unitary matrix, and the swap between matrix multiplication and the integral. 

As $\phi$ is a $p$-dimensional vector, the first integral above can be evaluated at each entry, denoted by $\phi_m$, of $\phi$ separately. For $m \neq p$, we have
\begin{align*}
    \begin{split}
        & \int_{\phi_p > 0} \phi_m \cdot \frac{1}{{(2\pi c^2)}^{\frac{p}{2}}} e^{-\frac{\phi^T \phi}{2c^2}} d \phi \\
        & \mathop = \limits^{(a)} \underbrace{\int_{-\infty}^{\infty} \frac{\phi_m}{{(2\pi c^2)}^{\frac{1}{2}}} e^{-\frac{\phi_m^2}{2c^2}} d \phi_m}_{0}  \cdot \int_{0}^{\infty}  \frac{1}{{(2\pi c^2)}^{\frac{1}{2}}} e^{-\frac{\phi_p^2}{2c^2}} d \phi_p \\
        &= 0
    \end{split}
\end{align*}
where $(a)$ follows from the expansion of the multiple integral, and all of the other $p-2$ integrals over $\phi_k$ for $k \notin \{p,m\}$ are 1. For $m=p$, we have
\begin{align*}
    \begin{split}
        & \int_{\phi_p > 0} \phi_p \cdot \frac{1}{{(2\pi c^2)}^{\frac{p}{2}}} e^{-\frac{\phi^T \phi}{2c^2}} d \phi \\
        & \mathop = \limits^{(a)}  \int_{0}^{\infty} \phi_p \cdot \frac{1}{{(2\pi c^2)}^{\frac{1}{2}}} e^{-\frac{\phi_p^2}{2c^2}} d \phi_p \\
        &\mathop = \limits^{(b)} \sqrt{\frac{1}{2\pi}}c
    \end{split}
\end{align*}
where $(a)$ also follows from the expansion of the multiple integral, and all other $p-1$ integrals over $\phi_k$ for $k \neq p$ are 1; $(b)$ follows from evaluating the integral by the the change of variables $t = \frac{\phi_p^2}{2c^2}$.

The second integral in Eq. (\ref{w_exp}) can also be evaluated at each entry in the following.
First, we have 
\begin{align} \label{p_proj}
    p(\phi_{A_p}) = \frac{1}{{(2\pi c^2)}^{\frac{p-1}{2}}} e^{-\frac{\sum_{i=1}^{p-1}\phi_i^2}{2c^2}}
\end{align}
Then for $m \neq p$, we have
\begin{align*}
    & \int \phi_m p(\phi_{A_p}) \delta_{A_p}(\phi) d \phi \\
    & \mathop = \limits^{\left( a \right)} \int \phi_m \frac{1}{{(2\pi c^2)}^{\frac{p-1}{2}}} e^{-\frac{\sum_{i=1}^{p-1}\phi_i^2}{2c^2}} d\phi_1 \cdots d\phi_{p-1} \\
    &  = \int \phi_m \frac{1}{{(2\pi c^2)}^{\frac{1}{2}}} e^{-\frac{\phi_m^2}{2c^2}} d\phi_m \\
    & = 0
\end{align*}
where $(a)$ follows from the definition of $\delta_{A_p}(\phi)$. For $m=p$, we have
\begin{align*}
    & \int \phi_p p(\phi_{A_p}) \delta_{A_p}(\phi) d \phi \mathop = \limits^{\left( a \right)} 0
\end{align*}
where $(a)$ follows from $A_p = \{\phi | \phi_p = 0\}$.

Putting them together, (\ref{w_exp}) becomes
\begin{align} \label{w_exp_final}
\begin{split}
\mathbb{E} \left[ { \tilde{w}^{(i,j)}} \right] = \sqrt{\frac{1}{2\pi}}c \cdot U e^{(p)} \mathop = \limits^{(a)} \sqrt{\frac{1}{2\pi}}c \cdot \frac{y^{(j)}}{\| y^{(j)}\|_2}
\end{split}
\end{align}
where $(a)$ follows from the the definition of the unitary matrix $U$ satisfying $U^T \cdot \frac{y^{(j)}}{\| y^{(j)}\|_2} = e^{(p)}$.


Therefore, GBP at the $k$-th logit can be approximated as 
\begin{align*}
\begin{split}
s^{\text{GBP}}_k(x) &   \approx  \frac{c^2}{2\pi \tilde{Z}_k} \sum_{j=1}^{J} \frac{1}{\|y^{(j)}\|_2}{D_j}^Ty^{(j)} \\
& \mathop  = \limits^{\left( a \right)} \frac{c^2 }{2\pi \tilde{Z}_k} \left(\sum_{j=1}^{J} \frac{1}{\|y^{(j)}\|_2} \mathcal{I}_{p_j} \right) x
\end{split}
\end{align*}
where $(a)$ follows from the definition $$\mathcal{I}_{p_j} \triangleq {D_j}^T D_j = 
\begin{bmatrix}
0_{(j-1)b \times (j-1)b} & & \\
 & I_{p \times p} & \\
 & & 0 
\end{bmatrix}
\in \mathcal{R}^{d \times d}.$$ 
Ideally, if we assume ${\|y^{(j)}\|_2} = C_0, \text{ } \forall j$ (a constant) and ignore the boundary points (Note that using the ``SAME'' padding method instead of the ``VALID'' one is supposed to alleviate the boundary inconsistency to some extent), then $\sum_{j=1}^{J} \mathcal{I}_{p_j} \approx p I_{d \times d}$ and thus we can further approximate the GBP as
\begin{align*} 
	s^{\text{GBP}}_k(x) \approx  \frac{c^2 p}{2\pi C_0 \tilde{Z}_k} x 
\end{align*}
Thus, by setting the normalization coefficient $\tilde{Z}_k = \frac{2\pi C_0}{c^2 p}$, we get the result. \hfill $\square$

\section{Proof of Theorem 2}

In Eq. (4), we denote by $\Theta_j = \sum_{i=1}^{N} {h({V_{q_{ij},k}}) \tilde{w}^{(i,j)}}$, which is a sum of $N$ independent and identically distributed random variables. From the Central Limit Theorem, $\Theta_j$ is approximated as a Gaussian random variable if the number of filters $N$ is sufficiently large. Since $s_k(x)$ is a linear function of $\Theta_j$, i.e.
\begin{align} \label{s_k_theta}
    s_k(x) = \frac{1}{Z_k} \sum_{j=1}^{J} {D_j}^T \Theta_j
\end{align}
we have $s_k(x)$ can also be approximated as a Gaussian random variable for both saliency map and DeconvNet. 

In the first part of the proof, we evaluate the mean and variance of saliency map. 

Since for saliency map we know $\mathbb{E}\left[ h({V_{q_{ij},k}}) \right] = \mathbb{E}\left[ {V_{q_{ij},k}} \right] = 0$, we can evaluate the mean of $\Theta_j$ as 
\begin{align*}
\begin{split}
    \mathbb{E}\left[\Theta_j \right] &= \sum_{i=1}^{N} \mathbb{E}\left[{h({V_{q_{ij},k}}) \tilde{w}^{(i,j)}}\right] \\
    & \mathop = \limits^{(a)} \sum_{i=1}^{N} \mathbb{E}\left[{{V_{q_{ij},k}}}\right] \mathbb{E}\left[{ \tilde{w}^{(i,j)}}\right] \\
    &= 0
\end{split}
\end{align*}
where $(a)$ is from the fact that $V_{q_{ij},k}$ and ${ \tilde{w}^{(i,j)}}$ are independent. Apparently, from Eq. (\ref{s_k_theta}) we have
\begin{align*} 
    \mathbb{E} \left[s^{\text{Sal}}_k(x) \right] = 0
\end{align*}
Then to evaluate the variance of saliency map, we can also first evaluate the variance of $\Theta_j$ as 
\begin{align*}
    &\text{Var}\left[ \Theta_j \right] = N \cdot \text{Var} \left[ {h({V_{q_{ij},k}}) \tilde{w}^{(i,j)}} \right] \\
    & \mathop = \limits^{(a)} N \cdot \left\{  \text{Var}\left[ {V_{q_{ij},k}} \right] \mathbb{E}\left[ \tilde{w}^{(i,j)} \right]^2 + \right. \\ 
    & \qquad \left. \text{Var}\left[ {V_{q_{ij},k}} \right] \text{Var} \left[ \tilde{w}^{(i,j)} \right] + \text{Var} \left[ \tilde{w}^{(i,j)} \right]  \mathbb{E}\left[ {V_{q_{ij},k}} \right]^2 \right\}\\
    & \mathop = \limits^{(b)} N \cdot c^2 \mathbb{E} \left[ \tilde{w}^{(i,j)} \tilde{w}^{(i,j)T} \right]
\end{align*}
where $(a)$ is also from the fact that $V_{q_{ij},k}$ and ${ \tilde{w}^{(i,j)}}$ are independent and $(b)$ follows from $\mathbb{E}\left[ {V_{q_{ij},k}} \right] = 0$ and $\text{Var}\left[ {V_{q_{ij},k}} \right] = c^2$. 
According to Eq. (\ref{w_pdf}), we get 
\begin{align*} 
\begin{split}
& \mathbb{E} \left[ { \tilde{w}^{(i,j)}  \tilde{w}^{(i,j)T} } \right] \\ 
&= \int_{{y^{(j)T}} w > 0} w w^T \cdot \frac{1}{{(2\pi c^2)}^{\frac{p}{2}}} e^{-\frac{w^T w}{2c^2}} dw \\
& \qquad + \frac{1}{2} \int ww^T  p(w_A) \delta_A(w) dw \\
& \mathop = \limits^{(a)} \int_{\phi_p > 0} U \phi \phi^T U^T \cdot \frac{1}{{(2\pi c^2)}^{\frac{p}{2}}} e^{-\frac{\phi^T \phi}{2c^2}} |U| d \phi \\
& \qquad + \frac{1}{2} \int U \phi \phi^T U^T  p(\phi_{A_p}) \delta_{A_p}(\phi) |U| d\phi \\
& =  U \underbrace{\left[ \int_{\phi_p > 0} \phi \phi^T \cdot \frac{1}{{(2\pi c^2)}^{\frac{p}{2}}} e^{-\frac{\phi^T \phi}{2c^2}} d \phi \right]}_{B^{(0)}} U^T \\
& \qquad + \frac{1}{2} U \underbrace{\left[ \int \phi \phi^T p(\phi_{A_p}) \delta_{A_p}(\phi) d\phi \right]}_{B^{(1)}} U^T \\
\end{split}
\end{align*}
where $(a)$ also follows from the change of variables $w= U \phi$ and $U$ is an unitary matrix satisfying the condition that $U^T \cdot \frac{y^{(j)}}{\| y^{(j)}\|_2} = e^{(p)}$, and $A_p = \{\phi | \phi_p = 0\}$.

As $\phi \phi^T$ is a $p \times p$ matrix, the integral above $B^{(0)}$ can be evaluated at each entry, denoted by $\phi_m \phi_n$, of $\phi \phi^T$ separately where $m,n \in \{1,\cdots, p\}$. 

First, for $m \neq n \neq p$, we have
\begin{align*}
    \begin{split}
        & B^{(0)}_{mn} = \int_{\phi_p > 0} \phi_m \phi_n \cdot \frac{1}{{(2\pi c^2)}^{\frac{p}{2}}} e^{-\frac{\phi^T \phi}{2c^2}} d \phi \\
        & \mathop = \limits^{(a)}  \underbrace{\int_{-\infty}^{\infty} \frac{\phi_m}{{(2\pi c^2)}^{\frac{1}{2}}} e^{-\frac{\phi_m^2}{2c^2}} d \phi_m}_{0}  \cdot \underbrace{\int_{-\infty}^{\infty}  \frac{\phi_n}{{(2\pi c^2)}^{\frac{1}{2}}} e^{-\frac{\phi_n^2}{2c^2}} d \phi_n}_{0} \\
        &= 0 
    \end{split}
\end{align*}
where $(a)$ follows from the expansion of the multiple integral, and all the other $p-2$ integrals over $\phi_k$ for $k \notin \{m,n\}$ are 1. Similarly, we can easily get that $B^{(0)}_{mn} = 0$ for $m \neq n$ with $m = p$ or $n = p$. 

Also, for $m = n \neq p$, we have 
\begin{align*}
    \begin{split}
        &B^{(0)}_{mn} = \int_{\phi_p > 0} \phi_m^2 \cdot \frac{1}{{(2\pi c^2)}^{\frac{p}{2}}} e^{-\frac{\phi^T \phi}{2c^2}} d \phi \\
        & \mathop = \limits^{(a)}  \underbrace{\int_{-\infty}^{\infty} \frac{\phi_m^2}{{(2\pi c^2)}^{\frac{1}{2}}} e^{-\frac{\phi_m^2}{2c^2}} d \phi_m}_{c^2}  \cdot \underbrace{\int_{0}^{\infty}  \frac{1}{{(2\pi c^2)}^{\frac{1}{2}}} e^{-\frac{\phi_p^2}{2c^2}} d \phi_p}_{\frac{1}{2}} \\
        &= \frac{1}{2} c^2
    \end{split}
\end{align*}
where $(a)$ follows from the expansion of the multiple integral, and all other $p-2$ integrals are 1. 

Finally, for $m = n = p$, we have 
\begin{align*}
    \begin{split}
        B^{(0)}_{mn} &= \int_{\phi_p > 0} \phi_p^2 \cdot \frac{1}{{(2\pi c^2)}^{\frac{p}{2}}} e^{-\frac{\phi^T \phi}{2c^2}} d \phi \\
        & \mathop = \limits^{(a)}  \underbrace{\int_{0}^{\infty} \frac{\phi_p^2}{{(2\pi c^2)}^{\frac{1}{2}}} e^{-\frac{\phi_p^2}{2c^2}} d \phi_p}_{\frac{1}{2}c^2}  \\
        &= \frac{1}{2}c^2
    \end{split}
\end{align*}
where $(a)$ follows from the expansion of the multiple integral, and all other $p-1$ integrals are 1. 

Putting them together, we get $B^{(0)} = \frac{1}{2}c^2 I$. Similarly, we can evaluate the integral $B^{(1)}$ in the following. For for $m = n \neq p$, we have 
\begin{align*}
    \begin{split}
    B^{(1)}_{mn} &= \int \phi_m \phi_n p(\phi_{A_p}) \delta_{A_p}(\phi) d \phi \\
    & \mathop = \limits^{\left( a \right)} \int \phi_m \phi_n \frac{1}{{(2\pi c^2)}^{\frac{p-1}{2}}} e^{-\frac{\sum_{i=1}^{p-1}\phi_i^2}{2c^2}} d\phi_1 \cdots d\phi_{p-1} \\
        &= 0 
    \end{split}
\end{align*}
where $(a)$ is from the definition of $\delta_{A_p}(\phi)$. For $m = n \neq p$, we have 
\begin{align*}
    \begin{split}
        B^{(1)}_{mn} & = \int \phi_m^2 p(\phi_{A_p}) \delta_{A_p}(\phi) d \phi \\
    & = \int \phi_m^2 \frac{1}{{(2\pi c^2)}^{\frac{p-1}{2}}} e^{-\frac{\sum_{i=1}^{p-1}\phi_i^2}{2c^2}} d\phi_1 \cdots d\phi_{p-1} \\
        &= c^2
    \end{split}
\end{align*}
Finally, for $m = p$ (or $n = p$), we have 
\begin{align*}
    \begin{split}
        B^{(1)}_{mn} &= \int \phi_p \phi_n p(\phi_{A_p}) \delta_{A_p}(\phi) d \phi \mathop = \limits^{\left( a \right)} 0 
    \end{split}
\end{align*}
where $(a)$ follows from $A_p = \{\phi | \phi_p = 0\}$.

Putting them together, we have $B^{(1)} = c^2 (I - e_p e_p^T)$, and thus $\mathbb{E} \left[ { \tilde{w}^{(i,j)}  \tilde{w}^{(i,j)T} } \right] = c^2 U (I- \frac{1}{2}e_p e_p^T) U^T$ which further implies 
\begin{align*}
    \begin{split}
        \text{Var}\left[ \Theta_j \right] \mathop = \limits^{\left( a \right)} N c^4 \left(I -  \frac{y^{(j)}y^{(j)T}}{2 \| y^{(j)} \|^2 } \right) 
    \end{split}
\end{align*}
where $(a)$ is from $U e^{(p)} = \frac{y^{(j)}}{\| y^{(j)}\|_2} $. Accordingly, from Eq. (\ref{s_k_theta}) we have
\begin{align*}
    \text{Var}\left[ s_k^{\text{sal}} (x) \right] &= \frac{1}{Z_k^2} \sum_{j=1}^{J}{D_j^T \text{Var}[\Theta_j] D_j}\\
    &= \frac{N c^4}{Z_k^2} {\sum_{j=1}^{J}\left({D_j^T D_j} - \frac{D_j^T y^{(j)} y^{(j)T} D_j}{2 \| y^{(j)} \|^2} \right)} \\
    & \mathop \approx \limits^{(a)} \frac{Np c^4}{Z_k^2} \left( I - \frac{1}{2p} \underbrace{ \sum_{j=1}^{J} \frac{D_j^T y^{(j)} y^{(j)T} D_j}{\| y^{(j)} \|^2}}_{\Lambda} \right) \\
    & \mathop = \limits^{(b)} I - \frac{1}{2p} \Lambda
\end{align*}
where $(a)$ is from the approximation that the patching matrix $D_j$ satisfies $\sum_{j=1}^{J}{D_j^T D_j} \approx p I$, and $(b)$ follows from setting the normalization coefficient to be $Z_k = c^2\sqrt{Np}$. 
Therefore, we have 
\begin{align*}
    s_k^{\text{Sal}} \sim \mathcal{N}(0, I - \frac{1}{2p} \Lambda)
\end{align*}
where $\Lambda \triangleq \sum_{j=1}^{J} \frac{D_j^T y^{(j)} y^{(j)T} D_j}{\| y^{(j)} \|^2}$ includes image information that is `buried' in the noise. Let $\tilde{y}^{(j)} \triangleq D_j^T y^{(j)}$, a vectorized image patch, augmented with zeros representing all the other pixels. Then we have
\begin{align*}
    \Lambda = \sum_{j=1}^{J} \frac{\tilde{y}^{(j)} \tilde{y}^{(j)T}}{\| \tilde{y}^{(j)} \|^2}
\end{align*}
As we can see that $\text{Trace}(\Lambda) = J$, we have $\text{Trace}( I - \frac{1}{2p} \Lambda) = d - \frac{J}{2p}$. Since the input image dimension $d \gg \frac{J}{2p}$ (which works for a typical image patch size of $3 \times 3$ or $7 \times 7$), we have $\text{Trace}( I - \frac{1}{2p} \Lambda) \approx d$. That is, the identity term $I$ dominates the term $\frac{1}{2p} \Lambda$ with image information. 
This implies that for an input image with reasonably large dimension $d$, we approximately have
\begin{align*}
    s_k^{\text{Sal}} \mathop \sim \limits^{.} \mathcal{N}(0, I)
\end{align*}

In the second part of the proof, we evaluate the mean and variance of the DeconvNet. 

Similarly, for DeconvNet we have $\mathbb{E}\left[ \tilde{w}^{(i,j)} \right] = \mathbb{E} \left[ w^{(i)}\right] = 0$, 
then we can evaluate the mean of $\Theta_j$ as 
\begin{align*}
    \mathbb{E}\left[\Theta_j \right] &= \sum_{i=1}^{N} \mathbb{E}\left[{h({V_{q_{ij},k}}) \tilde{w}^{(i,j)}}\right] \\
    & \mathop = \limits^{(a)} \sum_{i=1}^{N} \mathbb{E}\left[{\sigma({V_{q_{ij},k}})}\right] \mathbb{E}\left[{ {w}^{(i)}}\right] \\
    &= 0
\end{align*}
where $(a)$ is from the fact that $V_{q_{ij},k}$ and ${ \tilde{w}^{(i,j)}}$ are independent. Apparently, from Eq. (\ref{s_k_theta}) we have
\begin{align*} 
    \mathbb{E} \left[s^{\text{Deconv}}_k(x) \right] = 0
\end{align*}
Then to evaluate the variance of DeconvNet, we can also first evaluate the variance of
$\Theta_j$ as 
\begin{align*}
    &\text{Var}\left[ \Theta_j \right] = N \cdot \text{Var} \left[ {\sigma({V_{q_{ij},k}}) {w}^{(i)}} \right] \\
    & \mathop = \limits^{(a)} N \cdot \left\{  \text{Var}\left[ \sigma({V_{q_{ij},k}}) \right] \text{Var} \left[ {w}^{(i)} \right] + \right. \\ 
    & \;\;\; \left. \text{Var}\left[ {\sigma (V_{q_{ij},k})} \right] \mathbb{E}\left[ {w}^{(i)} \right]^2 + \text{Var} \left[ {w}^{(i)} \right]  \mathbb{E}\left[ \sigma({V_{q_{ij},k}}) \right]^2 \right\}\\
    & \mathop = \limits^{(b)} N c^2 \mathbb{E} \left[ \sigma(V_{q_{ij},k})^2 \right] \cdot I \\
    & \mathop = \limits^{(c)} N c^4 I
\end{align*}
where $(a)$ is also from the fact that $\sigma(V_{q_{ij},k})$ and ${ {w}^{(i)}}$ are independent, $(b)$ follows from $\mathbb{E}\left[ {V_{q_{ij},k}} \right] = 0$ and $\text{Var}\left[ w^{(i)} \right] = c^2 I$ and $(c)$ follows from the fact that $\mathbb{E} \left[ \sigma(V_{q_{ij},k})^2 \right] = c^2$. 
Then, the rest of the proof follows the same derivations with saliency map, which yields
\begin{align*}
    s_k^{\text{Deconv}} \sim \mathcal{N}(0, I)
\end{align*}
Thus, we finish our proof by showing that both saliency map and DeconvNet are standard Gaussians which preserve no input information. \hfill $\square$

\section{Proof of Proposition 1}

First, let us focus on the GBP case. From Eq. (9), we know 
\begin{align} \label{GBP_V2}
    \hat{V}_{\cdot,k}^{(2)} \triangleq \frac{\partial {o}^{(3)}}{\partial o^{(2)}}  \cdots  \sigma \left( \frac{\partial {o}^{(L-1)}}{\partial o^{(L-2)}} \sigma \left(\Gamma^{(L)}_k \right) \right)
\end{align}
where $\hat{V}_{q_{ij},k}^{(2)} \in \mathbb{R}^{d_3 \times 1}$ as we know $\Gamma^{(l)} \in \mathbb{R}^{d_{l} \times d_{l+1}}$ in the $l$-th layer, and then $\hat{V}_{\cdot,k}^{(1)}$ for GBP becomes
\begin{align*} 
    \begin{split}
        \hat{V}_{\cdot,k}^{(1)} &= \frac{\partial \sigma( {\Gamma^{(2)T}} o^{(1)} )}{\partial o^{(1)}} \sigma(\hat{V}_{\cdot,k}^{(2)}) \\
        =& \begin{bmatrix}
            \Gamma^{(2)}_{\cdot,1} \mathbb{I}({\Gamma^{(2)T}_{\cdot,1}} o^{(1)}) & \cdots & \Gamma^{(2)}_{\cdot, d_3} \mathbb{I}({\Gamma^{(2)T}_{\cdot,d_3}} o^{(1)})
        \end{bmatrix} \sigma(\hat{V}_{\cdot,k}^{(2)}) \\
        =& \sum_{t=1}^{d_3} \Gamma^{(2)}_{\cdot,t} \mathbb{I}({\Gamma^{(2)T}_{\cdot,t}} o^{(1)}) \sigma(\hat{V}_{t,k}^{(2)})
    \end{split}
\end{align*}
which yields
\begin{align*} 
    \hat{V}_{q_{ij},k}^{(1)} = \sum_{t=1}^{d_3} \Gamma^{(2)}_{q_{ij},t} \mathbb{I}({\Gamma^{(2)T}_{\cdot,t}} o^{(1)}) \sigma(\hat{V}_{t,k}^{(2)})
\end{align*}

Since every entry of $\Gamma^{(2)}$ is \textit{i.i.d.} Gaussian distributed with zero-mean, we have
\begin{align*}
     \mathbb{I}({\Gamma^{(2)T}_{\cdot,t}} o^{(1)}) 
    &=  \mathbb{I}(\Gamma^{(2)}_{q_{ij},t} o^{(1)}_{q_{ij}} + \sum_{v \neq q_{ij}} \Gamma^{(2)}_{v,t} o^{(1)}_v ) \\
    &\mathop  \approx \limits^{(a)}   \underbrace{ \mathbb{I} ( \sum_{v \neq q_{ij}} \Gamma^{(2)}_{v,t} o^{(1)}_v ) }_{b_t}
\end{align*}
where $(a)$ follows from the assumption of the dimension $d_2$ is sufficiently high in the CNN, and thus the impact of $\Gamma^{(2)}_{q_{ij},t} o^{(1)}_{q_{ij}}$ can be ignored. Therefore, $b_t$ is independent of $\Gamma^{(2)}_{q_{ij},t}$. Similarly, $\sigma(\hat{V}_{t,k}^{(2)})$ is also independent of $\Gamma^{(2)}_{q_{ij},t}$ under the same approximation. 

As $d_3$ is also sufficiently large and $\hat{V}_{q_{ij},k}^{(1)}$ for GBP becomes
\begin{align} 
    \hat{V}_{q_{ij},k}^{(1)} \approx \sum_{t=1}^{d_3} \Gamma^{(2)}_{q_{ij},t} b_t \sigma(\hat{V}_{t,k}^{(2)})
\end{align}

which approximately is a Gaussian random variable with zero mean due to the central limit theorem. 
Next, in order to show the independence of two Gaussian random variables, it is equivalent to show they are uncorrelated.
Since for any $q' \neq q_{ij}$, we know 
\begin{align*}
    & \mathbb{E} \left[\hat{V}_{q',k}^{(1)} \hat{V}_{q_{ij},k}^{(1)} \right] \\
    & \approx \mathbb{E} \left[
    \sum_{t'=1}^{d_3} \Gamma^{(2)}_{q',t'} b_{t'} \sigma(\hat{V}_{t',k}^{(2)}) \sum_{t=1}^{d_3} \Gamma^{(2)}_{q_{ij},t} b_t \sigma(\hat{V}_{t,k}^{(2)}) \right] \\
    &\mathop = \limits^{(a)} \sum_{t'=1}^{d_3} \sum_{t=1}^{d_3} \mathbb{E} \left[\Gamma^{(2)}_{q',t'} \Gamma^{(2)}_{q_{ij},t} \right]
    \mathbb{E} \left[b_{t'} b_t \right]
    \mathbb{E} \left[ \sigma(\hat{V}_{t',k}^{(2)}) \sigma(\hat{V}_{t,k}^{(2)}) \right]\\
    &\mathop = \limits^{(b)} 0
\end{align*}
where $(a)$ is from the mutual independence of $b_t$, $\Gamma^{(2)}_{q_{ij},t}$ and $\sigma(\hat{V}_{t,k}^{(2)})$, and $(b)$ is from the independence of two \textit{i.i.d.} zero-mean Gaussians $\Gamma^{(2)}_{q',t'}$ and $\Gamma^{(2)}_{q_{ij},t}$, by our assumption. 
Therefore, $\hat{V}_{q',k}^{(1)}$ and $\hat{V}_{q_{ij},k}^{(1)}$ are uncorrelated with each other for any $q' \neq q_{ij}$, as desired. 

Second, we consider the saliency map and DeconvNet cases. As $\hat{V}_{q_{ij},k}^{(1)}$ for saliency map becomes
\begin{align} \label{V_sal}
    \hat{V}_{q_{ij},k}^{(1)} \approx \sum_{t=1}^{d_3} \Gamma^{(2)}_{q_{ij},t} b_t \hat{V}_{t,k}^{(2)}
\end{align}
where $\hat{V}_{t,k}^{(2)}$ comes from the definition
\begin{align*} 
    \begin{split}
        \hat{V}_{\cdot, k}^{(2)} = \frac{\partial {{o}}^{(3)}}{\partial o^{(2)}} \cdots \frac{\partial {{o}}^{(L-1)}}{\partial o^{(L-2)}} \cdot \Gamma^{(L)}_k
    \end{split}
\end{align*}
which is also approximately independent of $\Gamma^{(2)}_{q_{ij},t}$ as before, and the other parameters are exactly the same with the GBP case, the independence approximation also holds for saliency map.

For DeconvNet, $\hat{V}_{q_{ij},k}^{(1)}$ becomes
\begin{align} \label{V_deconv}
    \hat{V}_{q_{ij},k}^{(1)} \approx \sum_{t=1}^{d_3} \Gamma^{(2)}_{q_{ij},t} \hat{V}_{t,k}^{(2)}
\end{align}
where $\hat{V}_{t,k}^{(2)}$ is defined identically as (\ref{GBP_V2}) for GBP. Since this is a special case of GBP, the analysis in the case trivially holds for DeconvNet as well. 
\hfill $\square$

\section{More Experiments on Random/Trained VGG-16 Net}

\begin{figure*}[!h]
	\begin{minipage}[c]{0.48\textwidth}
	\centering
	\begin{subfigure}[b]{0.3\textwidth}
		\centering
		\includegraphics[width=\textwidth]{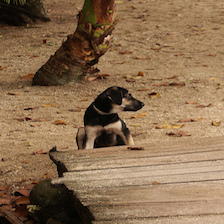}
		\caption*{\footnotesize dog}
	\end{subfigure}
	
	\begin{subfigure}[b]{0.3\textwidth}
		\centering
		\includegraphics[width=\textwidth]{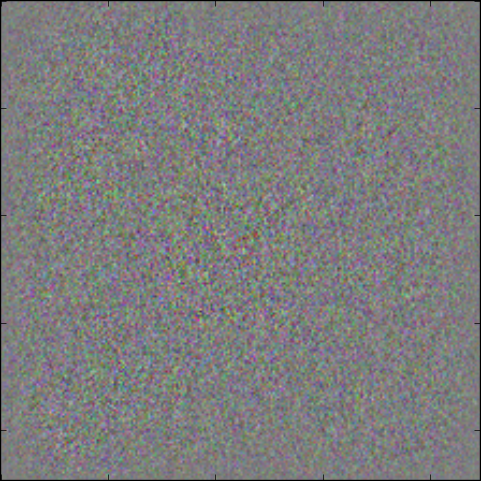}
		\caption*{\footnotesize Sal-max}
	\end{subfigure}
	\begin{subfigure}[b]{0.3\textwidth}
		\centering
		\includegraphics[width=\textwidth]{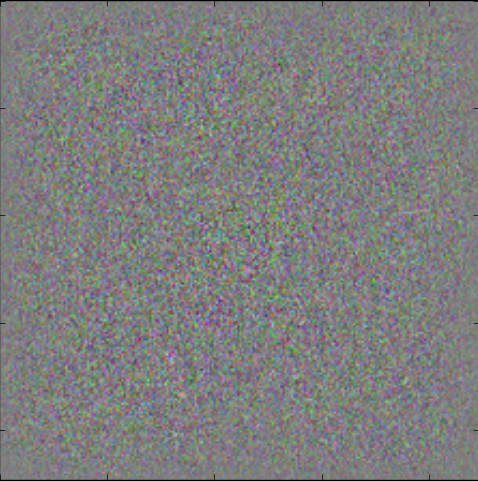}
		\caption*{\footnotesize Sal-731}
	\end{subfigure}
	\begin{subfigure}[b]{0.3\textwidth}
		\centering
		\includegraphics[width=\textwidth]{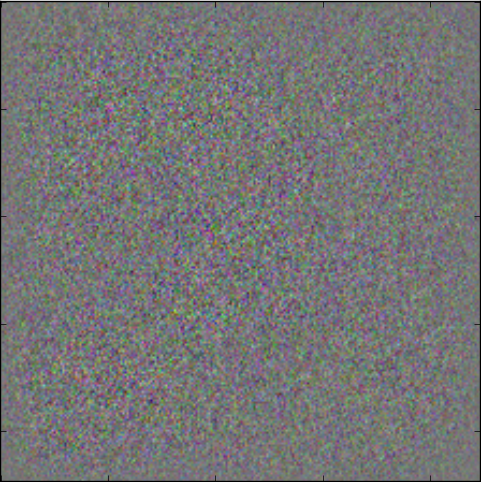}
		\caption*{\footnotesize Sal-815}
	\end{subfigure}
    
    \begin{subfigure}[b]{0.3\textwidth}
		\centering
		\includegraphics[width=\textwidth]{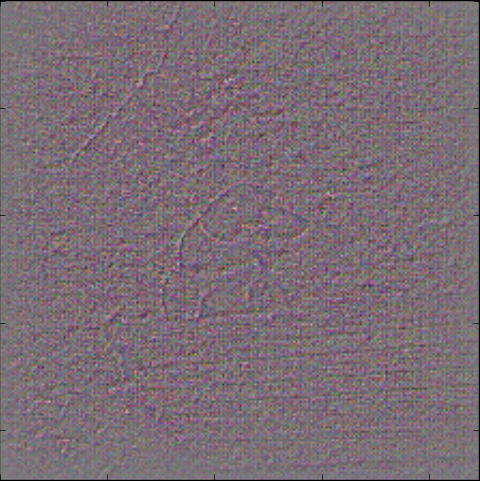}
		\caption*{\footnotesize Deconv-max}
	\end{subfigure}
	\begin{subfigure}[b]{0.3\textwidth}
		\centering
		\includegraphics[width=\textwidth]{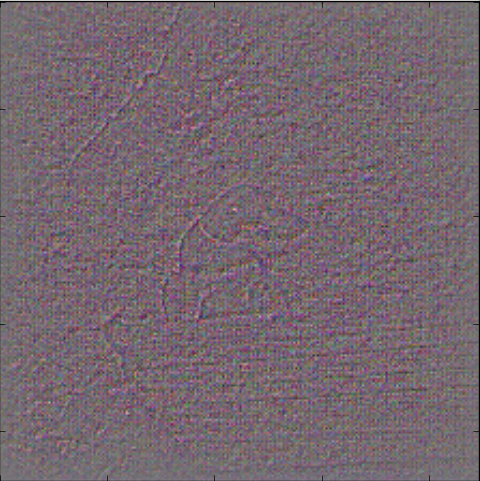}
		\caption*{\footnotesize Deconv-731}
	\end{subfigure}
	\begin{subfigure}[b]{0.3\textwidth}
		\centering
		\includegraphics[width=\textwidth]{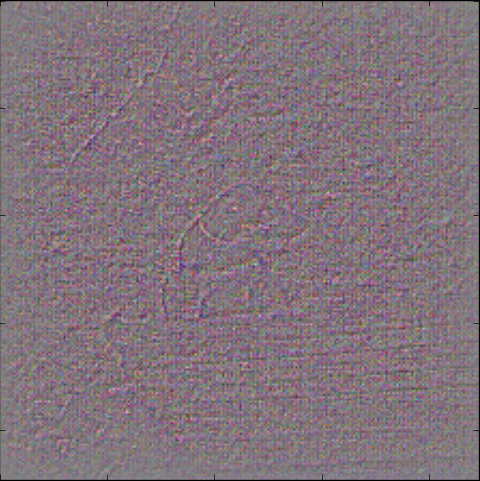}
		\caption*{\footnotesize Deconv-815}
	\end{subfigure}

	\begin{subfigure}[b]{0.3\textwidth}
		\centering
		\includegraphics[width=\textwidth]{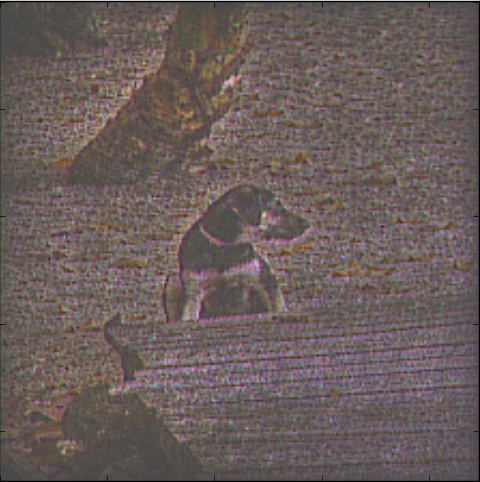}
		\caption*{\footnotesize GBP-max}
	\end{subfigure}
	\begin{subfigure}[b]{0.3\textwidth}
		\centering
		\includegraphics[width=\textwidth]{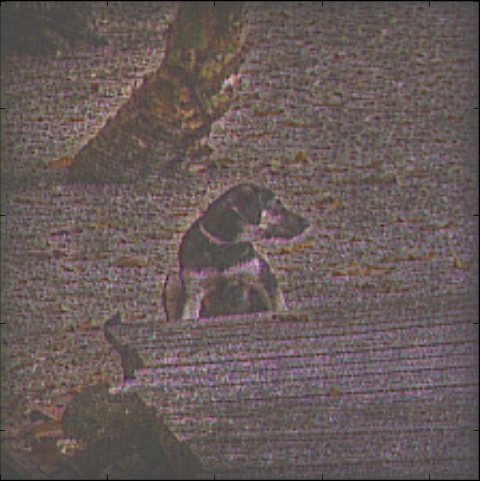}
		\caption*{\footnotesize GBP-731}
	\end{subfigure}
	\begin{subfigure}[b]{0.3\textwidth}
		\centering
		\includegraphics[width=\textwidth]{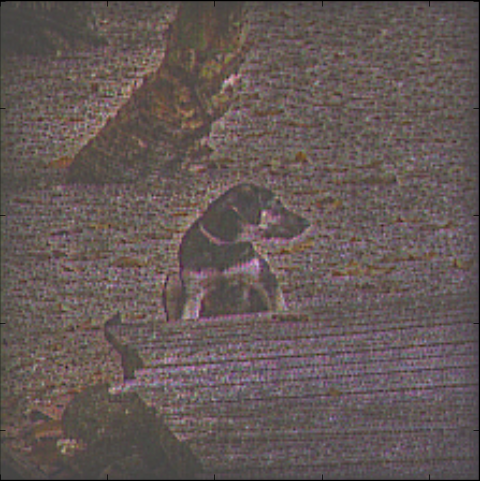}
		\caption*{\footnotesize GBP-815}
	\end{subfigure}
	
	\caption{Saliency map, DeconvNet and GBP visualizations for the random VGG-16 net with the input image ``dog''.}\label{fig_1}
	\end{minipage}
	\hfill
	\begin{minipage}[c]{0.48\textwidth}
	\centering
	\begin{subfigure}[b]{0.3\textwidth}
		\centering
		\includegraphics[width=\textwidth]{figures/01312018_1/Dog_1.png}
		\caption*{\footnotesize dog}
	\end{subfigure}
	
	\begin{subfigure}[b]{0.3\textwidth}
		\centering
		\includegraphics[width=\textwidth]{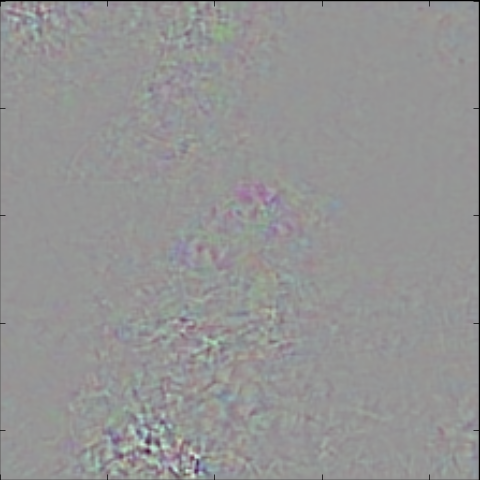}
		\caption*{\footnotesize Sal-max}
	\end{subfigure}
	\begin{subfigure}[b]{0.3\textwidth}
		\centering
		\includegraphics[width=\textwidth]{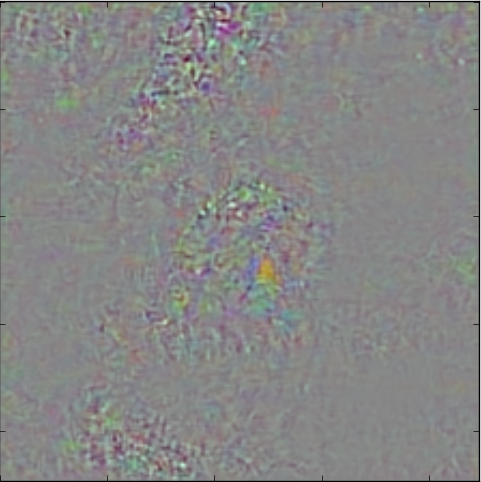}
		\caption*{\footnotesize Sal-731}
	\end{subfigure}
	\begin{subfigure}[b]{0.3\textwidth}
		\centering
		\includegraphics[width=\textwidth]{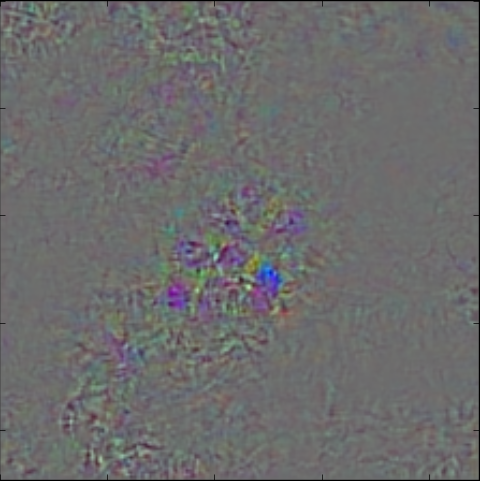}
		\caption*{\footnotesize Sal-815}
	\end{subfigure}
    
    \begin{subfigure}[b]{0.3\textwidth}
		\centering
		\includegraphics[width=\textwidth]{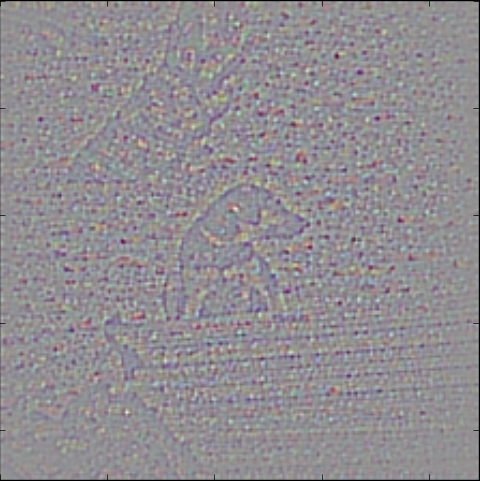}
		\caption*{\footnotesize Deconv-max}
	\end{subfigure}
	\begin{subfigure}[b]{0.3\textwidth}
		\centering
		\includegraphics[width=\textwidth]{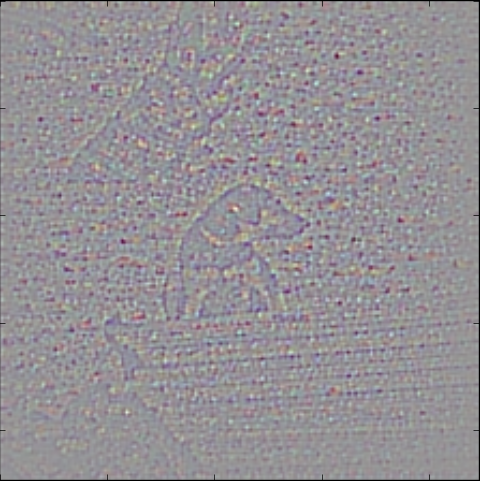}
		\caption*{\footnotesize Deconv-731}
	\end{subfigure}
	\begin{subfigure}[b]{0.3\textwidth}
		\centering
		\includegraphics[width=\textwidth]{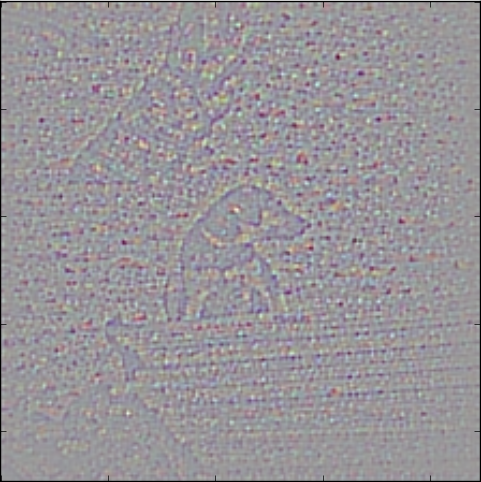}
		\caption*{\footnotesize Deconv-815}
	\end{subfigure}

	\begin{subfigure}[b]{0.3\textwidth}
		\centering
		\includegraphics[width=\textwidth]{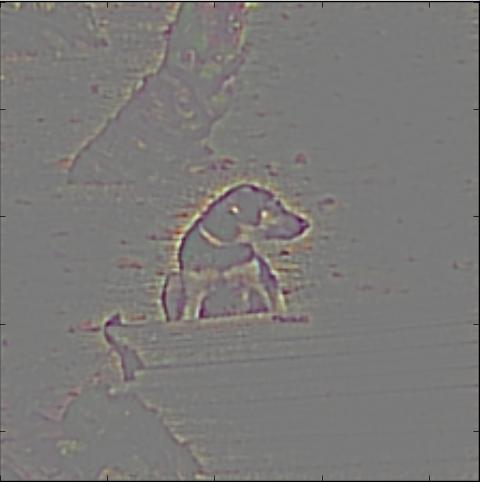}
		\caption*{\footnotesize GBP-max}
	\end{subfigure}
	\begin{subfigure}[b]{0.3\textwidth}
		\centering
		\includegraphics[width=\textwidth]{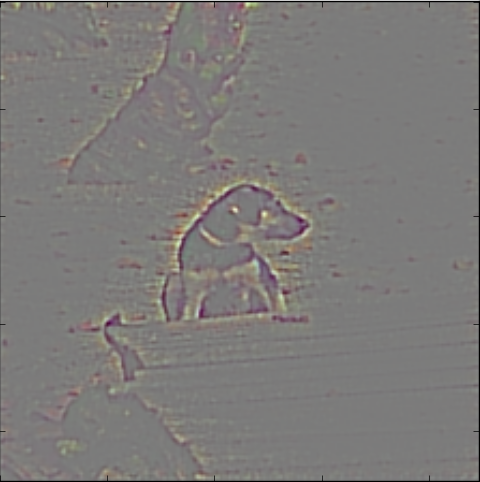}
		\caption*{\footnotesize GBP-731}
	\end{subfigure}
	\begin{subfigure}[b]{0.3\textwidth}
		\centering
		\includegraphics[width=\textwidth]{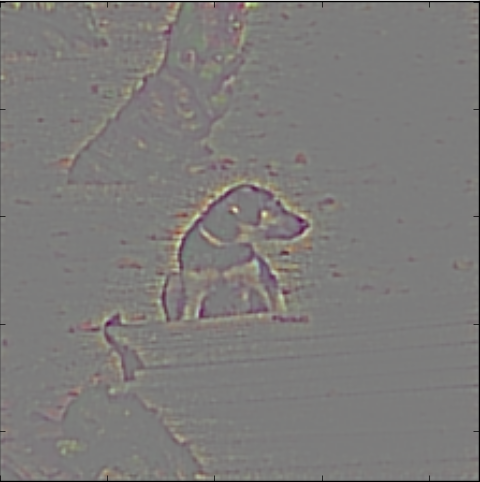}
		\caption*{\footnotesize GBP-815}
	\end{subfigure}
	
	\caption{Saliency map, DeconvNet and GBP visualizations for the trained VGG-16 net with the input image ``dog''.}\label{fig_2}
	\end{minipage}
	
\end{figure*}

\begin{figure*}[!h]
	\begin{minipage}[c]{0.48\textwidth}
	\centering
	\begin{subfigure}[b]{0.3\textwidth}
		\centering
		\includegraphics[width=\textwidth]{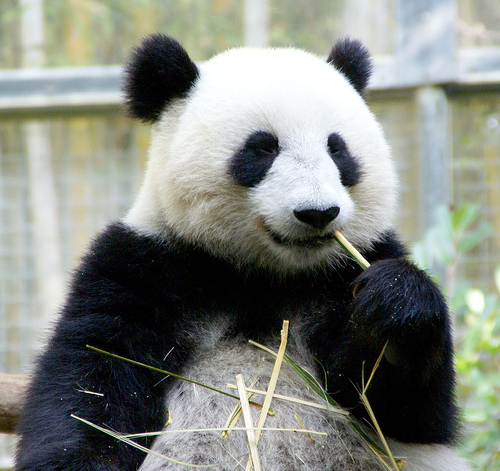}
		\caption*{\footnotesize panda}
	\end{subfigure}
	
	\begin{subfigure}[b]{0.3\textwidth}
		\centering
		\includegraphics[width=\textwidth]{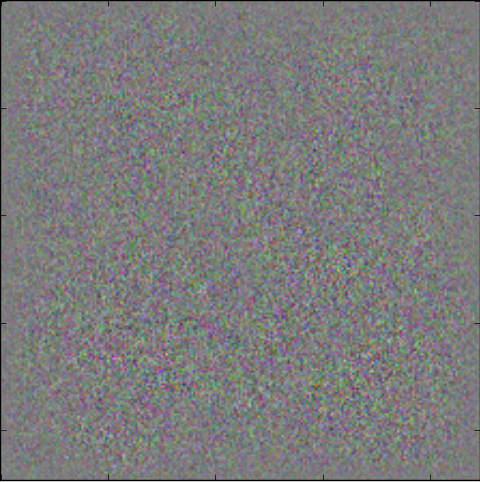}
		\caption*{\footnotesize Sal-max}
	\end{subfigure}
	\begin{subfigure}[b]{0.3\textwidth}
		\centering
		\includegraphics[width=\textwidth]{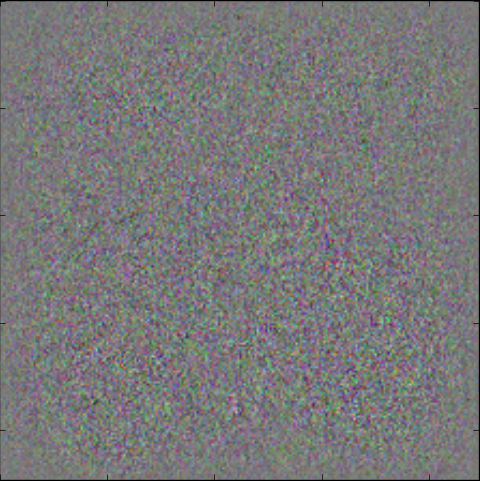}
		\caption*{\footnotesize Sal-731}
	\end{subfigure}
	\begin{subfigure}[b]{0.3\textwidth}
		\centering
		\includegraphics[width=\textwidth]{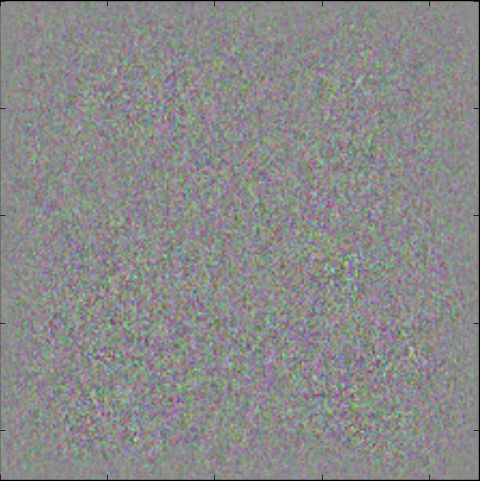}
		\caption*{\footnotesize Sal-815}
	\end{subfigure}
    
    \begin{subfigure}[b]{0.3\textwidth}
		\centering
		\includegraphics[width=\textwidth]{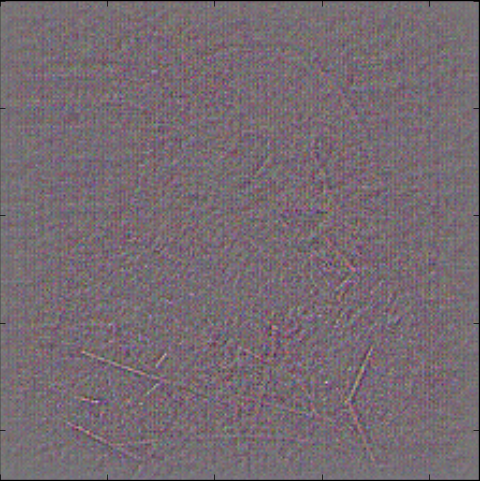}
		\caption*{\footnotesize Deconv-max}
	\end{subfigure}
	\begin{subfigure}[b]{0.3\textwidth}
		\centering
		\includegraphics[width=\textwidth]{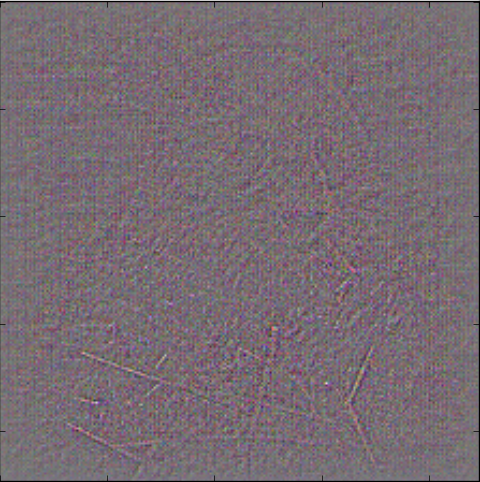}
		\caption*{\footnotesize Deconv-731}
	\end{subfigure}
	\begin{subfigure}[b]{0.3\textwidth}
		\centering
		\includegraphics[width=\textwidth]{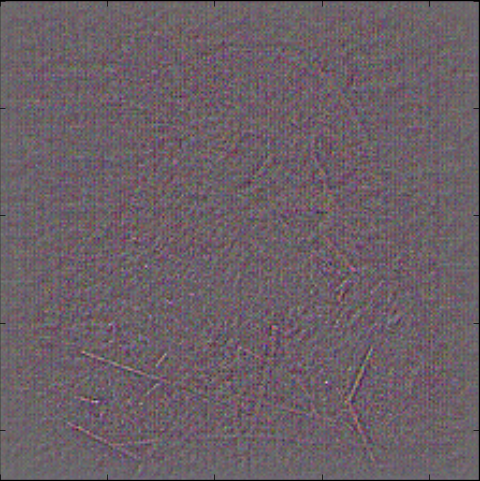}
		\caption*{\footnotesize Deconv-815}
	\end{subfigure}

	\begin{subfigure}[b]{0.3\textwidth}
		\centering
		\includegraphics[width=\textwidth]{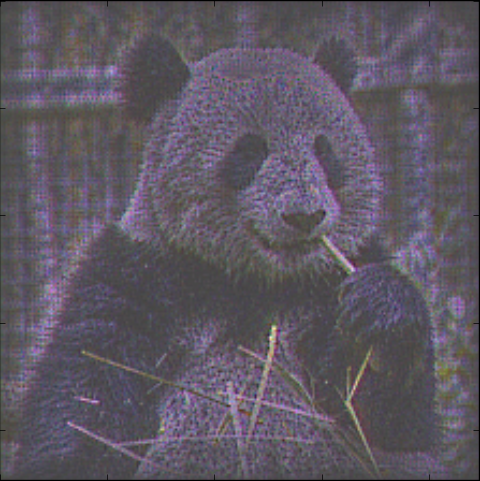}
		\caption*{\footnotesize GBP-max}
	\end{subfigure}
	\begin{subfigure}[b]{0.3\textwidth}
		\centering
		\includegraphics[width=\textwidth]{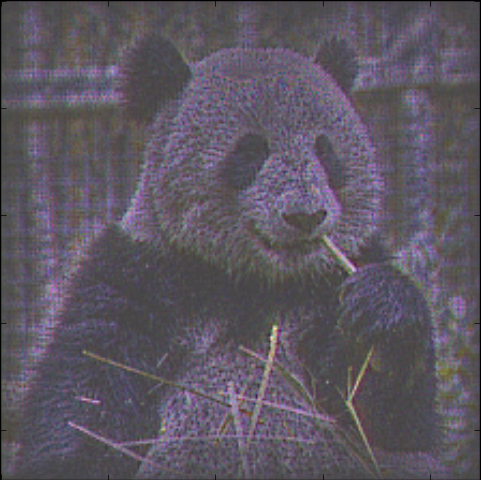}
		\caption*{\footnotesize GBP-731}
	\end{subfigure}
	\begin{subfigure}[b]{0.3\textwidth}
		\centering
		\includegraphics[width=\textwidth]{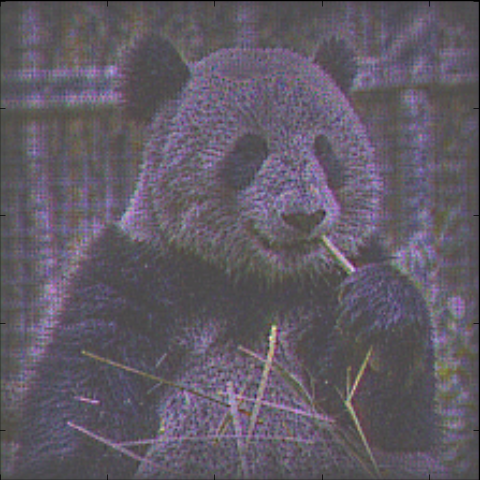}
		\caption*{\footnotesize GBP-815}
	\end{subfigure}

	\caption{Saliency map, DeconvNet and GBP visualizations for the random VGG-16 net with the input image ``panda''.}\label{fig_3}
	\end{minipage}
	\hfill
	\begin{minipage}[c]{0.48\textwidth}
	\centering
	\begin{subfigure}[b]{0.3\textwidth}
		\centering
		\includegraphics[width=\textwidth]{figures/01312018_1/panda.png}
		\caption*{\footnotesize panda}
	\end{subfigure}
	
	\begin{subfigure}[b]{0.3\textwidth}
		\centering
		\includegraphics[width=\textwidth]{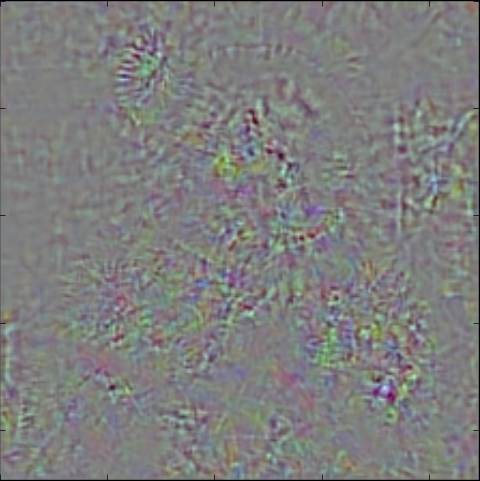}
		\caption*{\footnotesize Sal-max}
	\end{subfigure}
	\begin{subfigure}[b]{0.3\textwidth}
		\centering
		\includegraphics[width=\textwidth]{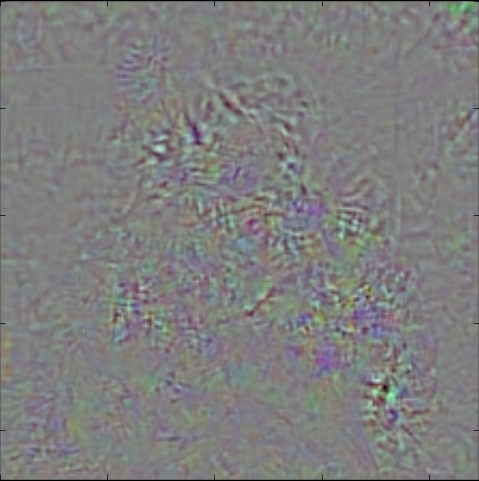}
		\caption*{\footnotesize Sal-731}
	\end{subfigure}
	\begin{subfigure}[b]{0.3\textwidth}
		\centering
		\includegraphics[width=\textwidth]{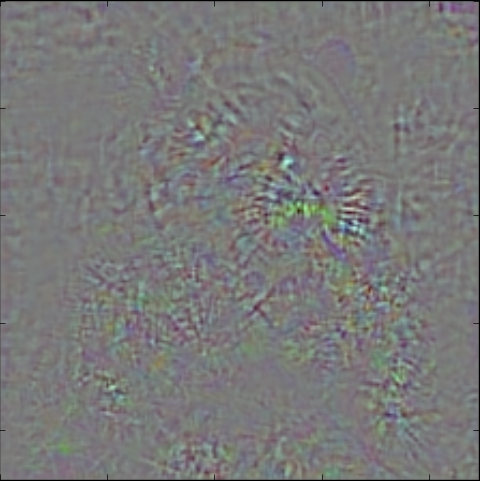}
		\caption*{\footnotesize Sal-815}
	\end{subfigure}
    
    \begin{subfigure}[b]{0.3\textwidth}
		\centering
		\includegraphics[width=\textwidth]{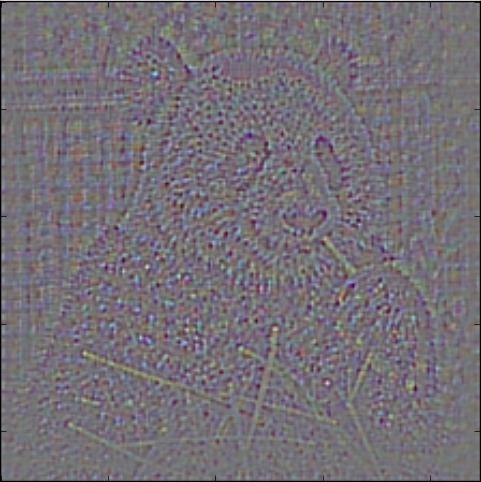}
		\caption*{\footnotesize Deconv-max}
	\end{subfigure}
	\begin{subfigure}[b]{0.3\textwidth}
		\centering
		\includegraphics[width=\textwidth]{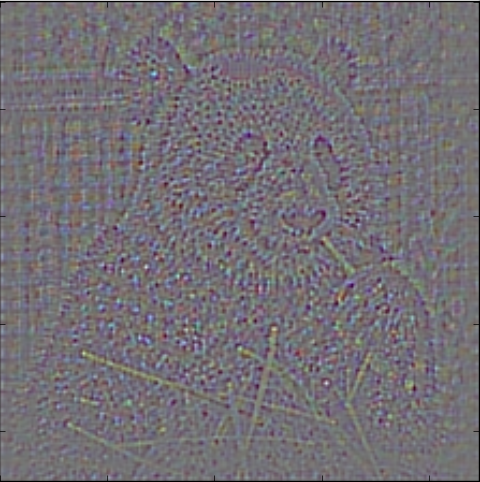}
		\caption*{\footnotesize Deconv-731}
	\end{subfigure}
	\begin{subfigure}[b]{0.3\textwidth}
		\centering
		\includegraphics[width=\textwidth]{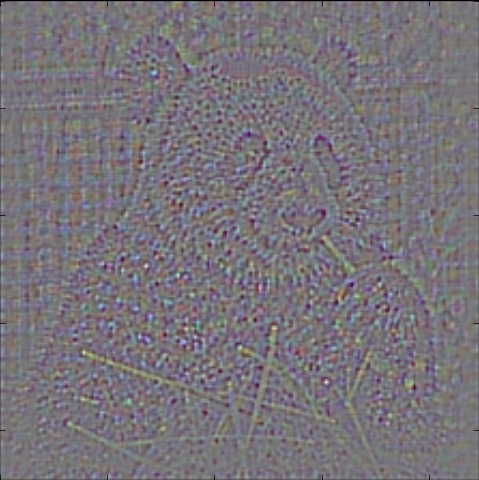}
		\caption*{\footnotesize Deconv-815}
	\end{subfigure}

	\begin{subfigure}[b]{0.3\textwidth}
		\centering
		\includegraphics[width=\textwidth]{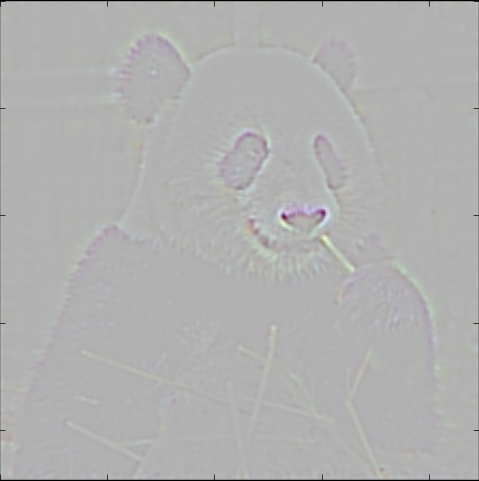}
		\caption*{\footnotesize GBP-max}
	\end{subfigure}
	\begin{subfigure}[b]{0.3\textwidth}
		\centering
		\includegraphics[width=\textwidth]{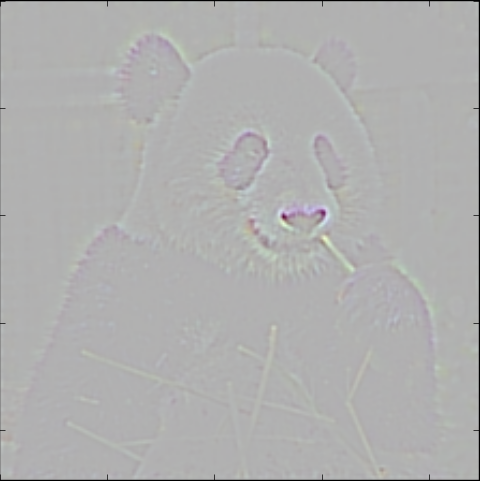}
		\caption*{\footnotesize GBP-731}
	\end{subfigure}
	\begin{subfigure}[b]{0.3\textwidth}
		\centering
		\includegraphics[width=\textwidth]{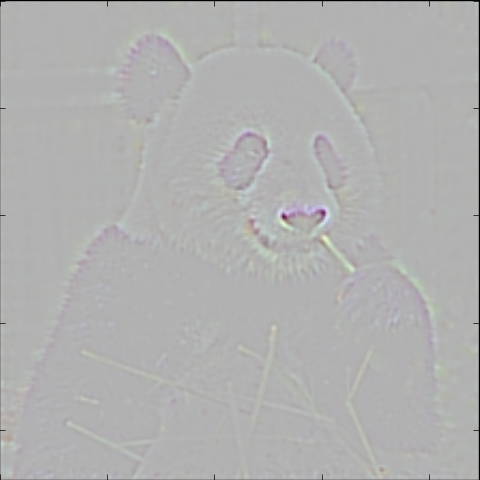}
		\caption*{\footnotesize GBP-815}
	\end{subfigure}

	\caption{Saliency map, DeconvNet and GBP visualizations for the trained VGG-16 net with the input image ``panda''.}\label{fig_4}
	\end{minipage}
	
\end{figure*}

\begin{figure*}[!h]
	\begin{minipage}[c]{0.48\textwidth}
	\centering
	\begin{subfigure}[b]{0.3\textwidth}
		\centering
		\includegraphics[width=\textwidth]{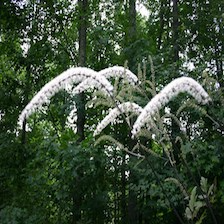}
		\caption*{\footnotesize forest}
	\end{subfigure}
	
	\begin{subfigure}[b]{0.3\textwidth}
		\centering
		\includegraphics[width=\textwidth]{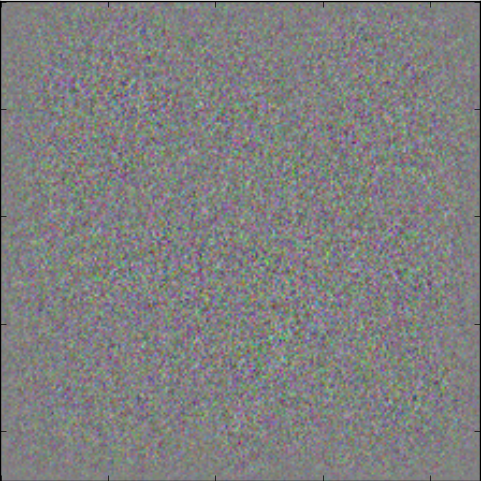}
		\caption*{\footnotesize Sal-max}
	\end{subfigure}
	\begin{subfigure}[b]{0.3\textwidth}
		\centering
		\includegraphics[width=\textwidth]{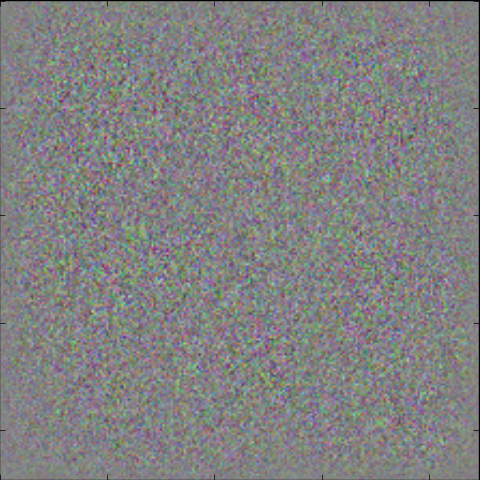}
		\caption*{\footnotesize Sal-731}
	\end{subfigure}
	\begin{subfigure}[b]{0.3\textwidth}
		\centering
		\includegraphics[width=\textwidth]{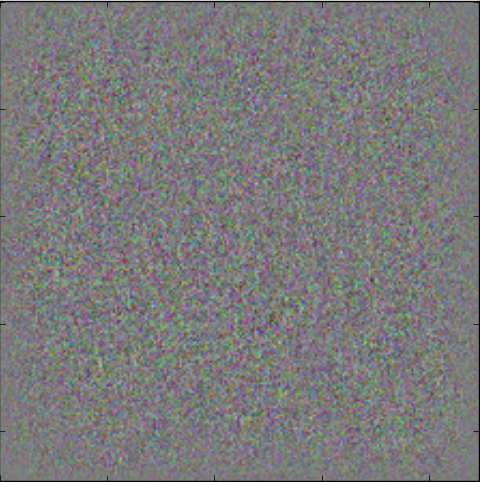}
		\caption*{\footnotesize Sal-815}
	\end{subfigure}
    
    \begin{subfigure}[b]{0.3\textwidth}
		\centering
		\includegraphics[width=\textwidth]{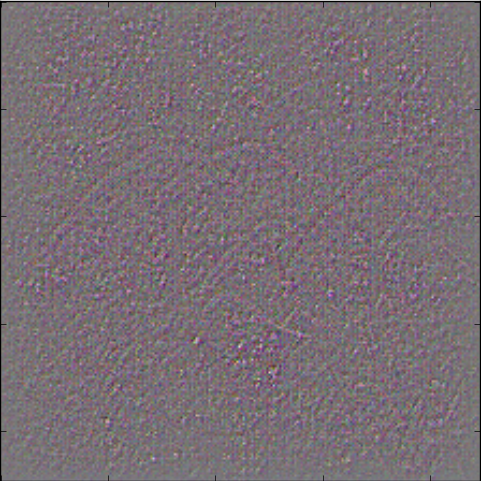}
		\caption*{\footnotesize Deconv-max}
	\end{subfigure}
	\begin{subfigure}[b]{0.3\textwidth}
		\centering
		\includegraphics[width=\textwidth]{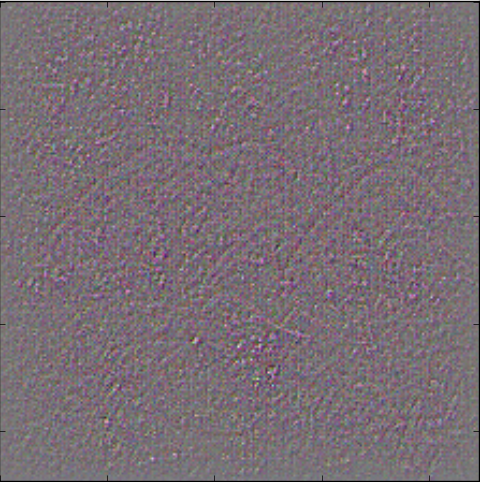}
		\caption*{\footnotesize Deconv-731}
	\end{subfigure}
	\begin{subfigure}[b]{0.3\textwidth}
		\centering
		\includegraphics[width=\textwidth]{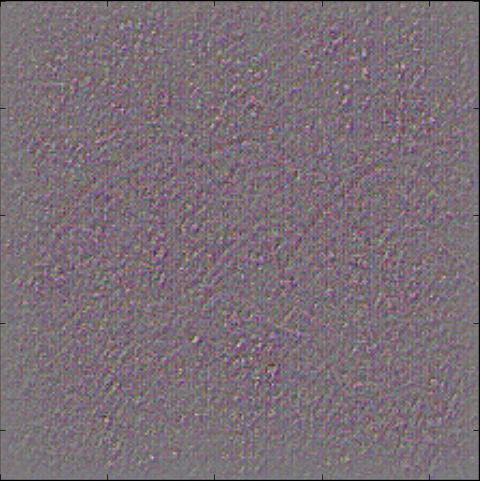}
		\caption*{\footnotesize Deconv-815}
	\end{subfigure}

	\begin{subfigure}[b]{0.3\textwidth}
		\centering
		\includegraphics[width=\textwidth]{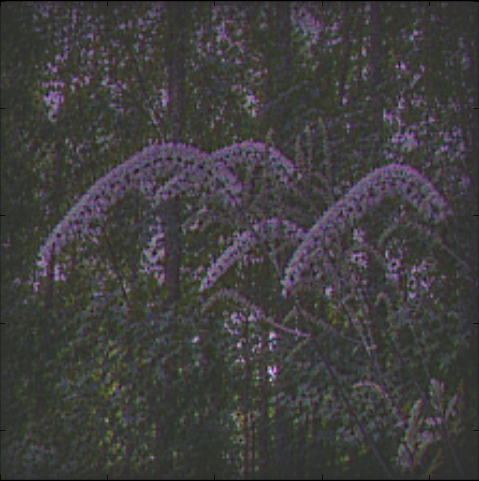}
		\caption*{\footnotesize GBP-max}
	\end{subfigure}
	\begin{subfigure}[b]{0.3\textwidth}
		\centering
		\includegraphics[width=\textwidth]{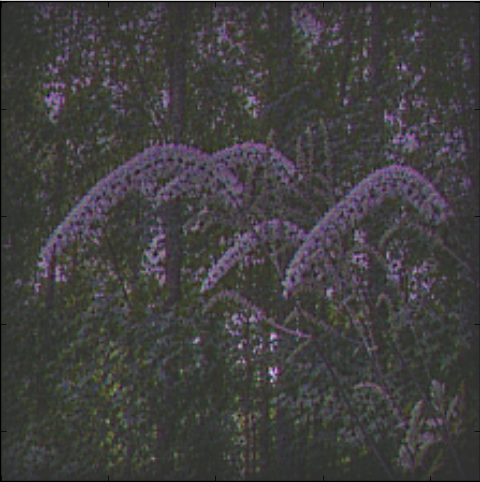}
		\caption*{\footnotesize GBP-731}
	\end{subfigure}
	\begin{subfigure}[b]{0.3\textwidth}
		\centering
		\includegraphics[width=\textwidth]{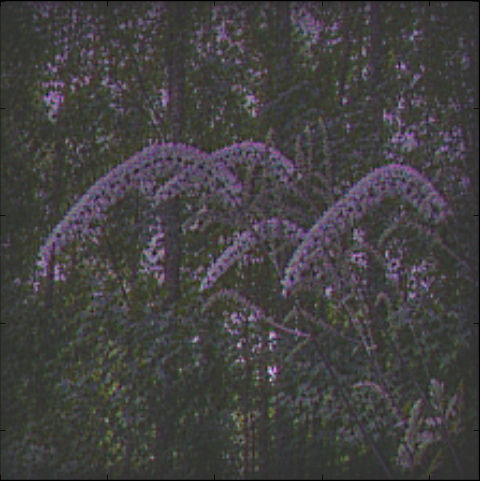}
		\caption*{\footnotesize GBP-815}
	\end{subfigure}

	\caption{Saliency map, DeconvNet and GBP visualizations for the random VGG-16 net with the input image ``forest''.}\label{fig_5}
	\end{minipage}
	\hfill
	\begin{minipage}[c]{0.48\textwidth}
	\centering
	\begin{subfigure}[b]{0.3\textwidth}
		\centering
		\includegraphics[width=\textwidth]{figures/01312018_1/forest.png}
		\caption*{\footnotesize forest}
	\end{subfigure}
	
	\begin{subfigure}[b]{0.3\textwidth}
		\centering
		\includegraphics[width=\textwidth]{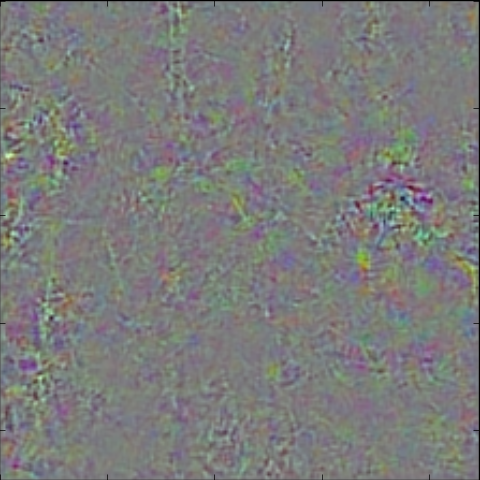}
		\caption*{\footnotesize Sal-max}
	\end{subfigure}
	\begin{subfigure}[b]{0.3\textwidth}
		\centering
		\includegraphics[width=\textwidth]{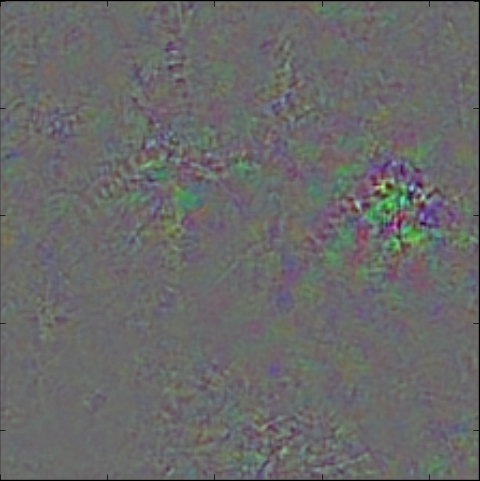}
		\caption*{\footnotesize Sal-731}
	\end{subfigure}
	\begin{subfigure}[b]{0.3\textwidth}
		\centering
		\includegraphics[width=\textwidth]{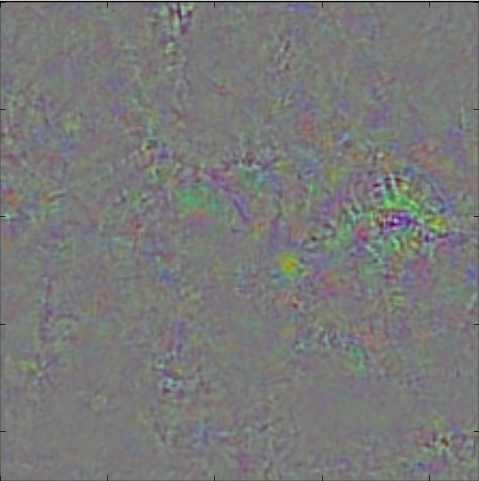}
		\caption*{\footnotesize Sal-815}
	\end{subfigure}
    
    \begin{subfigure}[b]{0.3\textwidth}
		\centering
		\includegraphics[width=\textwidth]{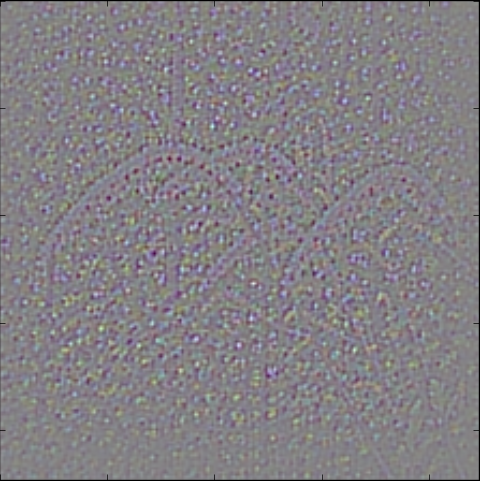}
		\caption*{\footnotesize Deconv-max}
	\end{subfigure}
	\begin{subfigure}[b]{0.3\textwidth}
		\centering
		\includegraphics[width=\textwidth]{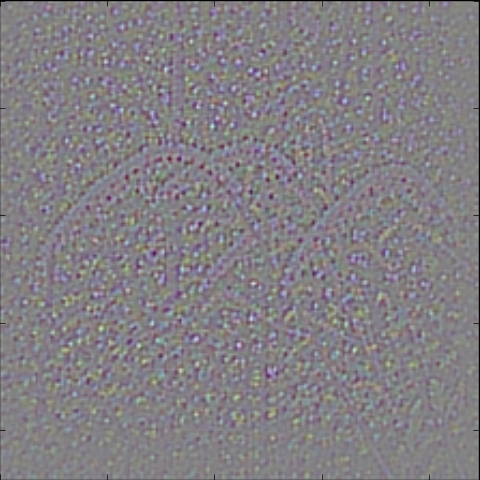}
		\caption*{\footnotesize Deconv-731}
	\end{subfigure}
	\begin{subfigure}[b]{0.3\textwidth}
		\centering
		\includegraphics[width=\textwidth]{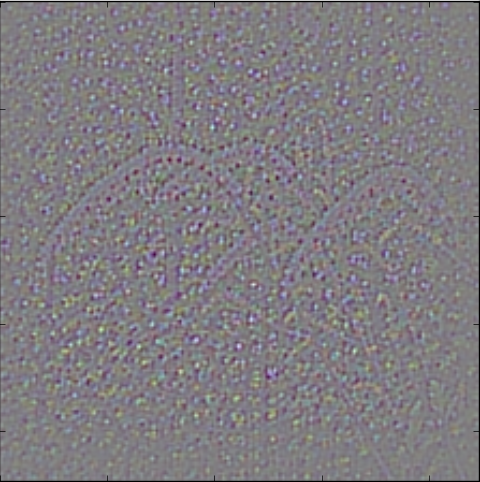}
		\caption*{\footnotesize Deconv-815}
	\end{subfigure}

	\begin{subfigure}[b]{0.3\textwidth}
		\centering
		\includegraphics[width=\textwidth]{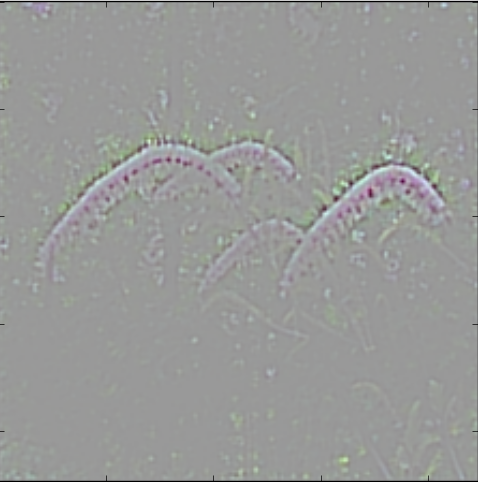}
		\caption*{\footnotesize GBP-max}
	\end{subfigure}
	\begin{subfigure}[b]{0.3\textwidth}
		\centering
		\includegraphics[width=\textwidth]{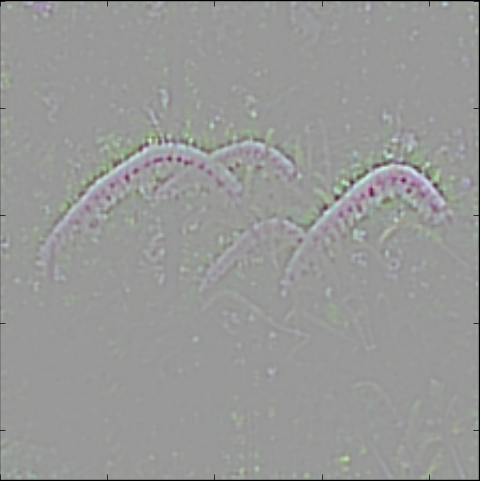}
		\caption*{\footnotesize GBP-731}
	\end{subfigure}
	\begin{subfigure}[b]{0.3\textwidth}
		\centering
		\includegraphics[width=\textwidth]{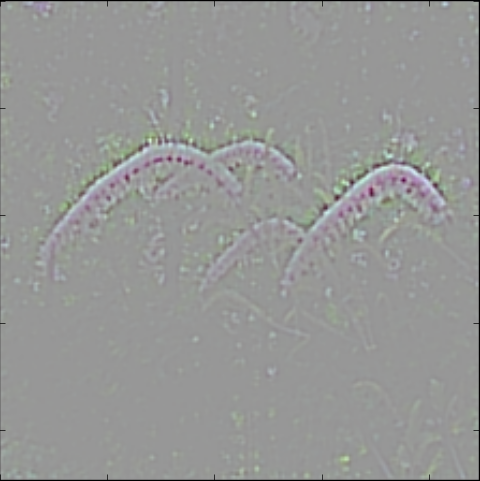}
		\caption*{\footnotesize GBP-815}
	\end{subfigure}
	
	\caption{Saliency map, DeconvNet and GBP visualizations for the trained VGG-16 net with the input image ``forest''.}\label{fig_6}
	\end{minipage}
	
\end{figure*}

\begin{figure*}[!h]
	\begin{minipage}[c]{0.48\textwidth}
	\centering
	\begin{subfigure}[b]{0.3\textwidth}
		\centering
		\includegraphics[width=\textwidth]{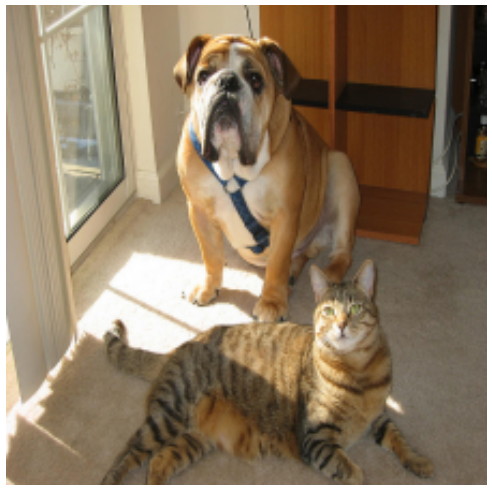}
		\caption*{\footnotesize mastiff}
	\end{subfigure}
	
	\begin{subfigure}[b]{0.3\textwidth}
		\centering
		\includegraphics[width=\textwidth]{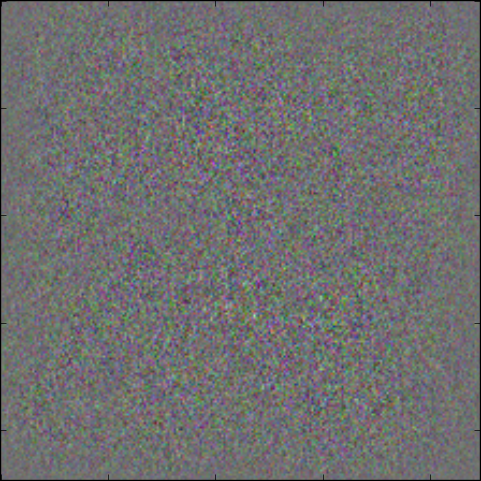}
		\caption*{\footnotesize Sal-max}
	\end{subfigure}
	\begin{subfigure}[b]{0.3\textwidth}
		\centering
		\includegraphics[width=\textwidth]{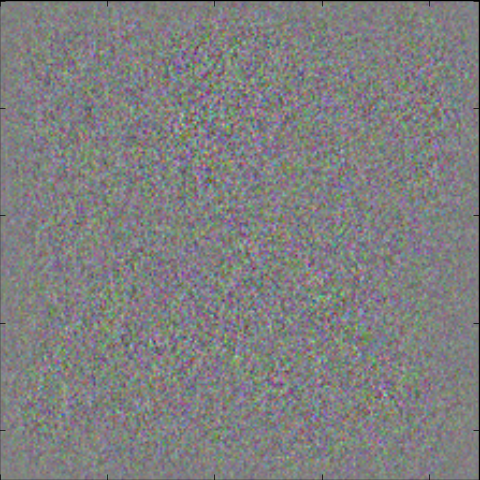}
		\caption*{\footnotesize Sal-731}
	\end{subfigure}
	\begin{subfigure}[b]{0.3\textwidth}
		\centering
		\includegraphics[width=\textwidth]{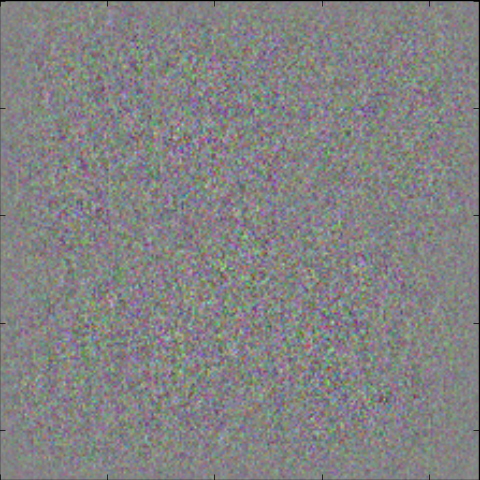}
		\caption*{\footnotesize Sal-815}
	\end{subfigure}
    
    \begin{subfigure}[b]{0.3\textwidth}
		\centering
		\includegraphics[width=\textwidth]{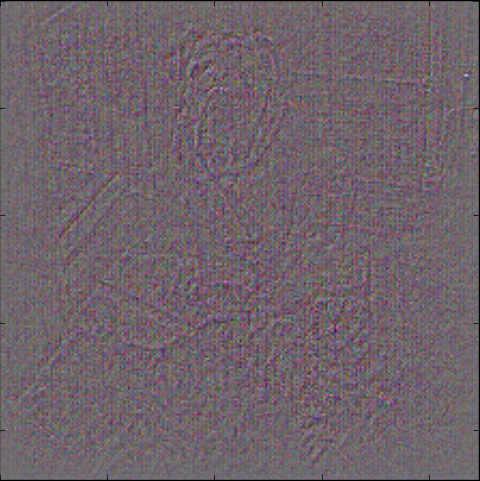}
		\caption*{\footnotesize Deconv-max}
	\end{subfigure}
	\begin{subfigure}[b]{0.3\textwidth}
		\centering
		\includegraphics[width=\textwidth]{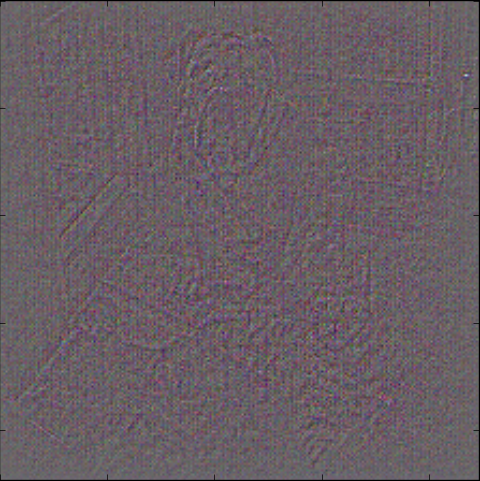}
		\caption*{\footnotesize Deconv-731}
	\end{subfigure}
	\begin{subfigure}[b]{0.3\textwidth}
		\centering
		\includegraphics[width=\textwidth]{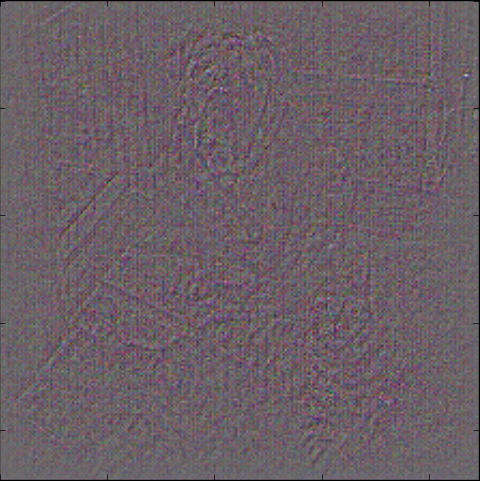}
		\caption*{\footnotesize Deconv-815}
	\end{subfigure}

	\begin{subfigure}[b]{0.3\textwidth}
		\centering
		\includegraphics[width=\textwidth]{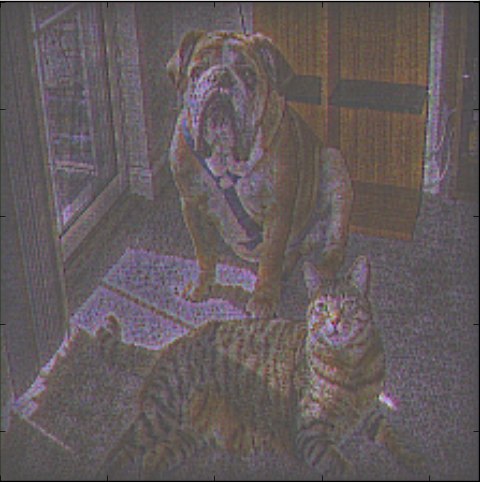}
		\caption*{\footnotesize GBP-max}
	\end{subfigure}
	\begin{subfigure}[b]{0.3\textwidth}
		\centering
		\includegraphics[width=\textwidth]{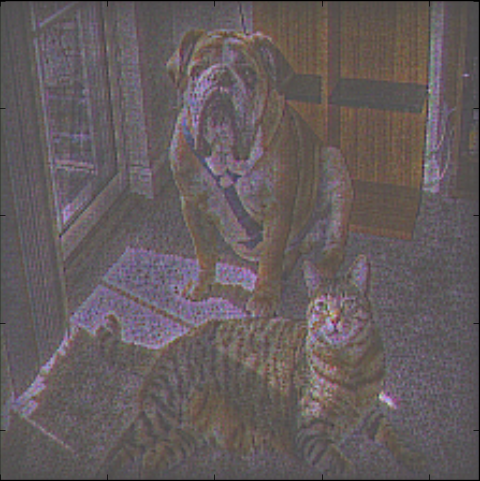}
		\caption*{\footnotesize GBP-731}
	\end{subfigure}
	\begin{subfigure}[b]{0.3\textwidth}
		\centering
		\includegraphics[width=\textwidth]{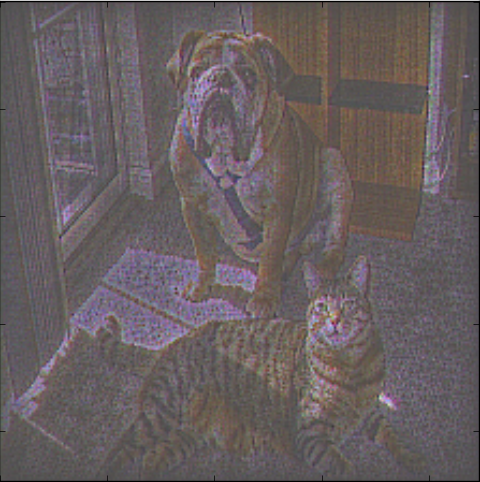}
		\caption*{\footnotesize GBP-815}
	\end{subfigure}
	
	\caption{Saliency map, DeconvNet and GBP visualizations for the random VGG-16 net with the input image ``mastiff''.}\label{fig_7}
	
	\end{minipage}
	\hfill
	\begin{minipage}[c]{0.48\textwidth}
	\centering
	\begin{subfigure}[b]{0.3\textwidth}
		\centering
		\includegraphics[width=\textwidth]{figures/01312018_1/mastiff.png}
		\caption*{\footnotesize mastiff}
	\end{subfigure}
	
	\begin{subfigure}[b]{0.3\textwidth}
		\centering
		\includegraphics[width=\textwidth]{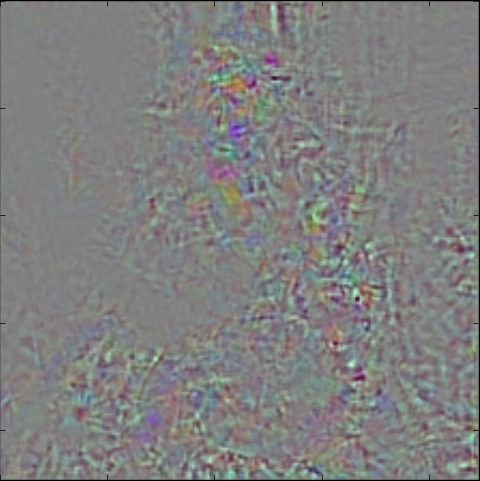}
		\caption*{\footnotesize Sal-max}
	\end{subfigure}
	\begin{subfigure}[b]{0.3\textwidth}
		\centering
		\includegraphics[width=\textwidth]{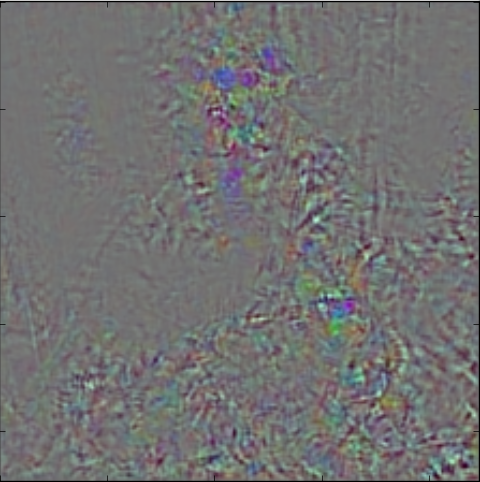}
		\caption*{\footnotesize Sal-731}
	\end{subfigure}
	\begin{subfigure}[b]{0.3\textwidth}
		\centering
		\includegraphics[width=\textwidth]{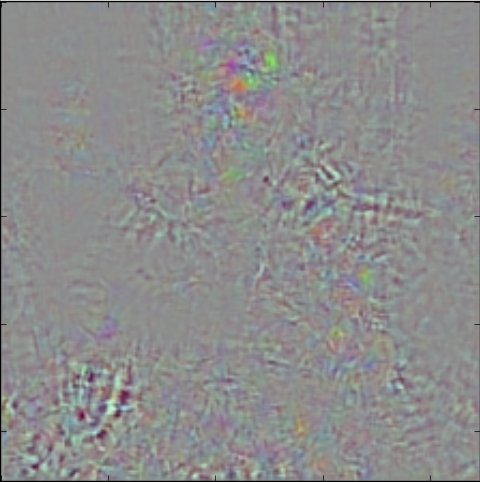}
		\caption*{\footnotesize Sal-815}
	\end{subfigure}
    
    \begin{subfigure}[b]{0.3\textwidth}
		\centering
		\includegraphics[width=\textwidth]{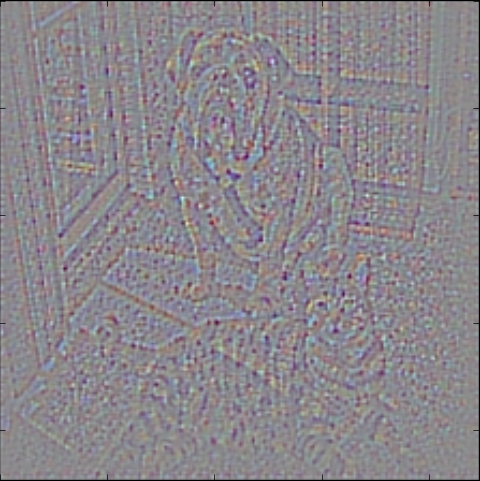}
		\caption*{\footnotesize Deconv-max}
	\end{subfigure}
	\begin{subfigure}[b]{0.3\textwidth}
		\centering
		\includegraphics[width=\textwidth]{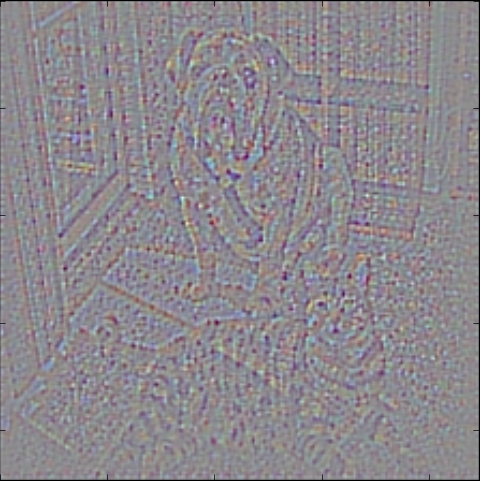}
		\caption*{\footnotesize Deconv-731}
	\end{subfigure}
	\begin{subfigure}[b]{0.3\textwidth}
		\centering
		\includegraphics[width=\textwidth]{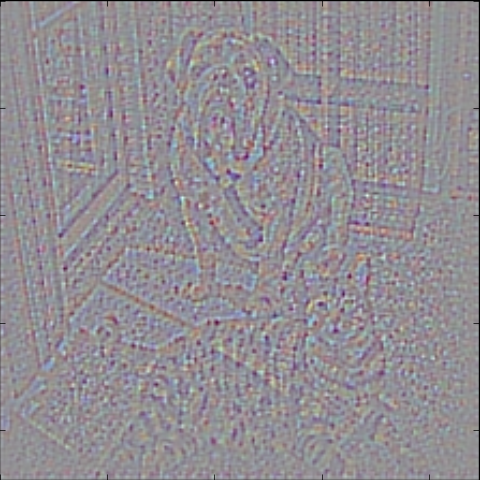}
		\caption*{\footnotesize Deconv-815}
	\end{subfigure}

	\begin{subfigure}[b]{0.3\textwidth}
		\centering
		\includegraphics[width=\textwidth]{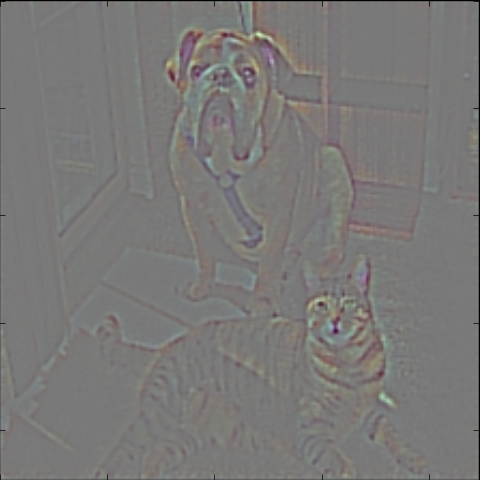}
		\caption*{\footnotesize GBP-max}
	\end{subfigure}
	\begin{subfigure}[b]{0.3\textwidth}
		\centering
		\includegraphics[width=\textwidth]{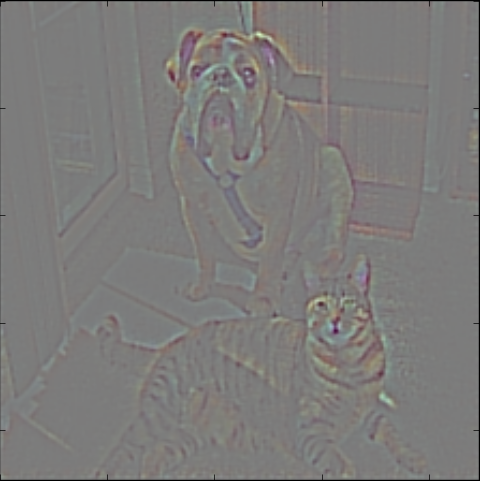}
		\caption*{\footnotesize GBP-731}
	\end{subfigure}
	\begin{subfigure}[b]{0.3\textwidth}
		\centering
		\includegraphics[width=\textwidth]{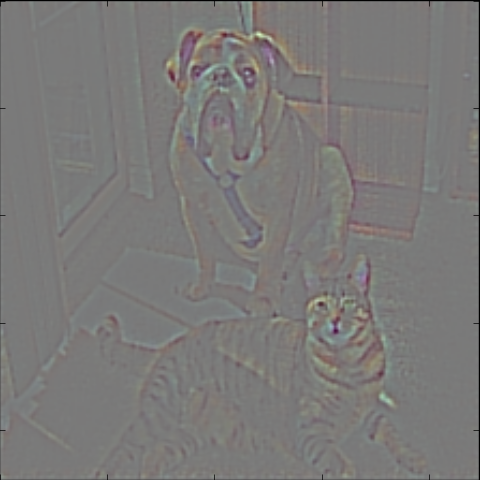}
		\caption*{\footnotesize GBP-815}
	\end{subfigure}
	
	\caption{Saliency map, DeconvNet and GBP visualizations for the trained VGG-16 net with the input image ``mastiff''.}\label{fig_8}
	\end{minipage}
	
\end{figure*}

We provide more results for backpropagation-based visualizations including saliency map, DeconvNet and GBP in both untrained (randomly initialized) and trained VGG-16 net. The input images -- labeled as ``dog'', ``panda'', ``forest'' and ``mastiff'' -- are randomly chosen from the ImageNet dataset. As we can see, all the results (Figures \ref{fig_1} - \ref{fig_8}) are consistent with our previous empirical observations that GBP and DeconvNet are more visually compelling but less class-sensitive than saliency map.

\section{Comparison Between GBP and Edge Detector}

\begin{figure*}[!h]
	\centering
		\begin{subfigure}[b]{0.19\textwidth}
			\centering
			\includegraphics[width=\textwidth]{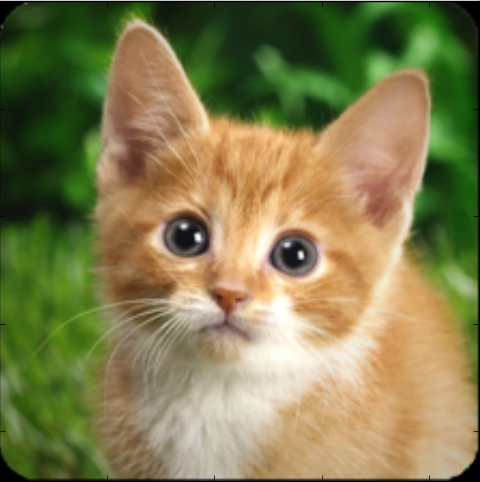}
		\end{subfigure}
		\begin{subfigure}[b]{0.19\textwidth}
			\centering
			\includegraphics[width=\textwidth]{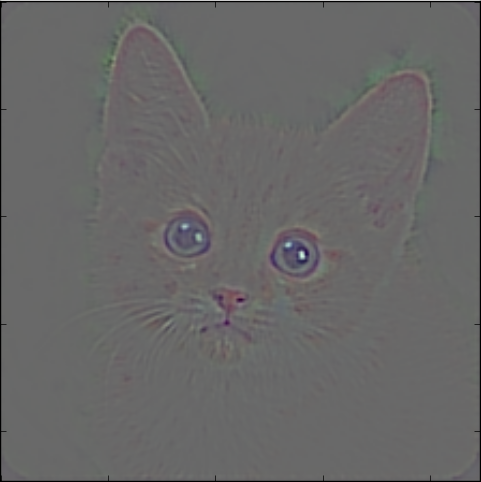}
		\end{subfigure}
		\begin{subfigure}[b]{0.19\textwidth}
			\centering
			\includegraphics[width=\textwidth]{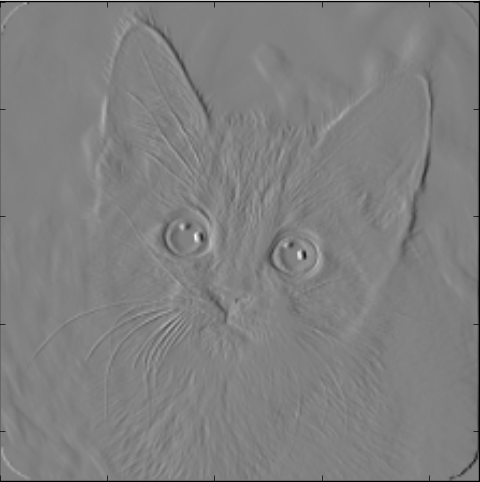}
		\end{subfigure}
		
		\begin{subfigure}[b]{0.19\textwidth}
			\centering
			\includegraphics[width=\textwidth]{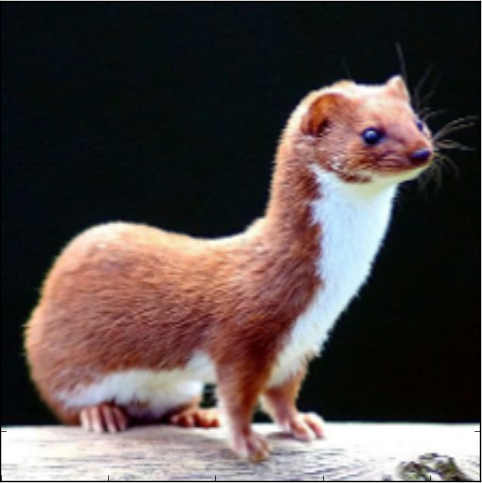}
		\end{subfigure}
		\begin{subfigure}[b]{0.19\textwidth}
			\centering
			\includegraphics[width=\textwidth]{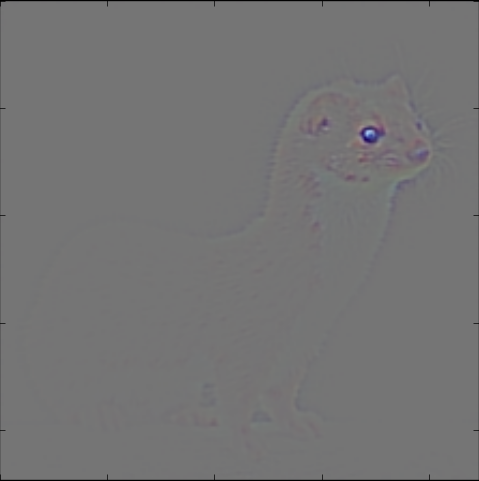}
		\end{subfigure}
		\begin{subfigure}[b]{0.19\textwidth}
			\centering
			\includegraphics[width=\textwidth]{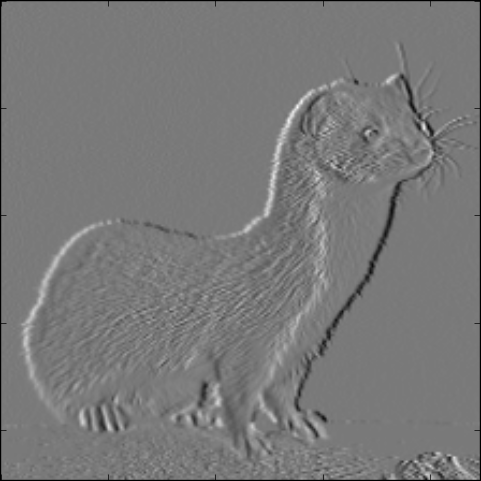}
		\end{subfigure}
		
		\begin{subfigure}[b]{0.19\textwidth}
			\centering
			\includegraphics[width=\textwidth]{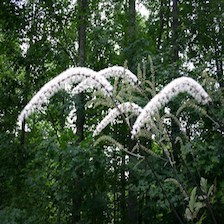}
		\end{subfigure}
		\begin{subfigure}[b]{0.19\textwidth}
			\centering
			\includegraphics[width=\textwidth]{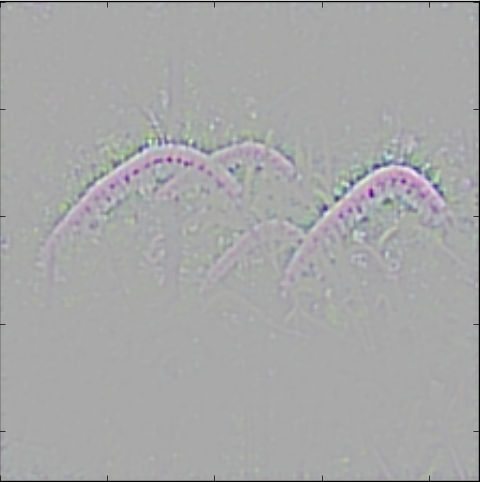}
		\end{subfigure}
		\begin{subfigure}[b]{0.19\textwidth}
			\centering
			\includegraphics[width=\textwidth]{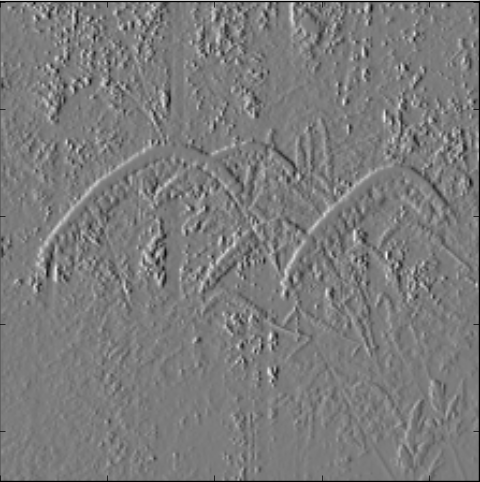}
		\end{subfigure}
	\caption{Comparison between the GBP visualization and the linear edge detector. The left column contains three sample inputs. The middle column contains the GBP visualization for each input. The right column is a linear vertical edge detector applied to each input. Specifically, the edge detector is designed by taking each pixel in the image and subtracting the neighboring pixel on the left. 
	}\label{fig_9}
	
\end{figure*}

Here we compare the GBP visualization with a linear vertical edge detector, as shown in Figure \ref{fig_9}. At the first glance, the GBP visualizations in a trained VGG-16 net are very similar to the results of an edge detector. In other words, GBP indeed pays much attention to the edge information like a Gabor filter. However, there exist subtle differences between GBP and linear edge detectors.  As we can see, the linear vertical edge detector will highlight all the horizontal intensity changes, while GBP has the additional ability to filter out some background image patches. 

\section{More Experiments on Partly Trained VGG-16 Net}

In this section, we provide more GBP visualizations by feeding more images to a partly trained VGG-16 net. Specifically, we consider two kinds of weights loading strategies for the VGG-16 net. The first one is to load trained weights \textbf{up to} a given layer as shown in Figure~\ref{fig_10}. The second one is to load trained weights for all the layers \textbf{except for} a given layer as shown in Figure~\ref{fig_11}. The results are consistent with our previous analysis: it is the trained weights in the convolutional layers rather than those in the dense layers that account for filtering out image patches. Also, earlier convolutional layers have a greater impact on the GBP visualization than later convolutional layers.

\section{More Experiments on ResNet}

\begin{figure*}[!h]
	\begin{minipage}[c]{0.48\textwidth}
	\centering
	\begin{subfigure}[b]{0.2\textwidth}
		\centering
		\includegraphics[width=\textwidth]{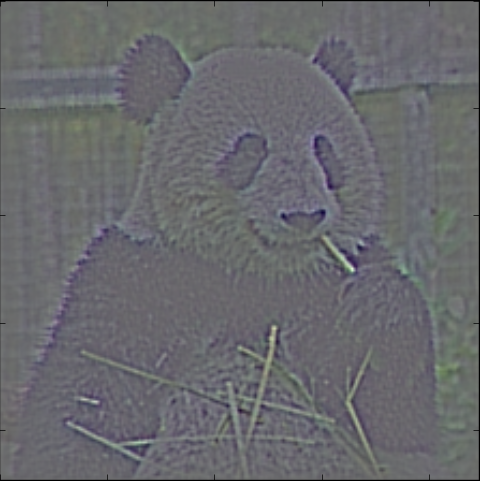}
	\end{subfigure}
	\begin{subfigure}[b]{0.2\textwidth}
		\centering
		\includegraphics[width=\textwidth]{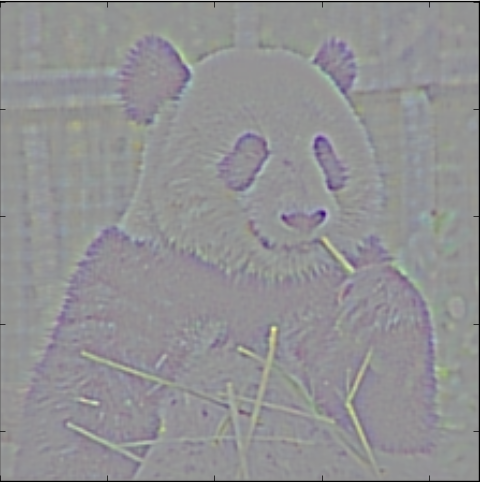}
	\end{subfigure}
	\begin{subfigure}[b]{0.2\textwidth}
		\centering
		\includegraphics[width=\textwidth]{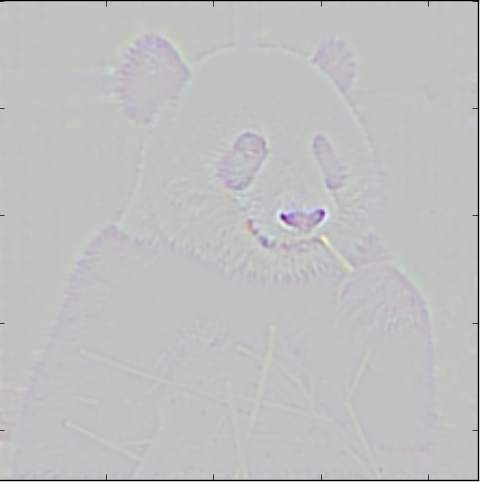}
	\end{subfigure}
	\begin{subfigure}[b]{0.2\textwidth}
		\centering
		\includegraphics[width=\textwidth]{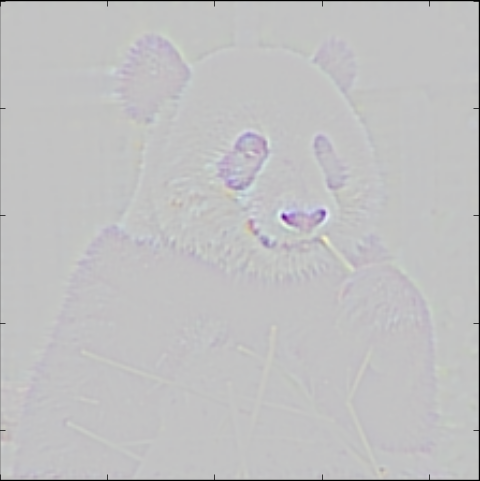}
	\end{subfigure}
	
	\begin{subfigure}[b]{0.2\textwidth}
		\centering
		\includegraphics[width=\textwidth]{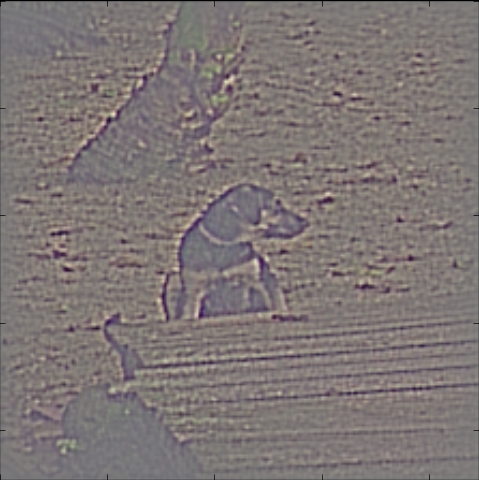}
	\end{subfigure}
	\begin{subfigure}[b]{0.2\textwidth}
		\centering
		\includegraphics[width=\textwidth]{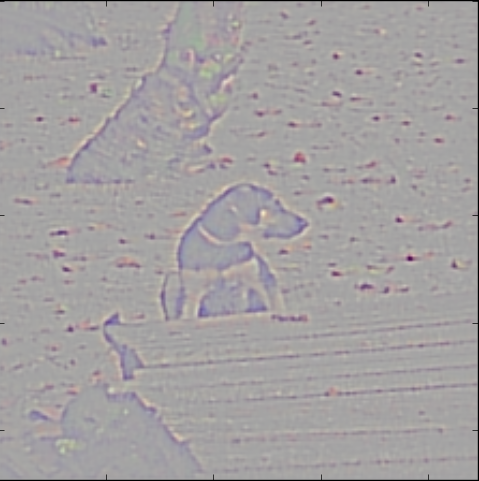}
	\end{subfigure}
	\begin{subfigure}[b]{0.2\textwidth}
		\centering
		\includegraphics[width=\textwidth]{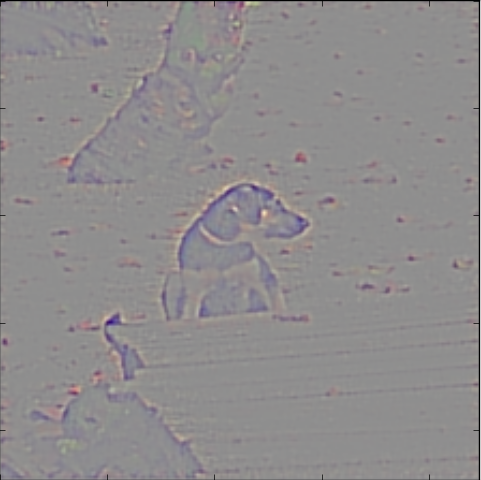}
	\end{subfigure}
	\begin{subfigure}[b]{0.2\textwidth}
		\centering
		\includegraphics[width=\textwidth]{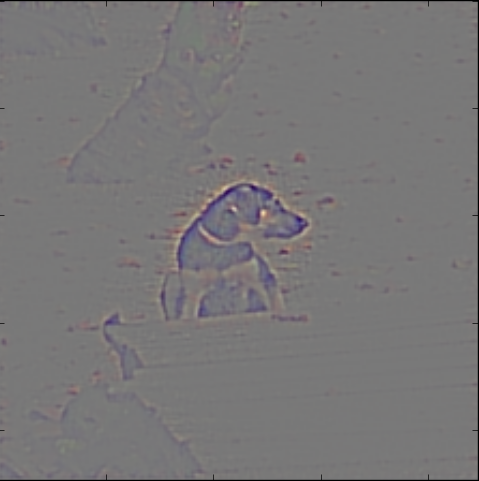}
	\end{subfigure}
	
	\begin{subfigure}[b]{0.2\textwidth}
		\centering
		\includegraphics[width=\textwidth]{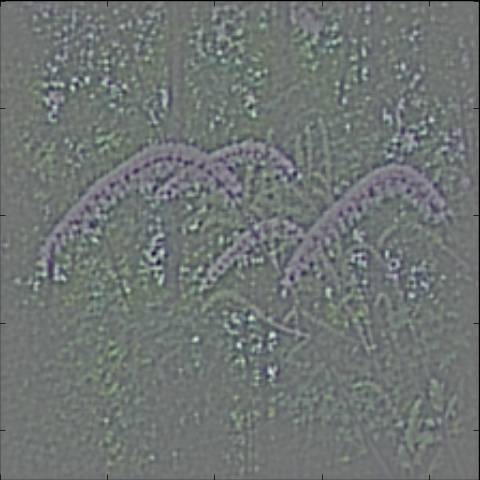}
	\end{subfigure}
	\begin{subfigure}[b]{0.2\textwidth}
		\centering
		\includegraphics[width=\textwidth]{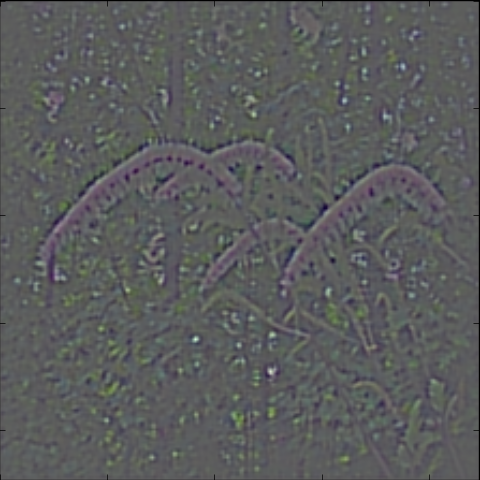}
	\end{subfigure}
	\begin{subfigure}[b]{0.2\textwidth}
		\centering
		\includegraphics[width=\textwidth]{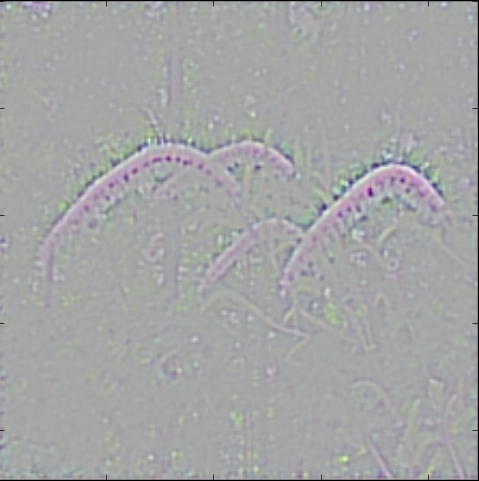}
	\end{subfigure}
	\begin{subfigure}[b]{0.2\textwidth}
		\centering
		\includegraphics[width=\textwidth]{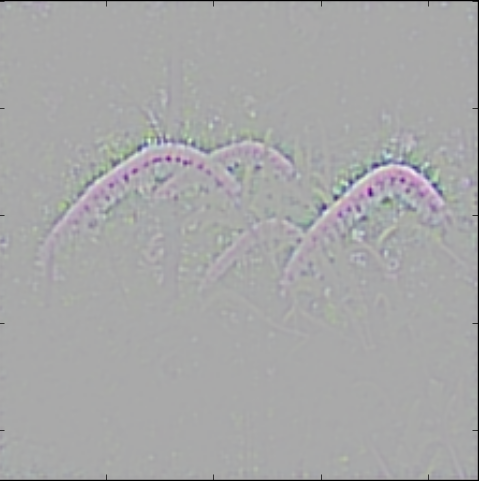}
	\end{subfigure}
	
	\begin{subfigure}[b]{0.2\textwidth}
		\centering
		\includegraphics[width=\textwidth]{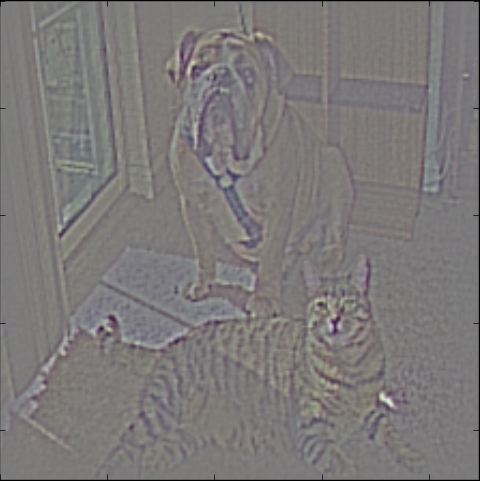}
		\caption*{Conv1-1*}
	\end{subfigure}
	\begin{subfigure}[b]{0.2\textwidth}
		\centering
		\includegraphics[width=\textwidth]{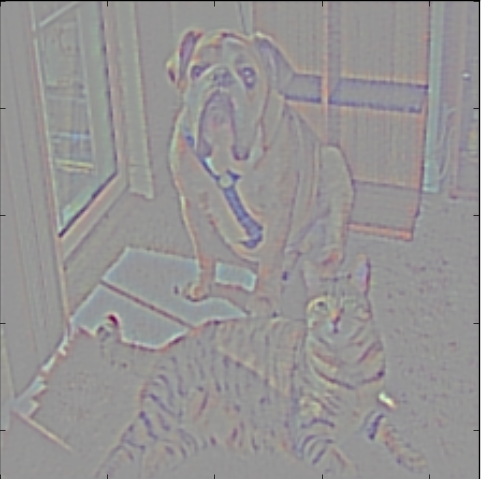}
		\caption*{Conv3-1*}
	\end{subfigure}
	\begin{subfigure}[b]{0.2\textwidth}
		\centering
		\includegraphics[width=\textwidth]{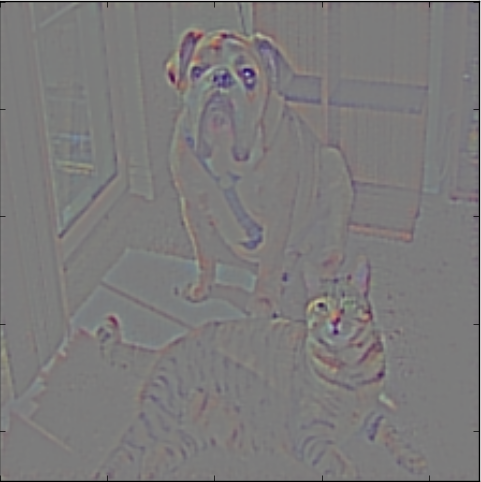}
		\caption*{Conv5-1*}
	\end{subfigure}
	\begin{subfigure}[b]{0.2\textwidth}
		\centering
		\includegraphics[width=\textwidth]{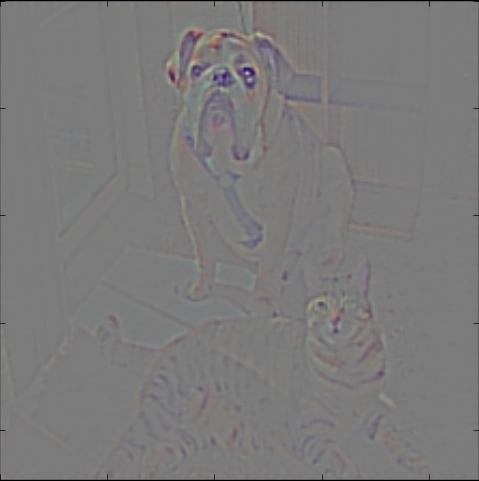}
		\caption*{FC3*}
	\end{subfigure}
		
	\caption{Load the trained weights of the VGG-16 net \textbf{up to} the indexed layer and leave the rest layers to be randomly initialized (denoted by the star sign) with different input images.}\label{fig_10}
	\end{minipage}
	\hfill
	\begin{minipage}[c]{0.48\textwidth}
	\centering
	\begin{subfigure}[b]{0.2\textwidth}
		\centering
		\includegraphics[width=\textwidth]{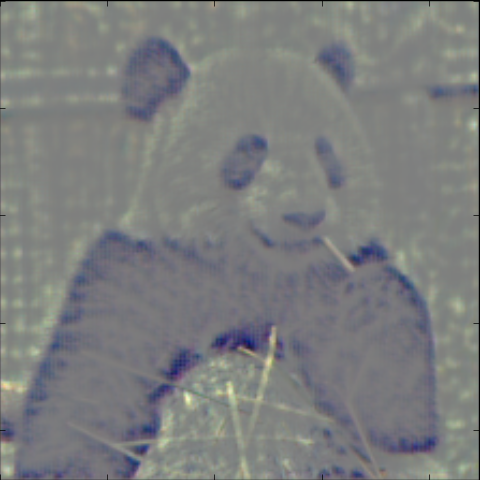}
	\end{subfigure}
	\begin{subfigure}[b]{0.2\textwidth}
		\centering
		\includegraphics[width=\textwidth]{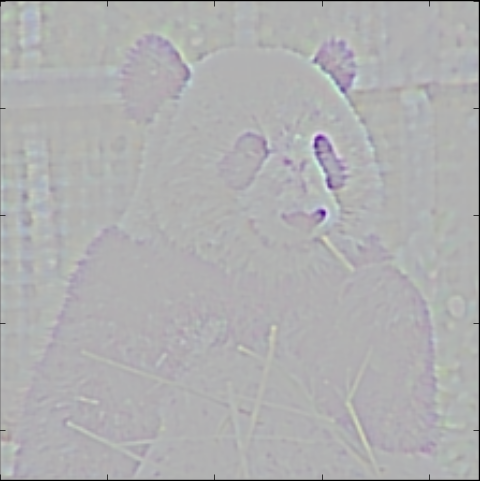}
	\end{subfigure}
	\begin{subfigure}[b]{0.2\textwidth}
		\centering
		\includegraphics[width=\textwidth]{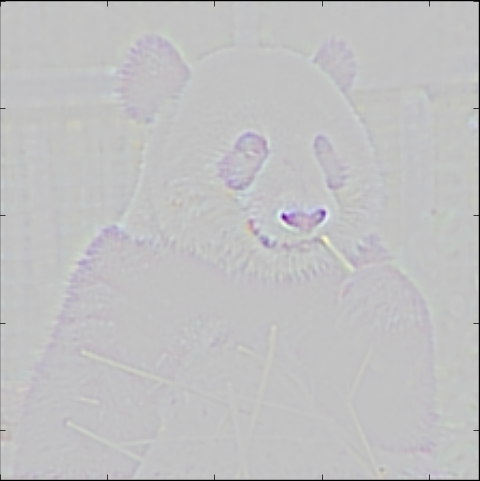}
	\end{subfigure}
	\begin{subfigure}[b]{0.2\textwidth}
		\centering
		\includegraphics[width=\textwidth]{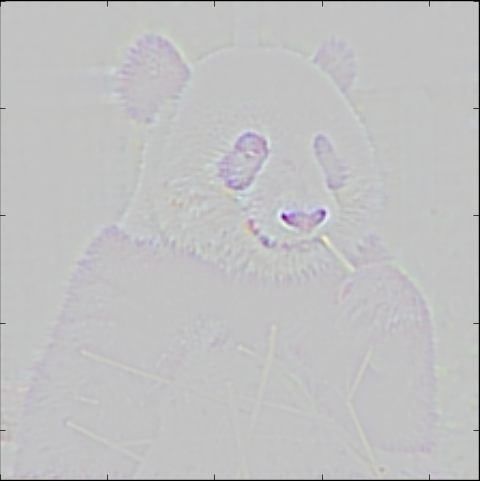}
	\end{subfigure}
	
	\begin{subfigure}[b]{0.2\textwidth}
		\centering
		\includegraphics[width=\textwidth]{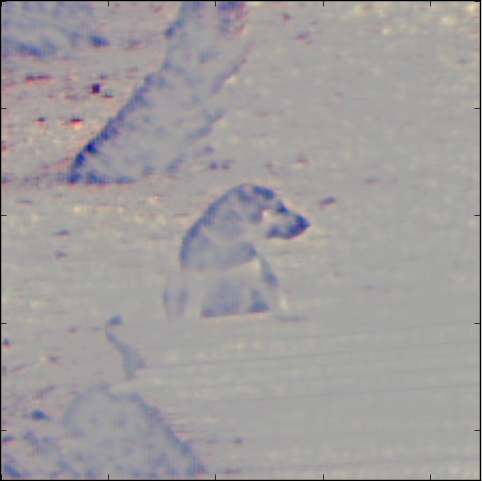}
	\end{subfigure}
	\begin{subfigure}[b]{0.2\textwidth}
		\centering
		\includegraphics[width=\textwidth]{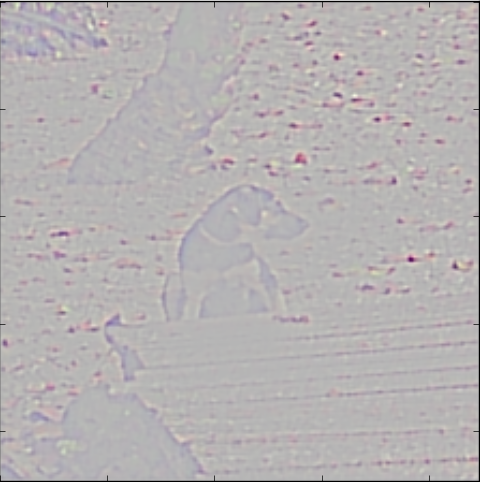}
	\end{subfigure}
	\begin{subfigure}[b]{0.2\textwidth}
		\centering
		\includegraphics[width=\textwidth]{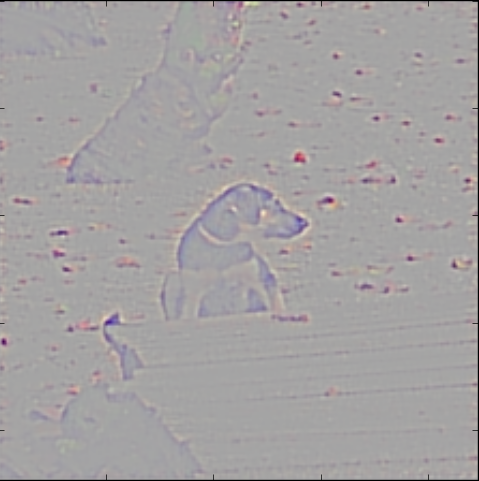}
	\end{subfigure}
	\begin{subfigure}[b]{0.2\textwidth}
		\centering
		\includegraphics[width=\textwidth]{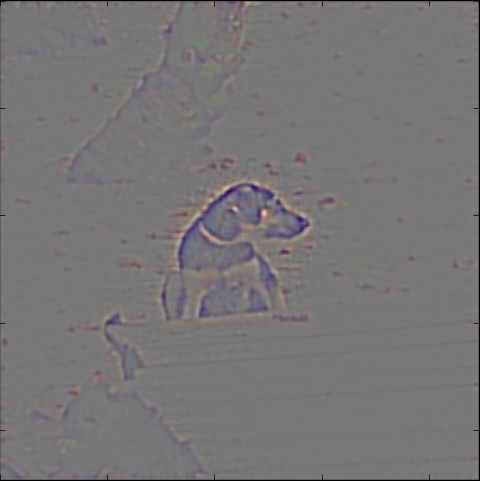}
	\end{subfigure}
	
	\begin{subfigure}[b]{0.2\textwidth}
		\centering
		\includegraphics[width=\textwidth]{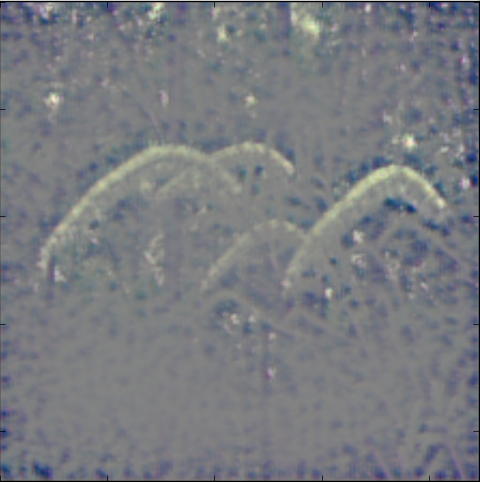}
	\end{subfigure}
	\begin{subfigure}[b]{0.2\textwidth}
		\centering
		\includegraphics[width=\textwidth]{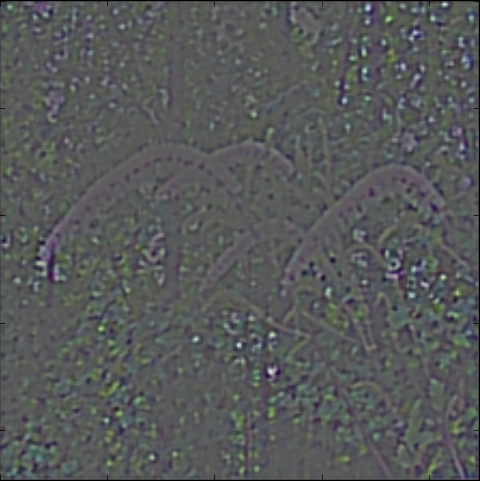}
	\end{subfigure}
	\begin{subfigure}[b]{0.2\textwidth}
		\centering
		\includegraphics[width=\textwidth]{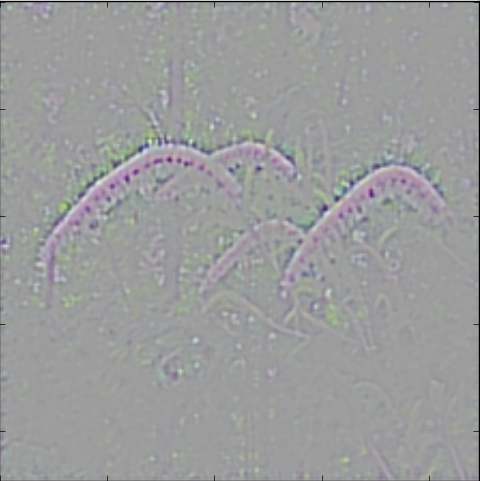}
	\end{subfigure}
	\begin{subfigure}[b]{0.2\textwidth}
		\centering
		\includegraphics[width=\textwidth]{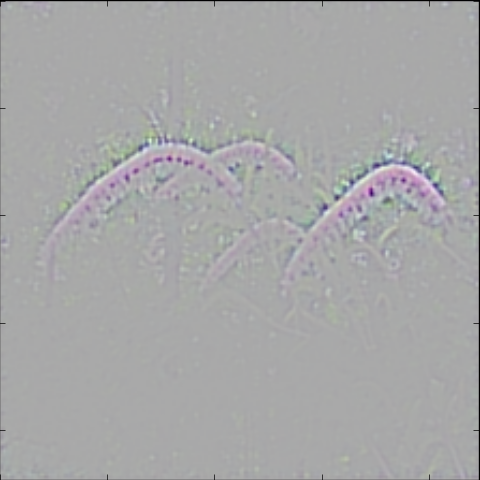}
	\end{subfigure}
	
	\begin{subfigure}[b]{0.2\textwidth}
		\centering
		\includegraphics[width=\textwidth]{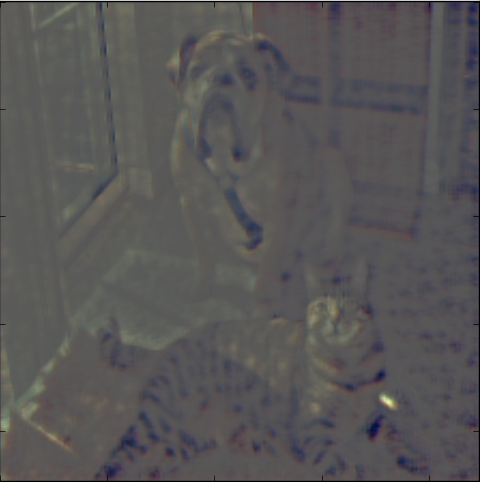}
		\caption*{Conv1-1$^\diamond$}
	\end{subfigure}
	\begin{subfigure}[b]{0.2\textwidth}
		\centering
		\includegraphics[width=\textwidth]{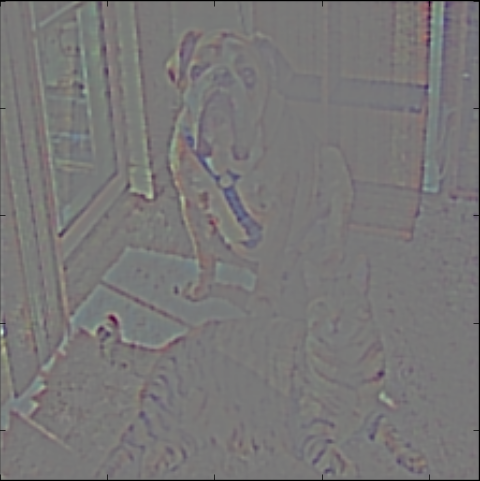}
		\caption*{Conv3-1$^\diamond$}
	\end{subfigure}
	\begin{subfigure}[b]{0.2\textwidth}
		\centering
		\includegraphics[width=\textwidth]{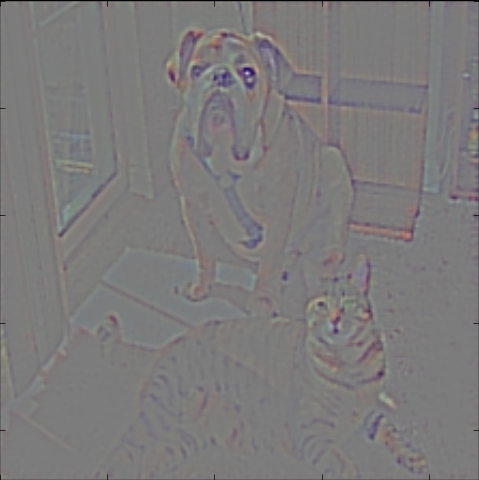}
		\caption*{Conv5-1$^\diamond$}
	\end{subfigure}
	\begin{subfigure}[b]{0.2\textwidth}
		\centering
		\includegraphics[width=\textwidth]{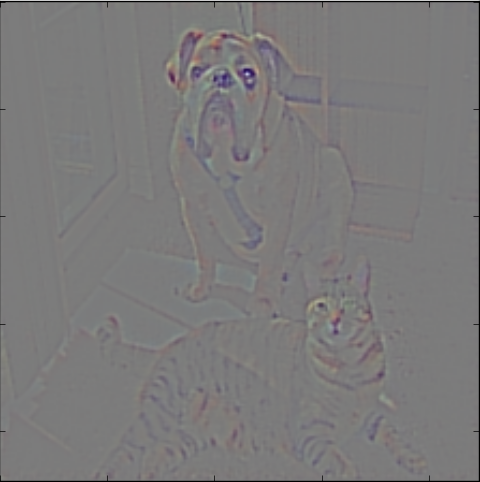}
		\caption*{FC3$^\diamond$}
	\end{subfigure}
	\caption{Load the trained weights of the VGG-16 net \textbf{except for} the indexed layer which is randomly initialized instead (denoted by the diamond sign) with different input images.}\label{fig_11}
	\end{minipage}
	
\end{figure*}

\begin{figure*}[!h]
	\begin{minipage}[c]{0.48\textwidth}
	\centering
	\begin{subfigure}[b]{0.3\textwidth}
		\centering
		\includegraphics[width=\textwidth]{tabby}
		\caption*{\footnotesize tabby}
	\end{subfigure}
	
	\begin{subfigure}[b]{0.3\textwidth}
		\centering
		\includegraphics[width=\textwidth]{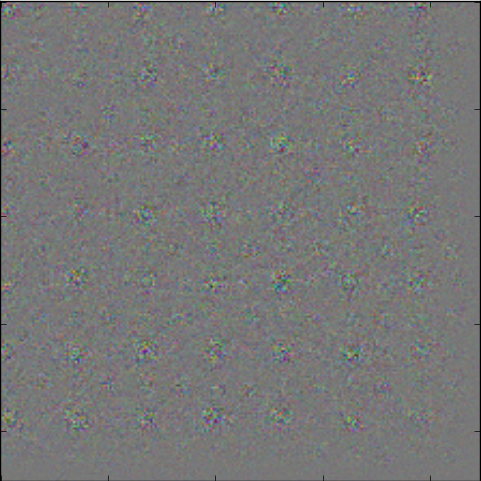}
		\caption*{\footnotesize Sal-max}
	\end{subfigure}
	\begin{subfigure}[b]{0.3\textwidth}
		\centering
		\includegraphics[width=\textwidth]{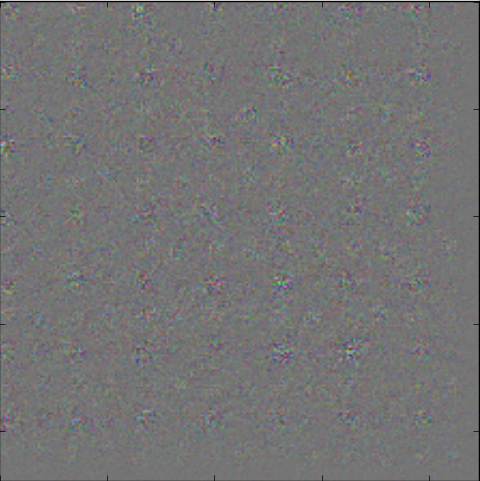}
		\caption*{\footnotesize Sal-31}
	\end{subfigure}
	\begin{subfigure}[b]{0.3\textwidth}
		\centering
		\includegraphics[width=\textwidth]{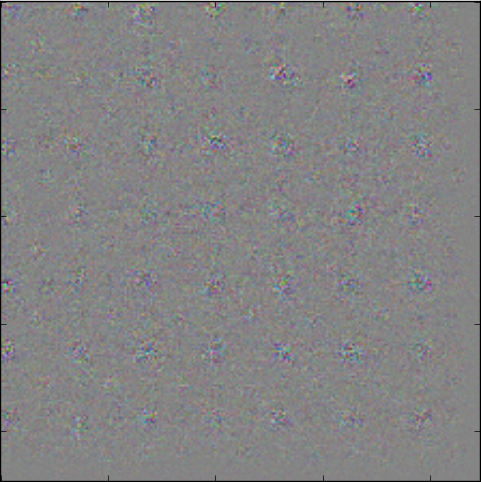}
		\caption*{\footnotesize Sal-85}
	\end{subfigure}
    
    \begin{subfigure}[b]{0.3\textwidth}
		\centering
		\includegraphics[width=\textwidth]{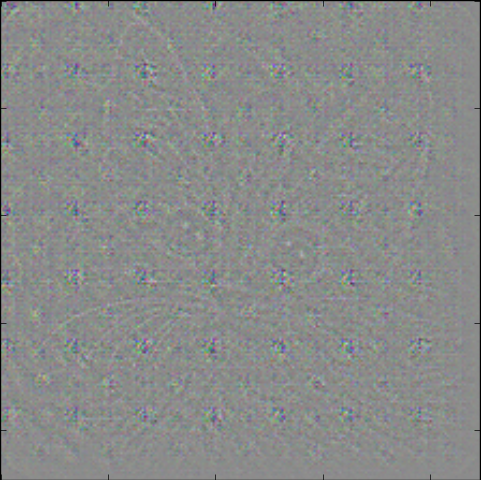}
		\caption*{\footnotesize Deconv-max}
	\end{subfigure}
	\begin{subfigure}[b]{0.3\textwidth}
		\centering
		\includegraphics[width=\textwidth]{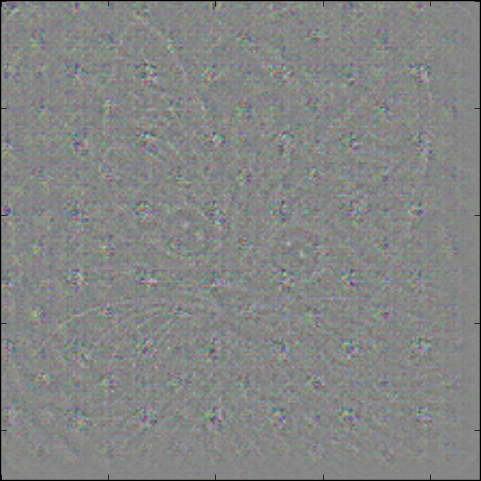}
		\caption*{\footnotesize Deconv-31}
	\end{subfigure}
	\begin{subfigure}[b]{0.3\textwidth}
		\centering
		\includegraphics[width=\textwidth]{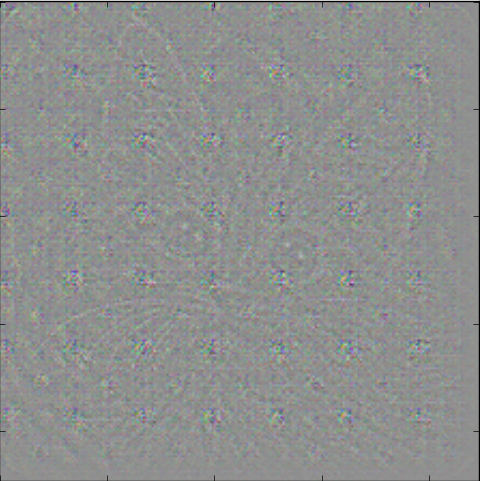}
		\caption*{\footnotesize Deconv-85}
	\end{subfigure}

	\begin{subfigure}[b]{0.3\textwidth}
		\centering
		\includegraphics[width=\textwidth]{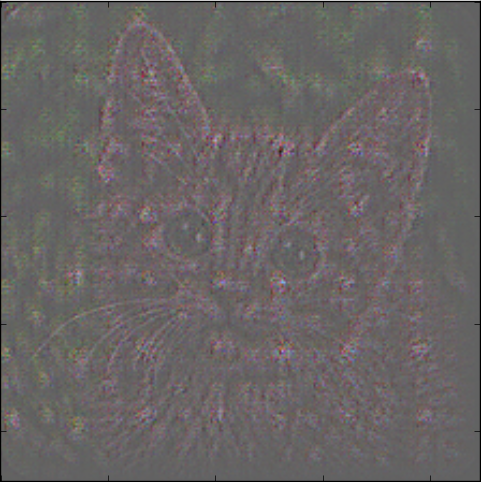}
		\caption*{\footnotesize GBP-max}
	\end{subfigure}
	\begin{subfigure}[b]{0.3\textwidth}
		\centering
		\includegraphics[width=\textwidth]{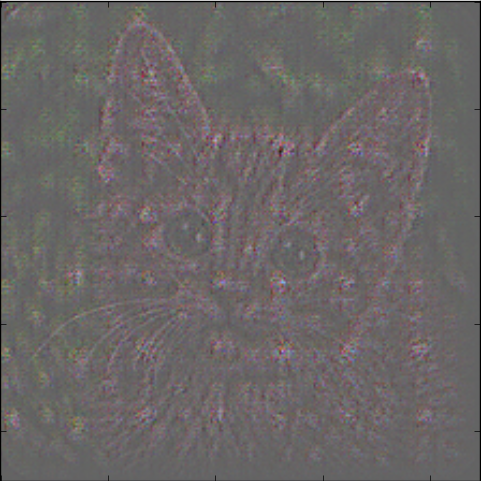}
		\caption*{\footnotesize GBP-31}
	\end{subfigure}
	\begin{subfigure}[b]{0.3\textwidth}
		\centering
		\includegraphics[width=\textwidth]{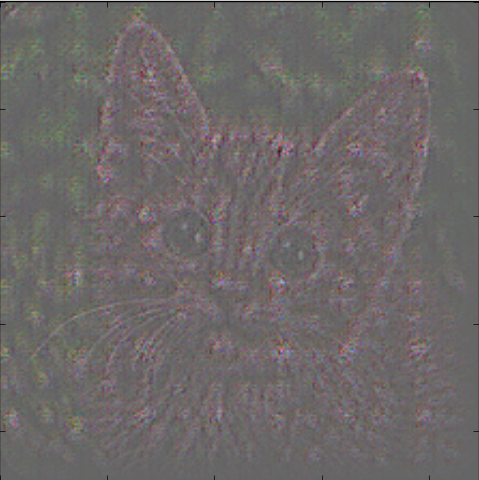}
		\caption*{\footnotesize GBP-85}
	\end{subfigure}
	
	\caption{Saliency map, DeconvNet and GBP visualizations for the random ResNet-50 with the input image ``tabby''.}\label{fig_12}
	\end{minipage}
	\hfill
	\begin{minipage}[c]{0.48\textwidth}
	\centering
	\begin{subfigure}[b]{0.3\textwidth}
		\centering
		\includegraphics[width=\textwidth]{tabby}
		\caption*{\footnotesize tabby}
	\end{subfigure}
	
	\begin{subfigure}[b]{0.3\textwidth}
		\centering
		\includegraphics[width=\textwidth]{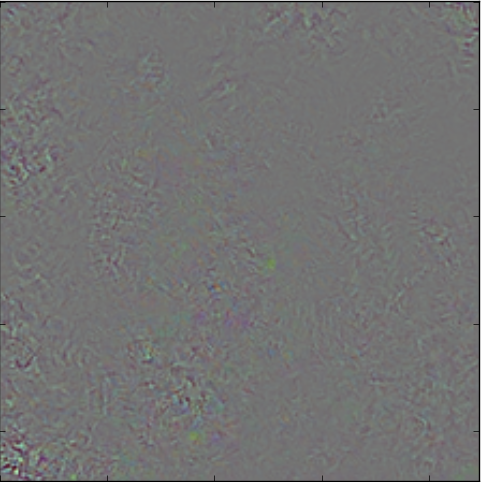}
		\caption*{\footnotesize Sal-max}
	\end{subfigure}
	\begin{subfigure}[b]{0.3\textwidth}
		\centering
		\includegraphics[width=\textwidth]{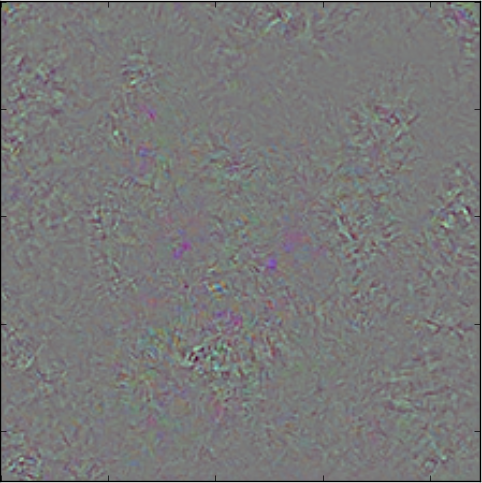}
		\caption*{\footnotesize Sal-31}
	\end{subfigure}
	\begin{subfigure}[b]{0.3\textwidth}
		\centering
		\includegraphics[width=\textwidth]{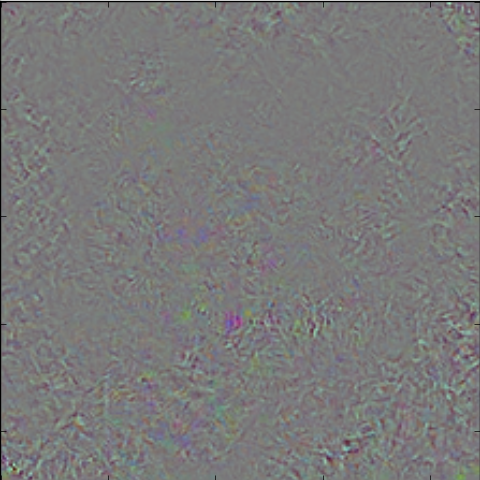}
		\caption*{\footnotesize Sal-85}
	\end{subfigure}
    
    \begin{subfigure}[b]{0.3\textwidth}
		\centering
		\includegraphics[width=\textwidth]{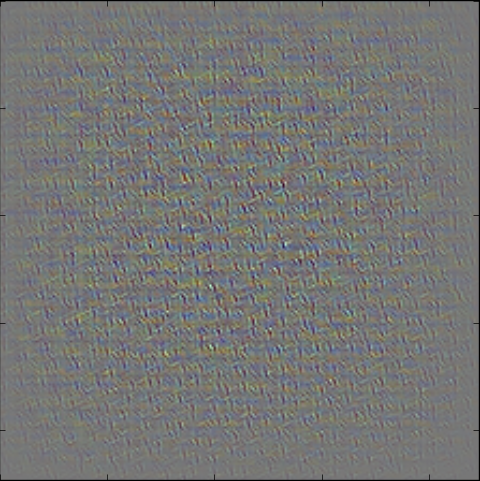}
		\caption*{\footnotesize Deconv-max}
	\end{subfigure}
	\begin{subfigure}[b]{0.3\textwidth}
		\centering
		\includegraphics[width=\textwidth]{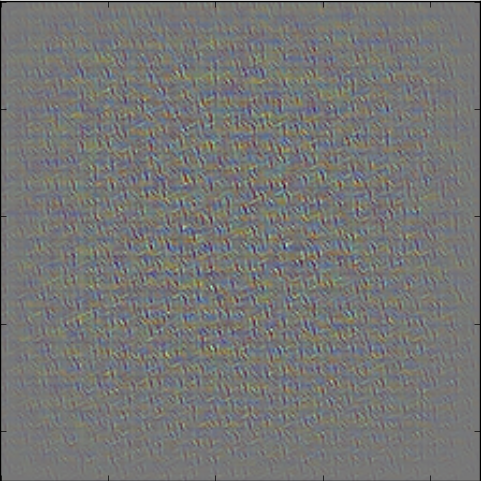}
		\caption*{\footnotesize Deconv-31}
	\end{subfigure}
	\begin{subfigure}[b]{0.3\textwidth}
		\centering
		\includegraphics[width=\textwidth]{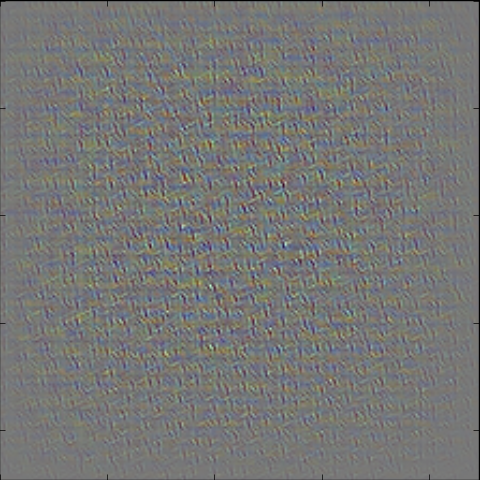}
		\caption*{\footnotesize Deconv-85}
	\end{subfigure}

	\begin{subfigure}[b]{0.3\textwidth}
		\centering
		\includegraphics[width=\textwidth]{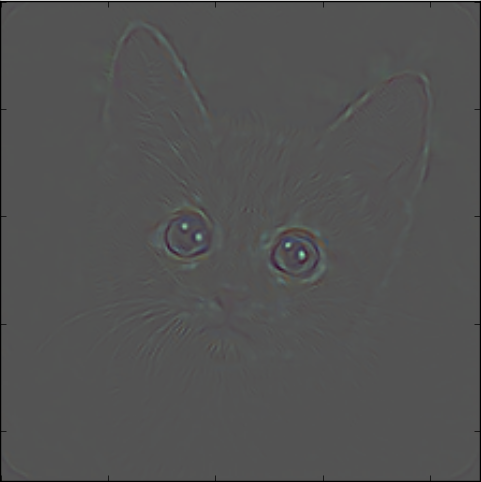}
		\caption*{\footnotesize GBP-max}
	\end{subfigure}
	\begin{subfigure}[b]{0.3\textwidth}
		\centering
		\includegraphics[width=\textwidth]{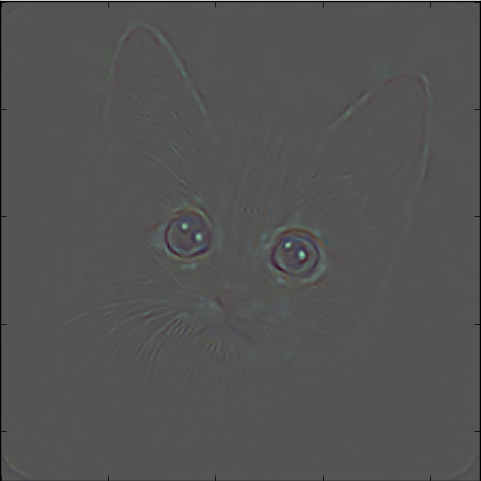}
		\caption*{\footnotesize GBP-31}
	\end{subfigure}
	\begin{subfigure}[b]{0.3\textwidth}
		\centering
		\includegraphics[width=\textwidth]{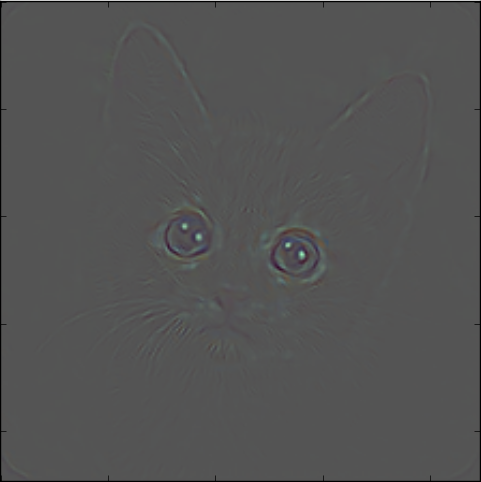}
		\caption*{\footnotesize GBP-85}
	\end{subfigure}
	
	\caption{Saliency map, DeconvNet and GBP visualizations for the trained ResNet-50 net with the input image ``tabby''.}\label{fig_13}
	\end{minipage}
	
\end{figure*}

Our theoretical analysis shows that for GBP it is the local connections in CNNs, together with the backward ReLU, that contribute to the clean-looking visualizations. Here we further investigate backpropagation-based visualizations on both randomly initialized (Figure~\ref{fig_11}) and trained (Figure~\ref{fig_12}) ResNet-50. In general, the results are very similar to those in the VGG-16 net.
However, we do observe some additional grid-like textures here and we conjecture that this deterioration of visual quality is due to the skip connections, as we have shown earlier that network structure has a significant impact on the visualizations.
We leave the rigorous analysis of this phenomenon for future work.

\end{document}